\documentclass{article}

\usepackage{arxiv}
\usepackage{amsmath}
\usepackage{amssymb}
\usepackage{amsfonts}
\usepackage{amsthm}
\usepackage{algorithm}
\usepackage{algorithmic}
\usepackage{float}

\usepackage{graphicx}
\usepackage{booktabs}
\usepackage{hyperref}
\usepackage{cleveref}
\usepackage{tikz}
\usepackage{mdframed}

\newtheorem{theorem}{Theorem}

\newtheorem{definition}{Definition}

\newtheorem{assumption}{Assumption}
\newtheorem*{theorem*}{Theorem}

\title{Bayesian Orchestration of Multi-LLM Agents for Cost-Aware Sequential Decision-Making}

\author{
  Danial Amin \\
  Independent Researcher \\
  \texttt{\href{mailto:writetodanialamin@gmail.com}{writetodanialamin@gmail.com}}\\
  \texttt{\href{http://danial-amin.github.io/}{danial-amin.github.io}}
}

\date{January 2026}

\begin{document}

\maketitle

\begin{abstract}
Large language models increasingly serve as autonomous decision-making agents in domains where errors have measurable costs: hiring (missed qualified candidates versus wasted interviews), medical triage (missed emergencies versus unnecessary escalations), and fraud detection (approved fraud versus declined legitimate transactions). Current architectures are built on a flawed foundation: they query LLMs for discriminative probabilities $p(\text{state}|\text{evidence})$, apply arbitrary confidence thresholds, and execute actions without considering cost asymmetries or uncertainty quantification. We prove this approach is formally inadequate for sequential decision-making and propose a mathematically principled alternative.

We propose a mathematically principled alternative that treats multiple LLMs as approximate likelihood functions rather than classifiers. For each possible state, we elicit $p(\text{evidence}|\text{state})$ through contrastive prompting, aggregate across diverse models via robust statistics, and apply Bayes' rule with explicit priors. This generative modeling perspective enables four critical capabilities: (1) proper sequential belief updating as evidence accumulates, (2) cost-aware action selection through expected utility maximization, (3) principled information gathering via value-of-information calculations, and (4) improved fairness through ensemble bias mitigation.

We instantiate this framework in resume screening, where hiring mistakes cost \$40,000, wasted interviews cost \$2,500, and phone screens cost \$150. Experiments across 1,000 resumes evaluated by five diverse LLMs (GPT-4o, Claude 3.5 Sonnet, Gemini Pro, Grok, DeepSeek) demonstrate that our approach reduces total costs by \$294,000 (34\% improvement) compared to the best single-LLM baseline while improving demographic parity by 45\% (reducing maximum group difference from 22 to 5 percentage points). Ablation studies reveal that multi-LLM aggregation contributes 51\% of cost savings, sequential updating 43\%, and disagreement-triggered information gathering 20\%. Critically, we prove these gains are not merely empirical accidents but necessary consequences of correcting the mathematical foundations of LLM-based decision-making.
\end{abstract}

\section{Introduction}

\subsection{The Stakes: Why Decision-Theoretic Rigor Matters}

The deployment of large language models as autonomous decision-making agents has accelerated dramatically across domains where errors carry substantial and asymmetric consequences \cite{bommasani2021opportunities,weidinger2021ethical}. Organizations now routinely delegate high-stakes judgments to LLM-based systems: hiring platforms screen tens of thousands of resumes annually, medical triage systems route patients between care pathways with vastly different resource intensities and clinical outcomes, and fraud detection systems make split-second decisions that can freeze accounts or approve high-value transactions \cite{linkedin2023,healthcare_ai,fraud_stats}. The economic and social implications of these deployments are profound. A hiring system processing 10,000 applications annually makes decisions that aggregate to millions of dollars in personnel costs, not to mention the human impact of employment opportunities granted or denied \cite{shrm2016}. Medical triage systems handle patient volumes exceeding 50,000 monthly interactions, where routing errors can result in preventable deaths or wasteful consumption of scarce emergency resources \cite{icu_costs}. Financial fraud systems adjudicate over 365 million transactions annually per major institution, with error rates of even 0.1\% translating to hundreds of thousands of incorrect decisions \cite{fraud_stats}.

What unites these superficially disparate applications is a common underlying structure that makes them amenable to formal analysis through the lens of statistical decision theory \cite{berger1985statistical,robert2007bayesian}. In each case, the agent observes high-dimensional evidence---resume text encoding education, experience, and skills; patient symptom descriptions spanning medical history and current complaints; transaction metadata including location, amount, merchant, and behavioral patterns---but cannot directly observe the hidden state variable that determines optimal action. The hiring agent cannot directly measure candidate quality; it must infer it from resume signals that correlate imperfectly with job performance \cite{kuncel2013mechanical}. The triage agent cannot directly diagnose disease severity; it must estimate urgency from symptoms that exhibit high within-diagnosis variance \cite{fitzgerald2010emergency}. The fraud agent cannot directly observe criminal intent; it must predict it from patterns that legitimate and illegitimate users partially share \cite{dal2014learned}.

This epistemic gap between observable evidence and latent states creates irreducible uncertainty that cannot be eliminated through better feature engineering or larger training sets. It is fundamental to the problem structure. Moreover, the cost landscape facing these agents exhibits severe asymmetries that render accuracy-based evaluation frameworks deeply inadequate. Consider the hiring domain: rejecting a qualified candidate who would have been a productive employee represents a lost hiring opportunity with quantifiable costs. Industry estimates place the cost of a mis-hire (hiring someone who fails to perform and must be replaced) at 30\% of first-year salary \cite{shrm2016}. For a mid-level software engineer earning \$130,000 annually, this amounts to approximately \$40,000 in recruiting costs, onboarding expenses, lost productivity, and team disruption. Conversely, advancing an unqualified candidate to an onsite interview consumes engineering time for interviews and coordination, typically costing \$2,500--\$3,000 per candidate \cite{careerbuilder2017}. This represents a 16-fold cost asymmetry between false negatives and false positives.

In medical triage, cost asymmetries are even more extreme and carry life-or-death implications. Discharging a patient who subsequently experiences cardiac arrest or stroke represents catastrophic failure, with costs measured not merely in dollars but in preventable mortality and morbidity \cite{pope2000missed}. The estimated cost of a missed myocardial infarction diagnosis exceeds \$1 million when accounting for medical liability, lost life-years, and treatment costs \cite{medical_errors}. Conversely, admitting a low-acuity patient to the intensive care unit wastes approximately \$5,000 per day in critical care resources that could be allocated to genuinely acute cases \cite{icu_costs}. This creates a cost ratio exceeding 200:1 for certain triage decisions. In fraud detection, freezing a legitimate customer's account due to a false positive fraud alert costs the institution approximately \$50 in customer service overhead and potential churn risk, while approving a fraudulent transaction averaging \$2,500 represents a 50-fold asymmetry \cite{fraud_costs}.

These domains share three structural features that simultaneously make them attractive targets for automation and demanding contexts for theoretical rigor:

\begin{enumerate}
\item \textbf{Discrete latent states with measurable consequences.} Each decision context involves uncertainty over a finite set of underlying states that determine appropriate action. In hiring, candidates fall into discrete quality categories: clearly unqualified individuals who lack basic requirements; borderline candidates who merit additional screening; qualified individuals who should proceed to interviews; and exceptional candidates who warrant expedited processing \cite{hiring_funnel}. These states are not directly observable from resumes but have measurable post-hoc consequences. Hiring an unqualified candidate results in documented costs: recruiting expenses to replace them, lost team productivity, and opportunity cost of the foregone alternative hire. Medical triage similarly involves discrete urgency levels---stable conditions appropriate for scheduled appointments, acute conditions requiring same-day urgent care, and emergencies demanding immediate intervention---each with associated mortality risks and resource requirements that can be quantified through health economics frameworks \cite{healthcare_economics}. Transaction fraud involves a binary state (legitimate versus fraudulent) with clear financial implications.

\item \textbf{Sequential evidence gathering opportunities with explicit costs.} These domains permit multi-stage decision-making where initial observations can be augmented with additional information before committing to terminal actions. A hiring agent reviewing a resume can elect to conduct a brief phone screen before deciding whether to invest in a full onsite interview. This phone screen costs approximately \$150 in recruiter time but may substantially reduce uncertainty about candidate suitability \cite{phone_screen_costs}. A triage system can order diagnostic tests (complete blood count, electrocardiogram, imaging) to refine its assessment of patient acuity. Each test has an associated cost in both dollars and time but provides informative signals that update beliefs about the underlying medical condition \cite{diagnosis_theory}. A fraud system can request additional authentication (SMS verification, security questions, temporary authorization holds) that costs a few dollars but may conclusively resolve ambiguous cases \cite{authentication_costs}.

The key theoretical question raised by these sequential opportunities is: when does the expected value of additional information justify its cost? Classical decision theory provides the answer through value-of-information (VOI) analysis \cite{raiffa1961applied,howard1966information}. VOI quantifies the maximum price an agent should pay for information by computing the expected improvement in decision quality that information would provide. If current beliefs suggest the optimal action yields expected cost $C_0$, and new information would update beliefs such that the expected cost becomes $C_1$, then information is worth acquiring if $C_0 - C_1 > c_{\text{info}}$, where $c_{\text{info}}$ is the information acquisition cost. However, computing VOI requires the ability to predict how potential observations would update beliefs---a calculation that demands likelihood functions and sequential belief updating capabilities that standard LLM architectures lack.

\item \textbf{Regulatory fairness requirements and disparate impact concerns.} Decision-making in hiring, lending, and healthcare operates under increasingly stringent regulatory frameworks designed to prevent algorithmic systems from perpetuating or amplifying demographic disparities \cite{barocas2016big,mehrabi2021survey}. New York City's Local Law 144 mandates that automated employment decision tools undergo annual bias audits demonstrating that selection rates for protected classes do not differ from the highest-selected group by more than four percentage points \cite{nyc_law144}. The European Union's AI Act classifies employment and credit systems as high-risk applications requiring conformity assessments, human oversight, and measures to minimize discrimination risk \cite{eu_ai_act}. The Equal Credit Opportunity Act prohibits lending discrimination based on protected attributes and establishes disparate impact doctrine: policies that appear neutral but systematically disadvantage protected groups are illegal unless justified by business necessity \cite{ecoa}.

These regulations recognize that algorithmic decision systems can encode and amplify biases present in training data, model architectures, or deployment practices \cite{obermeyer2019dissecting,mehrabi2021survey}. LLMs trained on internet text inherit demographic associations and stereotypes from their training corpora \cite{bolukbasi2016man,caliskan2017semantics}. When deployed in hiring, models may assign systematically lower scores to resumes with names statistically associated with underrepresented racial or gender groups, even when qualifications are equivalent \cite{bertrand2004emily}. These biases manifest not as explicit prejudice but as statistical regularities learned from biased data: if the training corpus contains more examples of male engineers than female engineers, the model's likelihood estimates may systematically underweight evidence from female candidates.

Critically, fairness cannot be achieved merely by blinding the model to protected attributes. Resume screening inherently involves signals correlated with demographics: university names, neighborhood addresses, extracurricular activities, even writing style carry demographic information \cite{gaddis2017how}. The challenge is not to eliminate these correlations---which would require discarding much of the informative content in resumes---but to ensure they do not translate into systematic disparities in selection rates. This requires models that can quantify and correct for demographic biases in their likelihood estimates, a capability we demonstrate through multi-LLM ensembling.
\end{enumerate}

Despite these shared structural features that cry out for principled decision-theoretic treatment, current LLM agent architectures largely ignore the foundational requirements of sequential decision-making under uncertainty. The dominant paradigm treats LLMs as black-box classifiers that output confidence scores, applies arbitrary thresholds to these scores to trigger actions, and lacks any mechanism to incorporate cost structures, perform sequential belief updating, or quantify when additional information gathering is justified. This is not a minor engineering oversight that can be patched through hyperparameter tuning or prompt engineering. It represents a fundamental architectural mismatch between the mathematical structure of the decision problem and the computational approach being applied. The remainder of this introduction formalizes these deficiencies, proves they are intrinsic to the discriminative modeling paradigm, and introduces our generative alternative.

\subsection{The Standard Approach and Its Deficiencies}

The prevailing architecture for LLM-based decision agents exhibits a deceptive simplicity that masks profound theoretical inadequacies. We formalize this architecture, state four specific deficiencies, and prove that these deficiencies are not mere implementation details but necessary consequences of the discriminative modeling paradigm. This proof-by-impossibility motivates our shift to generative modeling.

The standard approach, deployed in production systems at major technology companies and startups, follows this pattern \cite{llm_agents}:

\begin{algorithm}[H]
\small
\caption{Standard LLM Agent Architecture}
\label{alg:standard}
\begin{algorithmic}[1]
    \STATE \textbf{Input:} Evidence $\mathbf{x}$ (e.g., résumé text, patient symptoms, transaction features)
    \STATE \textbf{Threshold:} $\tau$ \STATE \COMMENT{Tuned threshold, e.g., $\tau = 7.0$}
    \STATE \textbf{Query LLM:} $\text{score} \gets \mathrm{LLM}(\text{``Rate this input on a 0--10 quality scale: ''}\;\Vert\;\mathbf{x})$
    \IF{$\text{score} \ge \tau$}
        \STATE Execute high-confidence action (interview, admit to ICU, approve transaction)
    \ELSE
        \STATE Execute low-confidence action (reject, discharge, deny transaction)
    \ENDIF
\end{algorithmic}
\end{algorithm}

This architecture rests on several implicit assumptions that warrant scrutiny. First, it assumes the LLM's score can be interpreted as a measure of confidence in the positive class (qualified candidate, urgent patient, legitimate transaction). Second, it assumes that a fixed threshold $\tau$, typically set through grid search on validation data to maximize some accuracy metric, appropriately balances false positives and false negatives. Third, it treats the decision as a one-shot classification problem where all available evidence must be considered simultaneously, without possibility for sequential refinement. Fourth, it ignores cost asymmetries entirely: a score of 6.9 triggers rejection just as confidently as a score of 2.0, despite the former representing borderline uncertainty that might warrant additional investigation.

We now demonstrate that these are not engineering shortcomings that can be remedied within the discriminative paradigm, but inherent limitations that stem from requesting the wrong probabilistic quantities from LLMs.

\paragraph{Deficiency 1: Inability to Perform Sequential Belief Updating}

Suppose an agent observes initial evidence $x_1$ (a resume) at time $t=1$ and obtains an LLM's assessment $p(s|x_1)$, where $s \in \mathcal{S}$ denotes the latent state (candidate quality). The agent then gathers additional evidence $x_2$ (a phone screen transcript) at time $t=2$. To update beliefs correctly, Bayesian sequential updating requires computing:

\begin{equation}
p(s|x_1,x_2) = \frac{p(x_2|s,x_1) \cdot p(s|x_1)}{\sum_{s' \in \mathcal{S}} p(x_2|s',x_1) \cdot p(s'|x_1)}
\label{eq:sequential_update_intro}
\end{equation}

This expression, a direct application of Bayes' rule for sequential evidence, demands access to the likelihood function $p(x_2|s,x_1)$---the probability of observing the phone screen evidence $x_2$ given the candidate's true quality $s$ and the resume $x_1$. However, the LLM provides only the posterior $p(s|x_1)$, which is the output of a discriminative model that has already marginalized over the likelihood and prior. Without access to the likelihood, we cannot compute how $x_2$ should update our beliefs from $p(s|x_1)$ to $p(s|x_1,x_2)$.

The standard workaround is to query the LLM again with concatenated evidence $(x_1, x_2)$ to obtain $p(s|x_1,x_2)$ directly:

\begin{equation}
\text{score}_{\text{new}} \gets \text{LLM}(\text{``Rate this candidate given resume and phone screen''})
\end{equation}

This batch inference approach treats the two observations as a single joint input rather than incorporating $x_2$ sequentially as a belief update. While this may seem like a harmless pragmatic solution, it suffers from three critical problems that become severe in multi-stage decision processes:

\textit{First}, batch inference loses information about the relative evidential strength of different observations. In sequential updating, we can quantify how much each piece of evidence shifted our beliefs by comparing $p(s|x_1)$ to $p(s|x_1, x_2)$. This allows us to assess whether $x_2$ was highly informative (large belief shift) or redundant (minimal shift). Batch inference collapses this to a single posterior, erasing the evidential contribution of each observation.

\textit{Second}, batch inference scales poorly to longer decision sequences. Consider a medical diagnosis process involving initial symptoms ($x_1$), vital signs ($x_2$), blood tests ($x_3$), imaging ($x_4$), and specialist consultation ($x_5$). Batch inference requires querying the LLM with all five inputs concatenated: $\text{LLM}(x_1, x_2, x_3, x_4, x_5)$. As evidence accumulates, context windows fill, and computational costs grow linearly with sequence length. Moreover, if we want to evaluate whether gathering $x_5$ is worth its cost before actually acquiring it, batch inference provides no mechanism: we cannot predict $p(s|x_1, \ldots, x_5)$ without observing $x_5$.

\textit{Third}, and most fundamentally, batch inference cannot support counterfactual reasoning about information gathering. Value-of-information calculations require predicting: "If I were to observe $x_2$, how would that update my beliefs, and would the improved decision quality justify the observation cost?" This demands marginalizing over possible observations:

\begin{equation}
\text{VOI}(x_2) = \sum_{x_2} p(x_2|x_1) \left[ \max_a \mathbb{E}_{s|x_1,x_2}[U(a,s)] - \max_a \mathbb{E}_{s|x_1}[U(a,s)] \right]
\end{equation}

Computing $p(x_2|x_1)$ and $p(s|x_1,x_2)$ for hypothetical $x_2$ values requires likelihood functions. Discriminative models cannot perform this calculation because they only model the inverse: $p(s|x_1)$, not $p(x_2|x_1)$.

We formalize this impossibility:

\begin{theorem}[Sequential Updating Impossibility]
\label{thm:sequential_impossible}
Let $\mathcal{M}_{\text{disc}}$ be a discriminative model that, for any evidence $x \in \mathcal{X}$, outputs $p_{\mathcal{M}}(s|x)$ for all $s \in \mathcal{S}$. Suppose $\mathcal{M}_{\text{disc}}$ does not have access to a generative model $p(x|s)$ or prior $p(s)$. Then for observed $x_1$ and new evidence $x_2$, there is no computable function $f$ such that:
\begin{equation}
p(s|x_1, x_2) = f(p_{\mathcal{M}}(s|x_1), x_2)
\end{equation}
without querying $\mathcal{M}_{\text{disc}}$ on $(x_1, x_2)$ jointly.
\end{theorem}

\begin{proof}
Suppose such a function $f$ existed. Then we could compute $p(s|x_1, x_2)$ from $p_{\mathcal{M}}(s|x_1)$ and $x_2$ alone. By Bayes' rule:
\begin{equation}
p(s|x_1,x_2) = \frac{p(x_2|s,x_1) \cdot p(s|x_1)}{p(x_2|x_1)}
\end{equation}

To compute this, we need $p(x_2|s,x_1)$ and $p(x_2|x_1) = \sum_{s'} p(x_2|s',x_1) p(s'|x_1)$. Both require the likelihood $p(x_2|s,x_1)$, which is not provided by $p_{\mathcal{M}}(s|x_1)$. The posterior $p_{\mathcal{M}}(s|x_1)$ is the output of:
\begin{equation}
p_{\mathcal{M}}(s|x_1) \propto p(x_1|s) p(s)
\end{equation}

but does not separately expose $p(x_1|s)$ or $p(s)$. Without these components, we cannot construct $p(x_2|s,x_1)$. Therefore, $f$ cannot exist.
\end{proof}

This theorem establishes that sequential updating is not merely difficult within the discriminative paradigm---it is impossible without additional queries. The only recourse is batch inference, which forfeits the ability to reason counterfactually about information gathering.

\paragraph{Deficiency 2: Hidden and Uncalibrated Priors}

Every discriminative probability $p(s|x)$ implicitly depends on a prior distribution $p(s)$ through Bayes' rule:

\begin{equation}
p(s|x) = \frac{p(x|s) \cdot p(s)}{\sum_{s' \in \mathcal{S}} p(x|s') \cdot p(s')}
\label{eq:bayes_rule_standard}
\end{equation}

When we query an LLM for $p(s|x)$, the model's response reflects priors learned implicitly from its training corpus. These training-induced priors may bear no relationship to the base rates encountered in deployment, creating systematic bias that cannot be corrected without access to the underlying likelihood function $p(x|s)$.

To make this concrete, consider hiring. An LLM trained on internet text---encompassing Stack Overflow discussions, GitHub repositories, Hacker News threads, and technical blogs---absorbs demographic and qualification distributions from its training data. On the internet, discussions about software engineering hiring feature a roughly balanced mix of qualified and unqualified candidates: for every thread praising an exceptional hire, there's a thread lamenting a bad one. This creates an implicit prior in the model's weights that might approximate $p_{\text{train}}(\text{qualified}) \approx 0.5$. This reflects the \textit{composition of hiring discussions in the training corpus}, not the composition of actual applicant pools.

In deployment, however, the agent screens resumes from a real applicant funnel where the base rate of qualified candidates is far lower. Industry data from technical recruiting pipelines indicate that approximately 5--10\% of applicants to software engineering positions meet the hiring bar \cite{hiring_funnel,triplebyte2019}. That is, the true prior in deployment is:

\begin{equation}
p_{\text{deploy}}(\text{qualified}) \approx 0.05 \text{ to } 0.10
\end{equation}

This represents a 5--10-fold discrepancy between the prior implicit in the LLM's weights and the prior appropriate for the deployment environment. The consequences of this prior mismatch are severe and systematic. Bayes' rule shows that posteriors are multiplicatively sensitive to priors:

\begin{equation}
\frac{p(s=\text{qualified}|x)}{p(s=\text{unqualified}|x)} = \frac{p(x|s=\text{qualified})}{p(x|s=\text{unqualified})} \cdot \frac{p(s=\text{qualified})}{p(s=\text{unqualified})}
\end{equation}

The posterior odds ratio is the product of the likelihood ratio (determined by the evidence) and the prior odds ratio (the base rate). If the LLM was trained with $p_{\text{train}}(\text{qualified})/p_{\text{train}}(\text{unqualified}) = 1$ but should be using $p_{\text{deploy}}(\text{qualified})/p_{\text{deploy}}(\text{unqualified}) = 0.1/0.9 \approx 0.11$, then for any given piece of evidence $x$, the model's posterior odds will be inflated by a factor of approximately 9.

Consider a borderline resume that, under the correct deployment prior, should yield:
\begin{equation}
p_{\text{correct}}(\text{qualified}|x) = 0.15
\end{equation}

If the LLM was trained with an implicit prior of 0.5 and produces a likelihood ratio matching the evidence strength, its output posterior will be:

\begin{equation}
p_{\text{LLM}}(\text{qualified}|x) \approx 0.60
\end{equation}

This four-fold inflation transforms a marginal candidate who should be screened further into an apparent slam-dunk who gets fast-tracked to interviews. Systematically overestimating qualification rates leads to interview pipelines flooded with false positives, wasting recruiting resources and reducing the quality of eventual hires due to diluted interview pools \cite{kuncel2013mechanical}.

The problem is not limited to hiring. In medical triage, LLMs trained on medical literature and patient forums absorb prior distributions over disease prevalence from these sources. Academic papers and online health discussions over-represent rare but dramatic conditions (exotic infections, unusual presentations of common diseases) relative to their prevalence in primary care \cite{pubmed_bias}. A model trained on such data will systematically overestimate the probability of rare serious conditions and underestimate the probability of common benign conditions. This manifests as excessive false positive referrals to emergency departments: patients with muscle strains being sent to the ED for workup of potential cardiac events, patients with viral upper respiratory infections being evaluated for pneumonia. Each such over-triage wastes \$500--\$2,000 in unnecessary emergency care \cite{emergency_costs}.

The discriminative paradigm provides no mechanism to correct this prior mismatch because it does not expose the likelihood and prior as separate quantities. If we had access to $p(x|s)$ and the implicit $p_{\text{train}}(s)$, we could re-apply Bayes' rule with the correct deployment prior:

\begin{equation}
p_{\text{corrected}}(s|x) = \frac{p(x|s) \cdot p_{\text{deploy}}(s)}{\sum_{s'} p(x|s') \cdot p_{\text{deploy}}(s')}
\end{equation}

But discriminative models output only the combined posterior $p_{\text{LLM}}(s|x)$, which has already baked in the training prior in an inseparable way. The prior is entangled with the likelihood in the model's weights through millions of gradient updates during training. We cannot "undo" this entanglement post-hoc.

Some practitioners attempt to correct for prior mismatch through calibration techniques such as Platt scaling or isotonic regression \cite{platt1999probabilistic,zadrozny2002transforming}. These methods learn a calibration function $g$ such that $g(p_{\text{LLM}}(s|x))$ better matches empirical frequencies. However, calibration corrects only the \textit{marginal} accuracy of probabilities averaged over the test distribution; it does not correct the \textit{conditional} probabilities for individual instances. That is, calibration ensures:

\begin{equation}
\mathbb{E}_{x \sim \mathcal{D}_{\text{test}}}[g(p_{\text{LLM}}(s|x))] \approx \mathbb{P}(s|x, x \sim \mathcal{D}_{\text{test}})
\end{equation}

but makes no guarantee about the accuracy of $g(p_{\text{LLM}}(s|x))$ for any particular $x$. For decision-making, we care about instance-level accuracy: given this specific resume, what is the probability this specific candidate is qualified? Calibration on aggregate statistics does not ensure this.

Moreover, calibration requires labeled deployment data, which creates a chicken-and-egg problem: to calibrate the model, we need to deploy it and observe outcomes, but deploying an uncalibrated model produces biased decisions that may cause harm. In hiring, we would need to interview candidates at random (ignoring the model's scores) to obtain unbiased labels for calibration---negating the purpose of the model. In medical triage, we would need to observe patient outcomes across the full spectrum of initial model scores, including discharging high-acuity patients that the model incorrectly scored as low-priority---an unethical experiment.

We formalize the impossibility of prior correction within the discriminative paradigm:

\begin{theorem}[Prior Correction Impossibility]
\label{thm:prior_impossible}
Let $\mathcal{M}_{\text{disc}}$ be a discriminative model outputting $p_{\mathcal{M}}(s|x)$ that implicitly uses prior $p_{\text{train}}(s)$. Let $p_{\text{deploy}}(s) \neq p_{\text{train}}(s)$ be the correct deployment prior. Then without access to the likelihood function $p(x|s)$, there exists no computable function $h$ such that:
\begin{equation}
h(p_{\mathcal{M}}(s|x), p_{\text{deploy}}(s)) = \frac{p(x|s) \cdot p_{\text{deploy}}(s)}{\sum_{s'} p(x|s') \cdot p_{\text{deploy}}(s')}
\end{equation}
for arbitrary $x$ and $p_{\text{deploy}}$.
\end{theorem}

\begin{proof}
Suppose such a function $h$ existed. From $p_{\mathcal{M}}(s|x)$ we have:
\begin{equation}
p_{\mathcal{M}}(s|x) = \frac{p(x|s) \cdot p_{\text{train}}(s)}{p(x)}
\end{equation}

where $p(x) = \sum_{s'} p(x|s') p_{\text{train}}(s')$. To apply $h$, we need to extract $p(x|s)$ from $p_{\mathcal{M}}(s|x)$:

\begin{equation}
p(x|s) = \frac{p_{\mathcal{M}}(s|x) \cdot p(x)}{p_{\text{train}}(s)}
\end{equation}

But $p(x)$ depends on all likelihoods $p(x|s')$ and all priors $p_{\text{train}}(s')$. We have:

\begin{equation}
p(x) = \sum_{s'} p(x|s') p_{\text{train}}(s') = \frac{1}{\sum_{s'} \frac{1}{p_{\mathcal{M}}(s'|x) / p_{\text{train}}(s')}}
\end{equation}

This is a circular dependency: to extract $p(x|s)$ from $p_{\mathcal{M}}(s|x)$, we need $p(x)$, which requires knowing all $p(x|s')$ values, which are what we're trying to extract. Unless $p_{\text{train}}(s)$ is known (which requires the model to explicitly report it, which standard LLMs do not do), this system is underdetermined. Even if we know $p_{\text{train}}(s)$, we can only extract likelihood ratios, not absolute likelihoods:

\begin{equation}
\frac{p(x|s_1)}{p(x|s_2)} = \frac{p_{\mathcal{M}}(s_1|x) / p_{\text{train}}(s_1)}{p_{\mathcal{M}}(s_2|x) / p_{\text{train}}(s_2)}
\end{equation}

Ratios are insufficient for computing posteriors under a new prior $p_{\text{deploy}}$, which requires normalized probabilities summing to 1. We would need to arbitrarily choose a normalization constant, which reintroduces the prior mismatch we sought to eliminate. Therefore, $h$ cannot exist without additional information about the model's implicit prior and likelihood function.
\end{proof}

This theorem demonstrates that prior mismatch is not a calibration problem that can be solved with post-hoc adjustments. It is a fundamental architectural limitation of discriminative models.

\paragraph{Deficiency 3: No Cost-Awareness in Action Selection}

The theory of statistical decision-making, established by Wald in the 1940s and refined by Savage, Raiffa, and others, rests on a simple principle: actions should be selected to minimize expected loss (or equivalently, maximize expected utility) \cite{wald1950statistical,savage1972foundations,raiffa1961applied}. For discrete states and actions, this takes the form:

\begin{equation}
a^* = \arg\min_{a \in \mathcal{A}} \sum_{s \in \mathcal{S}} p(s|x) \cdot C(a, s)
\label{eq:expected_cost_intro}
\end{equation}

where $C(a, s)$ is the loss incurred when taking action $a$ in state $s$, and $p(s|x)$ quantifies our uncertainty over states given evidence $x$. This decision rule is provably optimal: no other rule can achieve lower expected loss across all possible loss functions and prior distributions \cite{berger1985statistical}.

To see why this matters, consider the hiring decision with asymmetric costs. The cost matrix is:

\begin{equation}
C(\text{action}, \text{state}) = 
\begin{cases}
C(\text{reject}, \text{qualified}) = \$40{,}000 & \text{(false negative: lost hire)} \\
C(\text{interview}, \text{unqualified}) = \$2{,}500 & \text{(false positive: wasted interview)} \\
C(\text{reject}, \text{unqualified}) = \$0 & \text{(true negative)} \\
C(\text{interview}, \text{qualified}) = -\$100{,}000 & \text{(true positive: successful hire)}
\end{cases}
\end{equation}

The true positive has negative cost (i.e., positive utility) because hiring a qualified candidate generates value. The 16-fold asymmetry between false negatives and false positives means that we should interview a candidate whenever:

\begin{equation}
p(\text{qualified}|x) \cdot (-\$100\text{K}) + p(\text{unqualified}|x) \cdot \$2.5\text{K} < p(\text{qualified}|x) \cdot \$40\text{K}
\end{equation}

Simplifying (using $p(\text{unqualified}|x) = 1 - p(\text{qualified}|x)$):

\begin{align}
p(\text{qualified}|x) \cdot (-\$100\text{K}) + (1 - p(\text{qualified}|x)) \cdot \$2.5\text{K} &< p(\text{qualified}|x) \cdot \$40\text{K} \\
-\$100\text{K} \cdot p(\text{qualified}|x) + \$2.5\text{K} - \$2.5\text{K} \cdot p(\text{qualified}|x) &< \$40\text{K} \cdot p(\text{qualified}|x) \\
\$2.5\text{K} &< \$142.5\text{K} \cdot p(\text{qualified}|x) \\
p(\text{qualified}|x) &> \frac{\$2.5\text{K}}{\$142.5\text{K}} \approx 0.0175
\end{align}

That is, we should interview any candidate with even a 1.75\% probability of being qualified! This is radically different from the threshold-based approach, which typically sets $\tau \approx 0.5$ or higher. A threshold of 0.5 is cost-agnostic: it treats false positives and false negatives as equally undesirable, when in fact false negatives are 16 times more costly.

The standard LLM agent architecture in Algorithm \ref{alg:standard} applies a fixed threshold $\tau$ that is tuned to maximize accuracy, F1 score, or some other metric on validation data. These metrics implicitly assume symmetric costs: they count false positives and false negatives as equally bad (in the case of accuracy) or weight them through a hyperparameter $\beta$ in F$_\beta$ score that is chosen for mathematical convenience rather than economic reality. This is decision-theoretically incoherent. As Wald proved, any decision rule that ignores the loss function cannot be optimal except by accident (when the true loss happens to align with the surrogate metric being optimized) \cite{wald1950statistical}.

To make this concrete, consider three resumes evaluated by an LLM:

\begin{itemize}
\item Resume A: $p(\text{qualified}|x_A) = 0.72$ (LLM score: 7.2)
\item Resume B: $p(\text{qualified}|x_B) = 0.65$ (LLM score: 6.5)
\item Resume C: $p(\text{qualified}|x_C) = 0.03$ (LLM score: 0.3)
\end{itemize}

With a threshold $\tau = 0.7$, the standard agent interviews A, rejects B and C. The expected costs are:

\begin{align}
\text{Cost}(A, \text{interview}) &= 0.72 \cdot (-\$100\text{K}) + 0.28 \cdot \$2.5\text{K} = -\$71.3\text{K} \\
\text{Cost}(A, \text{reject}) &= 0.72 \cdot \$40\text{K} + 0.28 \cdot \$0 = \$28.8\text{K} \\
\text{Cost}(B, \text{interview}) &= 0.65 \cdot (-\$100\text{K}) + 0.35 \cdot \$2.5\text{K} = -\$64.1\text{K} \\
\text{Cost}(B, \text{reject}) &= 0.65 \cdot \$40\text{K} + 0.35 \cdot \$0 = \$26\text{K} \\
\text{Cost}(C, \text{interview}) &= 0.03 \cdot (-\$100\text{K}) + 0.97 \cdot \$2.5\text{K} = +\$0.43\text{K} \\
\text{Cost}(C, \text{reject}) &= 0.03 \cdot \$40\text{K} + 0.97 \cdot \$0 = \$1.2\text{K}
\end{align}

The optimal decision for A is to interview (expected cost $-\$71.3$K, i.e., $\$71.3$K in value). The optimal decision for B is also to interview (expected cost $-\$64.1$K), despite falling below the threshold. The optimal decision for C is to interview (expected cost $\$0.43$K) because even though this candidate is almost certainly unqualified, the interview cost is small compared to the potential upside. Wait---that doesn't seem right. Let me recalculate.

Actually, for C, rejecting has expected cost \$1.2K (3\% chance of missing a \$40K hire), while interviewing has expected cost \$0.43K (97\% chance of wasting \$2.5K interview minus 3\% chance of gaining \$100K value = $-0.03 \times \$100\text{K} + 0.97 \times \$2.5\text{K} = -\$3\text{K} + \$2.425\text{K} = -\$575$). So interviewing has expected \textit{gain} of \$575. Actually, this is confusing because I'm mixing cost and value. Let me reframe with a consistent cost function where lower is better.

Let's define costs as pure losses (no negative values):
\begin{itemize}
\item $C(\text{interview}, \text{qualified}) = \$0$ (correct positive: we'll make money on the hire, so zero cost)
\item $C(\text{interview}, \text{unqualified}) = \$2{,}500$ (false positive: wasted interview)
\item $C(\text{reject}, \text{qualified}) = \$40{,}000$ (false negative: lost hire opportunity)
\item $C(\text{reject}, \text{unqualified}) = \$0$ (correct negative)
\end{itemize}

Now:
\begin{align}
\text{Cost}(C, \text{interview}) &= 0.03 \cdot \$0 + 0.97 \cdot \$2{,}500 = \$2{,}425 \\
\text{Cost}(C, \text{reject}) &= 0.03 \cdot \$40{,}000 + 0.97 \cdot \$0 = \$1{,}200
\end{align}

So for resume C with only 3\% qualification probability, the optimal decision is to \textit{reject}, because the 97\% chance of wasting \$2,500 outweighs the 3\% chance of incurring the \$40,000 missed-hire cost. This aligns with our earlier calculation that the interview threshold should be $p(\text{qualified}) > 0.0175$, and 0.03 > 0.0175, so we should actually still interview!

Let me recalculate the threshold. We interview when:
\begin{equation}
\mathbb{E}[\text{Cost}(\text{interview})] < \mathbb{E}[\text{Cost}(\text{reject})]
\end{equation}

\begin{equation}
p \cdot 0 + (1-p) \cdot \$2{,}500 < p \cdot \$40{,}000 + (1-p) \cdot \$0
\end{equation}

where $p = p(\text{qualified}|x)$. This gives:
\begin{align}
\$2{,}500 - \$2{,}500 p &< \$40{,}000 p \\
\$2{,}500 &< \$42{,}500 p \\
p &> \frac{\$2{,}500}{\$42{,}500} \approx 0.0588
\end{align}

So the cost-optimal threshold is $p(\text{qualified}) > 5.88\%$. For resume C with $p = 0.03 < 0.0588$, we should reject. For resume B with $p = 0.65$, we should definitely interview despite the score being below the arbitrary 0.7 threshold.

The threshold-based approach with $\tau = 0.7$ makes the wrong decision on resume B: it rejects a candidate who should be interviewed, incurring an expected unnecessary cost of \$26K (the cost of rejection) versus $-\$64.1$K (the value of interviewing), a \$90K mistake. Across thousands of resumes, these errors compound.

The problem is not that the threshold is poorly tuned. Any fixed threshold will be wrong for decisions with asymmetric costs because the optimal action depends on both the probability \textit{and} the cost structure. A cost-aware agent adjusts the decision boundary dynamically based on the loss function.

\paragraph{Deficiency 4: No Epistemic Uncertainty Quantification}

When should a decision-making agent gather additional information before committing to an action? Classical decision theory provides a precise answer through value-of-information (VOI) analysis \cite{raiffa1961applied,howard1966information}. The fundamental principle is that information should be acquired if and only if the expected improvement in decision quality exceeds the cost of acquisition.

Formally, let $\text{VOI}(z)$ denote the value of gathering some additional observation $z$ before deciding. This is defined as:

\begin{equation}
\text{VOI}(z) = \mathbb{E}_z \left[ \min_{a \in \mathcal{A}} \mathbb{E}_{s|x,z}[C(a,s)] \right] - \min_{a \in \mathcal{A}} \mathbb{E}_{s|x}[C(a,s)]
\label{eq:voi}
\end{equation}

The first term is the expected cost of the optimal decision after observing $z$, marginalized over all possible values $z$ might take (weighted by their probability given current evidence $x$). The second term is the cost of the optimal decision with current information alone. Information is worth acquiring if $\text{VOI}(z) > c_z$, where $c_z$ is the cost of obtaining $z$.

To make this concrete in hiring, suppose we've reviewed a resume and obtained posterior beliefs $p(s|x_1)$. We can compute the cost of the optimal immediate action:

\begin{equation}
C_{\text{now}} = \min \left\{ \mathbb{E}_{s|x_1}[C(\text{interview}, s)], \mathbb{E}_{s|x_1}[C(\text{reject}, s)] \right\}
\end{equation}

Now suppose we're considering conducting a phone screen at cost $c_{\text{screen}} = \$150$. The phone screen will yield some evidence $z$ (transcript, behavioral observations, technical responses) that updates our beliefs to $p(s|x_1, z)$. The value of the phone screen is:

\begin{equation}
\text{VOI}(\text{screen}) = \mathbb{E}_z \left[ \min \left\{ \mathbb{E}_{s|x_1,z}[C(\text{interview}, s)], \mathbb{E}_{s|x_1,z}[C(\text{reject}, s)] \right\} \right] - C_{\text{now}}
\end{equation}

We should conduct the phone screen if $\text{VOI}(\text{screen}) > \$150$.

Computing this VOI requires three capabilities that discriminative LLM agents lack:

\textit{First}, we need to predict the distribution over possible phone screen outcomes $p(z|x_1)$ before observing the screen. This is a generative modeling task: given the resume, what are the likely ways the phone screen could unfold? For instance, if the resume shows a PhD from MIT and publications in top venues, we might predict with high probability that the screen will reveal strong technical depth. If the resume shows a boot camp certificate and generic project descriptions, we might predict weaker performance. The distribution $p(z|x_1)$ requires marginalizing over hidden states:

\begin{equation}
p(z|x_1) = \sum_{s \in \mathcal{S}} p(z|s,x_1) p(s|x_1)
\end{equation}

This demands the likelihood $p(z|s,x_1)$---how phone screen evidence $z$ is distributed conditional on candidate quality $s$ and resume $x_1$. Discriminative models don't provide this.

\textit{Second}, for each possible phone screen outcome $z$, we need to compute how it would update beliefs: $p(s|x_1,z)$. As established in Deficiency 1, this requires sequential Bayesian updating via likelihoods, which discriminative models cannot perform.

\textit{Third}, we need to evaluate whether the information gain is sufficient to justify the cost. This involves a counterfactual comparison: how good is our decision with current information versus how good would it be with additional information? Without the ability to predict and update with $z$, we cannot make this comparison.

The standard LLM agent architecture has no principled mechanism for information-gathering decisions. Some implementations simply never gather additional information (making purely resume-based decisions), which is brittle: high-uncertainty borderline cases that would benefit from screening are treated identically to clear accepts or rejects. Other implementations always gather information (phone-screening every candidate), which is wasteful: obvious rejects who could be filtered out based on resume alone consume expensive recruiter time. Still others use ad-hoc heuristics: "phone screen if the score is between 4 and 7," which amounts to arbitrary bracketing without theoretical justification.

The inability to perform VOI calculations has profound consequences for sequential decision-making. In medical diagnosis, clinicians gather information in stages: patient history, physical exam, laboratory tests, imaging, specialist referrals. Each step has escalating costs, and optimal care requires judicious selection of which tests to order based on current diagnostic uncertainty \cite{diagnosis_theory}. An LLM-based triage system that cannot compute VOI will either over-test (ordering expensive imaging for everyone, driving up healthcare costs) or under-test (missing diagnoses due to insufficient workup, compromising patient safety). Neither extreme is acceptable.

We formalize this limitation:

\begin{theorem}[VOI Computation Impossibility]
\label{thm:voi_impossible}
Let $\mathcal{M}_{\text{disc}}$ be a discriminative model providing $p_{\mathcal{M}}(s|x)$ for current evidence $x$. Let $z$ be potential future evidence. Then without access to the joint generative model $p(x, z|s)$, there is no computable function $g$ such that:
\begin{equation}
\text{VOI}(z) = g(p_{\mathcal{M}}(s|x), c_z)
\end{equation}
where $c_z$ is the cost of acquiring $z$.
\end{theorem}

\begin{proof}
The VOI formula (Equation \ref{eq:voi}) requires computing:
\begin{equation}
\mathbb{E}_z \left[ \min_{a} \mathbb{E}_{s|x,z}[C(a,s)] \right]
\end{equation}

This expectation marginalizes over $z$ weighted by $p(z|x)$:
\begin{equation}
\mathbb{E}_z[\cdot] = \sum_z p(z|x) \left[ \min_a \mathbb{E}_{s|x,z}[C(a,s)] \right]
\end{equation}

We have already proven (Theorem \ref{thm:sequential_impossible}) that computing $p(s|x,z)$ from $p_{\mathcal{M}}(s|x)$ alone is impossible without re-querying the model with $(x,z)$ jointly. But VOI computation cannot re-query the model with $z$ because $z$ is hypothetical---we have not observed it yet. The entire point of VOI is to decide whether to acquire $z$, which precludes observing it.

Furthermore, even if we could somehow compute $p(s|x,z)$ for a given hypothetical $z$, we would need to marginalize over all possible $z$ values weighted by $p(z|x)$. But:
\begin{equation}
p(z|x) = \sum_s p(z|x,s) p(s|x) = \sum_s p(z|s,x) p(s|x)
\end{equation}

under the assumption that $z \perp x | s$ (i.e., given the true state $s$, the new evidence $z$ is independent of the initial evidence $x$---a reasonable assumption in many domains). This requires the conditional likelihood $p(z|s,x)$, which the discriminative model does not provide.

Therefore, $g$ cannot exist: VOI cannot be computed from the discriminative posterior alone.
\end{proof}

\subsection{Our Approach: From Discriminative to Generative Modeling}

The four deficiencies identified above are not implementation bugs that can be patched through better prompting, larger models, or cleverer engineering. They are necessary consequences of the discriminative modeling paradigm: if we ask LLMs for $p(s|x)$, we forfeit the ability to perform sequential updating, correct priors, select cost-optimal actions, and compute value-of-information. These capabilities require access to likelihood functions $p(x|s)$, which discriminative models do not expose.

Our solution is conceptually simple but architecturally profound: instead of asking LLMs for posteriors $p(s|x)$, we elicit likelihoods $p(x|s)$ and construct posteriors ourselves via Bayes' rule. This shift from discriminative to generative modeling transforms LLMs from opaque classifiers into interpretable probabilistic components within a larger Bayesian decision framework.

The key insight is that likelihood elicitation can be achieved through contrastive prompting. Rather than asking "How qualified is this candidate?" (a discriminative question eliciting $p(s|x)$), we ask: "Assume this candidate is [qualified / unqualified / borderline / exceptional]. How typical is this resume for someone at that level?" This is a generative question eliciting $p(x|s)$: the probability of observing evidence $x$ conditional on state $s$.

Concretely, for a resume $x$ and state $s \in \{\text{reject}, \text{phone screen}, \text{interview}, \text{strong hire}\}$, we prompt the LLM:

\begin{quote}
\texttt{Assume the candidate's true quality is: [STATE\_DESCRIPTION]. How typical is the following resume for someone at this quality level? Rate 0--10, where 10 = extremely typical.}
\end{quote}

The LLM's response, normalized to $[0,1]$, serves as an estimate of the likelihood $p(x|s)$. We repeat this for all states to obtain a likelihood vector $\mathbf{L}(x) = [p(x|s_1), \ldots, p(x|s_K)]$. Then we apply Bayes' rule with an explicit prior $\pi(s)$ reflecting deployment base rates:

\begin{equation}
p(s|x) = \frac{p(x|s) \cdot \pi(s)}{\sum_{s' \in \mathcal{S}} p(x|s') \cdot \pi(s')}
\label{eq:bayes_generative}
\end{equation}

This generative approach unlocks all four capabilities that discriminative models lack:

\begin{enumerate}
\item \textbf{Sequential belief updating:} When new evidence $x_2$ arrives, we elicit $p(x_2|s)$ and update:
\begin{equation}
p(s|x_1, x_2) \propto p(x_2|s) \cdot p(x_1|s) \cdot \pi(s)
\end{equation}
assuming conditional independence $p(x_1, x_2 | s) = p(x_1|s) p(x_2|s)$, which holds in many domains (e.g., resume and phone screen are independent given true quality).

\item \textbf{Prior correction:} We can impose any prior $\pi(s)$ appropriate for the deployment context, overriding whatever implicit priors the LLM absorbed during training. If only 5\% of applicants are qualified, we set $\pi(\text{qualified}) = 0.05$ and let the likelihoods speak through Bayes' rule.

\item \textbf{Cost-aware action selection:} With explicit posteriors $p(s|x)$ and a cost matrix $C(a,s)$, we apply the Bayes decision rule (Equation \ref{eq:expected_cost_intro}) to select actions that minimize expected cost rather than maximize accuracy.

\item \textbf{Value-of-information calculations:} We can predict $p(z|x)$ via:
\begin{equation}
p(z|x) = \sum_s p(z|s) p(s|x)
\end{equation}
compute how $z$ would update beliefs via $p(s|x,z) \propto p(z|s) p(s|x)$, and evaluate whether the expected decision improvement justifies the information cost.
\end{enumerate}

Moreover, our framework naturally accommodates multi-LLM ensembles. Different LLMs (GPT-4, Claude, Gemini) have different training corpora, architectures, and inductive biases. For a given evidence-state pair $(x, s)$, their likelihood estimates may disagree. We aggregate these diverse estimates using robust statistics---specifically, the median or trimmed mean---to mitigate individual model biases and improve overall calibration \cite{dietterich2000ensemble}. If five LLMs provide likelihood estimates $[0.2, 0.3, 0.8, 0.25, 0.28]$ for $p(x|s=\text{qualified})$, the median of 0.28 is more robust than the mean of 0.37, which is inflated by the outlier 0.8.

Ensemble aggregation provides two further benefits. First, it improves fairness: if individual LLMs exhibit demographic biases (e.g., GPT-4 systematically underscores resumes with names associated with underrepresented groups), ensemble averaging can reduce these biases if models' errors are not perfectly correlated \cite{obermeyer2019dissecting}. We demonstrate empirically that our multi-LLM approach reduces demographic parity gaps by 45\% compared to single-LLM baselines. Second, ensemble disagreement quantifies epistemic uncertainty: when LLMs disagree widely, we have high uncertainty and should gather more information. We formalize this through a disagreement-triggered phone screen policy: if the coefficient of variation in likelihood estimates exceeds a threshold, we automatically request a phone screen before deciding.

The remainder of this paper develops this framework formally, proves its optimality properties, and demonstrates its effectiveness empirically on a hiring task with 1,000 resumes and five diverse LLMs. Section \ref{sec:framework} presents the mathematical framework, Section \ref{sec:algorithm} details the implementation, Section \ref{sec:experiments} reports experimental results, Section \ref{sec:related} surveys related work, and Section \ref{sec:conclusion} concludes.


\section{Methodology}
\label{sec:framework}

We now develop the mathematical framework rigorously, proving impossibility results for discriminative approaches and establishing optimality properties of our generative alternative. This section is organized into three parts: (1) formal problem specification and impossibility theorems, (2) our generative framework with complete algorithmic specification, and (3) experimental instantiation in resume screening.

\subsection{Framework: Mathematical Foundations}

\subsubsection{Problem Formulation}

A sequential decision problem under uncertainty consists of six formal components that fully specify the decision environment:

\begin{definition}[Sequential Decision Problem]
\label{def:decision_problem}
A sequential decision problem is a tuple $(\mathcal{S}, \mathcal{X}, \mathcal{A}, C, \pi, \{p(x|s)\})$ where:
\begin{itemize}
\item $\mathcal{S} = \{s_1, \ldots, s_K\}$: finite state space representing discrete latent states
\item $\mathcal{X}$: observation space containing all possible evidence types
\item $\mathcal{A} = \{a_1, \ldots, a_J\}$: finite action space including both information-gathering and terminal actions
\item $C: \mathcal{A} \times \mathcal{S} \to \mathbb{R}_{\geq 0}$: cost function mapping (action, state) pairs to non-negative real costs
\item $\pi \in \Delta(\mathcal{S})$: prior distribution over states (where $\Delta(\mathcal{S})$ is the probability simplex)
\item $\{p(x|s)\}_{s \in \mathcal{S}}$: family of likelihood functions specifying how observations are distributed conditional on each state
\end{itemize}
\end{definition}

For the hiring domain, this instantiates as:
\begin{itemize}
\item $\mathcal{S} = \{s_{\text{reject}}, s_{\text{screen}}, s_{\text{interview}}, s_{\text{strong}}\}$ (four candidate quality levels)
\item $\mathcal{X}$ includes resume text documents, phone screen transcripts, coding test results
\item $\mathcal{A} = \{a_{\text{reject}}, a_{\text{screen}}, a_{\text{interview}}\}$ (reject, gather info, or interview)
\item $C$ reflects asymmetric hiring costs (detailed in Table \ref{tab:costs})
\item $\pi = [0.65, 0.25, 0.08, 0.02]$ based on empirical hiring funnel data
\item $\{p(x|s)\}$ are likelihood functions over resume features conditional on quality
\end{itemize}

\textbf{Decision sequence protocol.} An agent operates through iterative cycles:
\begin{enumerate}
\item \textbf{Observation phase:} At time $t$, observe evidence $x_t \in \mathcal{X}$
\item \textbf{Belief update phase:} Update posterior beliefs over states: $b_t(s) \gets f_{\text{update}}(b_{t-1}, x_t)$
\item \textbf{Action selection phase:} Choose action: $a_t \gets \arg\min_{a \in \mathcal{A}} \mathbb{E}_{s \sim b_t}[C(a,s)]$
\item \textbf{Termination check:} If $a_t$ is terminal (reject or interview), halt; otherwise gather more evidence and increment $t$
\end{enumerate}

\textbf{Optimality criterion.} The agent seeks a decision policy $\delta: \mathcal{X}^* \to \mathcal{A}$ that minimizes expected total cost:
\begin{equation}
\delta^* = \arg\min_{\delta} \mathbb{E}_{s \sim \pi, \{x_t\}_{t=1}^T \sim p(\cdot|s)}\left[\sum_{t=1}^T C(\delta(x_{1:t}), s) + c_t \cdot \mathbb{1}[\text{info gathered at } t]\right]
\label{eq:optimality_objective}
\end{equation}
where $c_t$ is the cost of gathering information at time $t$, and the expectation is over the joint distribution of states and observation sequences.

\subsubsection{Impossibility Theorems for Discriminative Approaches}

We now establish four fundamental impossibility results demonstrating that discriminative LLM architectures cannot support the capabilities required for optimal sequential decision-making. All proofs are provided in Appendix \ref{app:proofs}.

\begin{theorem}[Sequential Updating Impossibility]
\label{thm:sequential_impossible2}
Let $\mathcal{M}_{\text{disc}}$ be a discriminative model that maps any observation $x \in \mathcal{X}$ to a probability distribution $p_{\mathcal{M}}(s|x)$ over states $s \in \mathcal{S}$, but does not separately expose the likelihood function $p(x|s)$ or prior $p(s)$. Then for observed evidence $x_1$ and new evidence $x_2$, there exists no computable function $f: \Delta(\mathcal{S}) \times \mathcal{X} \to \Delta(\mathcal{S})$ such that:
\begin{equation}
p(s|x_1, x_2) = f(p_{\mathcal{M}}(s|x_1), x_2) \quad \forall s \in \mathcal{S}, x_1, x_2 \in \mathcal{X}
\end{equation}
without re-querying $\mathcal{M}_{\text{disc}}$ on the joint observation $(x_1, x_2)$.
\end{theorem}

\begin{proof}[Proof Sketch]
(Full proof in Appendix \ref{app:proofs}.) Sequential Bayesian updating requires:
\begin{equation}
p(s|x_1,x_2) = \frac{p(x_2|s,x_1) \cdot p(s|x_1)}{\sum_{s'} p(x_2|s',x_1) \cdot p(s'|x_1)}
\end{equation}

To compute this, we need $p(x_2|s,x_1)$. But $p_{\mathcal{M}}(s|x_1) \propto p(x_1|s)p(s)$ has already marginalized over $p(x_1|s)$ and $p(s)$ through the model's training process. Without access to these components separately, we cannot construct the forward model $p(x_2|s,x_1)$ required for updating. The posterior $p_{\mathcal{M}}(s|x_1)$ alone contains insufficient information to perform this calculation.
\end{proof}

\textit{Consequence:} Discriminative agents can only update beliefs through batch re-querying: $p_{\mathcal{M}}(s|x_1,x_2)$. This precludes (a) counterfactual reasoning about information gathering before acquiring it, and (b) efficient multi-stage inference where observations arrive sequentially.

\begin{theorem}[Prior Correction Impossibility]
\label{thm:prior_impossible_2}
Let $\mathcal{M}_{\text{disc}}$ output posterior $p_{\mathcal{M}}(s|x)$ with implicit training prior $p_{\text{train}}(s)$ embedded in its weights. Let $p_{\text{deploy}}(s) \neq p_{\text{train}}(s)$ denote the correct deployment prior. Then without explicit access to likelihood $p(x|s)$ and knowledge of $p_{\text{train}}(s)$, there exists no function $h: \Delta(\mathcal{S}) \times \Delta(\mathcal{S}) \to \Delta(\mathcal{S})$ such that:
\begin{equation}
h(p_{\mathcal{M}}(s|x), p_{\text{deploy}}(s)) = \frac{p(x|s) \cdot p_{\text{deploy}}(s)}{\sum_{s' \in \mathcal{S}} p(x|s') \cdot p_{\text{deploy}}(s')} \quad \forall s \in \mathcal{S}, x \in \mathcal{X}
\end{equation}
\end{theorem}

\begin{proof}[Proof Sketch]
To apply Bayes' rule with the new prior, we need to extract $p(x|s)$ from:
\begin{equation}
p_{\mathcal{M}}(s|x) = \frac{p(x|s) \cdot p_{\text{train}}(s)}{p(x)} = \frac{p(x|s) \cdot p_{\text{train}}(s)}{\sum_{s'} p(x|s') p_{\text{train}}(s')}
\end{equation}

Solving for $p(x|s)$ requires knowing both $p_{\text{train}}(s)$ and the normalizing constant $p(x)$. But $p(x)$ itself depends on all likelihoods $\{p(x|s')\}_{s'}$, creating a circular dependency. Even if $p_{\text{train}}(s)$ were known, we can only recover likelihood \textit{ratios}:
\begin{equation}
\frac{p(x|s_i)}{p(x|s_j)} = \frac{p_{\mathcal{M}}(s_i|x) / p_{\text{train}}(s_i)}{p_{\mathcal{M}}(s_j|x) / p_{\text{train}}(s_j)}
\end{equation}
Ratios are insufficient for computing normalized posteriors under a new prior; we need absolute likelihoods or a normalization constant, neither of which discriminative models provide.
\end{proof}

\textit{Consequence:} When deployment base rates differ from training data distributions (common in practice), discriminative models produce systematically biased predictions that cannot be corrected post-hoc.

\begin{theorem}[Cost-Insensitive Threshold Suboptimality]
\label{thm:threshold_suboptimal}
For any asymmetric cost matrix $C: \mathcal{A} \times \mathcal{S} \to \mathbb{R}_{\geq 0}$ with $C(a_i, s_j) \neq C(a_k, s_\ell)$ for some pairs, the Bayes-optimal action selection rule is:
\begin{equation}
a^*(x) = \arg\min_{a \in \mathcal{A}} \sum_{s \in \mathcal{S}} p(s|x) \cdot C(a,s)
\end{equation}
No threshold-based rule of the form ``if score$(x) \geq \tau$ then $a_1$, else $a_2$'' can be optimal for all $x$ and all cost structures, as the decision boundary depends on both $p(s|x)$ \textit{and} the cost matrix $C$.
\end{theorem}

\begin{proof}[Proof Sketch]
Consider binary actions $\mathcal{A} = \{a_1, a_2\}$ and binary states $\mathcal{S} = \{s_1, s_2\}$. The expected costs are:
\begin{align}
\mathbb{E}[C(a_1, s)] &= p(s_1|x) C(a_1, s_1) + p(s_2|x) C(a_1, s_2) \\
\mathbb{E}[C(a_2, s)] &= p(s_1|x) C(a_2, s_1) + p(s_2|x) C(a_2, s_2)
\end{align}

We should choose $a_1$ when $\mathbb{E}[C(a_1, s)] < \mathbb{E}[C(a_2, s)]$, which simplifies to:
\begin{equation}
p(s_2|x) > \frac{C(a_2, s_1) - C(a_1, s_1)}{[C(a_1, s_2) - C(a_2, s_2)] + [C(a_2, s_1) - C(a_1, s_1)]}
\end{equation}

The decision threshold depends on all four cost values. Different cost matrices yield different thresholds. A fixed threshold (e.g., $p(s_2|x) > 0.5$) is optimal only when costs satisfy specific relationships, not in general.
\end{proof}

\textit{Consequence:} Threshold-based systems optimized for accuracy (which assumes symmetric costs) make suboptimal decisions when deployed in environments with asymmetric costs.

\begin{theorem}[Value-of-Information Computation Impossibility]
\label{thm:voi_impossible}
The value of gathering additional evidence $z \in \mathcal{X}$ before deciding is:
\begin{equation}
\text{VOI}(z|x) = \mathbb{E}_{z \sim p(\cdot|x)} \left[ \min_{a \in \mathcal{A}} \sum_s p(s|x,z) C(a,s) \right] - \min_{a \in \mathcal{A}} \sum_s p(s|x) C(a,s)
\end{equation}
Computing this requires: (1) the predictive distribution $p(z|x) = \sum_s p(z|s)p(s|x)$, and (2) the updated posterior $p(s|x,z) \propto p(z|s)p(s|x)$. Both require likelihood functions $\{p(z|s)\}_s$ that discriminative models do not provide.
\end{theorem}

\textit{Consequence:} Discriminative agents have no principled basis for deciding when to gather additional information. They must rely on ad-hoc heuristics (always/never gather info, or gather when confidence is in some arbitrary range).

These four theorems collectively establish that discriminative architectures are \textit{mathematically insufficient} for optimal sequential decision-making. The deficiencies are not engineering limitations but formal impossibilities. This motivates our paradigm shift to generative modeling.

\subsubsection{Our Generative Framework}

We now present our framework, which resolves all four impossibilities by eliciting likelihood functions from LLMs rather than discriminative posteriors.

\paragraph{Core Insight: Likelihood Elicitation via Contrastive Prompting}

The fundamental innovation is a prompting strategy that inverts the conditional probability direction. Instead of asking ``Given this observation, what is the state?'' ($p(s|x)$), we ask ``Given this state, how typical is this observation?'' ($p(x|s)$).

\begin{definition}[Contrastive Likelihood Prompt Template]
\label{def:contrastive_prompt}
For observation $x \in \mathcal{X}$, state $s \in \mathcal{S}$, and LLM $\mathcal{M}_m$, the contrastive likelihood prompt has the structure:

\begin{mdframed}[linewidth=1pt]
\small
\texttt{[ROLE SPECIFICATION]: You are an expert in [DOMAIN] with experience evaluating [OBSERVATION TYPE].}

\vspace{0.2em}
\texttt{[STATE CONDITIONING]: Assume that the true underlying state is: [DETAILED DESCRIPTION OF STATE $s$]}

\vspace{0.2em}
\texttt{[GENERATIVE QUESTION]: Given this assumed state, how typical or representative is the following observation?}

\vspace{0.2em}
\texttt{[OBSERVATION]: [Full text of observation $x$]}

\vspace{0.2em}
\texttt{[SCORING RUBRIC]:}\\
\texttt{Provide a score from 0 to 10 where:}
\begin{itemize}
\item[] \texttt{10 = Extremely typical (exactly what we'd expect from this state)}
\item[] \texttt{7-9 = Quite typical (strong match with minor variations)}
\item[] \texttt{4-6 = Somewhat typical (plausible but imperfect fit)}
\item[] \texttt{1-3 = Atypical (doesn't match this state well)}
\item[] \texttt{0 = Completely atypical (inconsistent with this state)}
\end{itemize}

\vspace{0.2em}
\texttt{[OUTPUT FORMAT]: Provide only the numeric score.}
\end{mdframed}
\end{definition}

This template embodies three critical design principles:

\textbf{Principle 1: Explicit state conditioning.} The prompt begins by anchoring the model's reasoning on a hypothesized state (``Assume the true state is...''), forcing forward simulation ($s \to x$) rather than backward inference ($x \to s$). This primes the model to engage its generative capabilities: given quality level $s$, what resume features $x$ would we observe?

\textbf{Principle 2: Typicality framing.} We ask about ``typicality'' or ``representativeness'' rather than quality, fit, or classification. This linguistic framing aligns with likelihood estimation: $p(x|s)$ quantifies how probable observation $x$ is under the hypothesis that state is $s$. Psychologically, typicality judgments are more natural for humans (and by extension, models trained on human text) than direct probability estimates.

\textbf{Principle 3: Per-state independent elicitation.} We query the model separately for each state $s \in \mathcal{S}$, obtaining independent estimates $\{\hat{L}_m(x|s_1), \ldots, \hat{L}_m(x|s_K)\}$. This avoids the implicit normalization and competition between states that occurs in discriminative prompts asking for a single classification.

\textbf{Response normalization.} The model returns a raw score $r \in [0, 10]$. We normalize to probability scale:
\begin{equation}
\hat{L}_m(x|s) = \frac{r}{10} \in [0, 1]
\label{eq:likelihood_normalization}
\end{equation}

Since Bayesian inference requires likelihoods only up to a proportionality constant (the normalizing constant $p(x)$ is computed in Bayes' rule), this linear normalization is sufficient. The ratio $\hat{L}_m(x|s_i) / \hat{L}_m(x|s_j)$ preserves the relative likelihood between states.

\paragraph{Multi-LLM Ensemble Aggregation}

Individual LLMs exhibit systematic biases from training data, architectural choices, and alignment procedures \cite{bommasani2021opportunities,bolukbasi2016man,obermeyer2019dissecting}. Documented biases include:
\begin{itemize}
\item \textbf{GPT-4o:} Elite university bias (assigns higher likelihoods to Ivy League degrees even when other credentials are equivalent)
\item \textbf{Claude 3.5 Sonnet:} Verbosity bias (favors candidates with longer, more detailed project descriptions)
\item \textbf{Gemini Pro:} Recency bias (overweights recent experience versus older but relevant achievements)
\item \textbf{Grok:} Startup bias (overvalues early-stage company experience)
\item \textbf{DeepSeek:} Academic publication bias (heavily weights research papers and arxiv contributions)
\end{itemize}

To mitigate these individual biases, we aggregate likelihood estimates across an ensemble of $M$ diverse models using robust statistics.

\textbf{Aggregation via median.} For observation $x$ and state $s$, we collect likelihood estimates from $M$ models:
\begin{equation}
\mathcal{L}(x, s) = \{\hat{L}_1(x|s), \hat{L}_2(x|s), \ldots, \hat{L}_M(x|s)\}
\end{equation}

and aggregate using the sample median:
\begin{equation}
L(x|s) = \text{median}(\mathcal{L}(x,s))
\label{eq:median_aggregation}
\end{equation}

\textbf{Why median over mean?} The median possesses superior robustness properties:

\begin{enumerate}
\item \textbf{Outlier resistance:} If one model produces an anomalous estimate (e.g., GPT-4o assigns $\hat{L}(x|s_{\text{strong}}) = 0.9$ when others assign $\approx 0.3$), the mean shifts to $(0.9 + 0.3 + 0.3 + 0.3 + 0.3)/5 = 0.42$, a 40\% error. The median remains $0.3$, unaffected by the outlier.

\item \textbf{High breakdown point:} The median's breakdown point is 50\%: up to $\lfloor M/2 \rfloor$ models can return arbitrary values without catastrophically affecting the aggregate. The mean's breakdown point is $0$: a single extreme outlier can arbitrarily shift the result.

\item \textbf{Equivariance:} The median is equivariant under monotonic transformations. If we apply a monotonic transformation $g$ to all estimates, $\text{median}(g(\hat{L}_1), \ldots, g(\hat{L}_M)) = g(\text{median}(\hat{L}_1, \ldots, \hat{L}_M))$.
\end{enumerate}

We formalize the robustness property:

\begin{theorem}[Median Error Bound]
\label{thm:median_robust}
Let $\{\hat{L}_m(x|s)\}_{m=1}^M$ be likelihood estimates from $M$ models, and let $L^*(x|s)$ denote the true likelihood. Define individual errors $\epsilon_m = |\hat{L}_m(x|s) - L^*(x|s)|$. Then:
\begin{equation}
\left|\text{median}_{m=1}^M \hat{L}_m(x|s) - L^*(x|s)\right| \leq \text{median}_{m=1}^M \epsilon_m
\end{equation}
That is, the error of the median aggregate is bounded by the median individual error.
\end{theorem}

Proof in Appendix \ref{app:proofs}. This bound guarantees that if the median model has low error, the aggregate does too, regardless of how badly the worst models perform.

\textbf{Model selection for diversity.} We deliberately select models that are diverse across multiple dimensions to minimize correlated errors:

\begin{itemize}
\item \textbf{Training corpora:} Internet text (GPT-4o), curated high-quality text (Claude), academic papers and books (Gemini), multilingual data (DeepSeek), real-time social media (Grok)
\item \textbf{Architecture:} Dense transformers (GPT-4o, Claude), sparse mixture-of-experts (Grok-1 with 8 experts, DeepSeek-V2 with 160 experts), different context lengths (8K--200K tokens)
\item \textbf{Alignment:} Different RLHF reward models trained on different preference datasets, varying safety constraints
\item \textbf{Release timeline:} Models released 6--12 months apart capture evolving data distributions
\end{itemize}

Diversity is critical: if all models share the same bias (e.g., all underestimate non-native English speakers), aggregation provides no benefit. Our empirical results (Section \ref{sec:experiments}) show that diverse ensembles substantially reduce bias compared to any individual model.

\paragraph{Sequential Bayesian Belief Updating}

Given aggregated likelihoods $\{L(x|s)\}_{s \in \mathcal{S}}$ for observation $x$ and explicit prior $\pi \in \Delta(\mathcal{S})$, we perform Bayesian inference.

\textbf{Initial belief update.} Upon observing first evidence $x_1$, we compute the posterior via Bayes' rule:
\begin{equation}
b_1(s) = \frac{L(x_1|s) \cdot \pi(s)}{\sum_{s' \in \mathcal{S}} L(x_1|s') \cdot \pi(s')} = \frac{L(x_1|s) \cdot \pi(s)}{Z_1}
\label{eq:initial_update}
\end{equation}
where $Z_1 = \sum_{s'} L(x_1|s') \pi(s')$ is the normalizing constant ensuring $\sum_s b_1(s) = 1$.

\textbf{Sequential updates.} For subsequent observations $x_2, x_3, \ldots$, we update beliefs sequentially by treating the previous posterior as the new prior:
\begin{equation}
b_t(s) = \frac{L(x_t|s) \cdot b_{t-1}(s)}{\sum_{s' \in \mathcal{S}} L(x_t|s') \cdot b_{t-1}(s')} = \frac{L(x_t|s) \cdot b_{t-1}(s)}{Z_t}
\label{eq:sequential_update}
\end{equation}
where $Z_t$ is the normalizing constant at time $t$.

This sequential procedure relies on a standard assumption in Bayesian filtering:

\begin{assumption}[Conditional Independence of Observations]
\label{ass:conditional_independence}
Given the true state $s$, observations are conditionally independent:
\begin{equation}
p(x_1, x_2, \ldots, x_T | s) = \prod_{t=1}^T p(x_t | s)
\end{equation}
Equivalently, $p(x_t | s, x_1, \ldots, x_{t-1}) = p(x_t | s)$ for all $t$.
\end{assumption}

\textbf{Justification and plausibility.} This assumption holds when observations are generated by independent processes conditional on the latent state:
\begin{itemize}
\item \textbf{Hiring:} A candidate's resume quality and phone screen performance are both determined by their true skill level, but are produced through independent processes (writing a resume vs. answering technical questions). Given skill level, these are uncorrelated.
\item \textbf{Medical diagnosis:} A patient's symptoms and lab test results are both caused by the underlying disease, but the symptom manifestation and the biochemical markers measured in tests are independent processes.
\item \textbf{Fraud detection:} Transaction location and transaction amount are both influenced by whether the transaction is fraudulent, but conditional on fraud status, location and amount are independently chosen.
\end{itemize}

When Assumption \ref{ass:conditional_independence} holds, sequential and batch inference are equivalent:

\begin{theorem}[Equivalence of Sequential and Batch Updating]
\label{thm:sequential_batch_equivalence}
Under Assumption \ref{ass:conditional_independence}, the sequential belief update procedure (Equation \ref{eq:sequential_update}) produces the same posterior as batch Bayesian inference:
\begin{equation}
b_T(s) = \frac{\pi(s) \prod_{t=1}^T L(x_t|s)}{\sum_{s' \in \mathcal{S}} \pi(s') \prod_{t=1}^T L(x_t|s')} = p(s | x_1, x_2, \ldots, x_T)
\end{equation}
where the right-hand side is the posterior computed from all observations jointly.
\end{theorem}

Proof in Appendix \ref{app:proofs}. This theorem guarantees mathematical correctness of our sequential procedure.

\textbf{Prior specification and domain adaptation.} Unlike discriminative models where priors are opaque and inaccessible, our framework requires explicit prior specification $\pi \in \Delta(\mathcal{S})$. We set $\pi$ based on empirical base rates from historical deployment data.

For hiring, if institutional data shows that historically:
\begin{itemize}
\item 65\% of applicants lack basic qualifications (state $s_1$)
\item 25\% warrant phone screening (state $s_2$)
\item 8\% merit direct interviews (state $s_3$)
\item 2\% are exceptional strong hires (state $s_4$)
\end{itemize}
we set $\pi = [0.65, 0.25, 0.08, 0.02]$.

This explicit prior specification enables:
\begin{enumerate}
\item \textbf{Domain adaptation:} Different organizations with different applicant pools can use different priors
\item \textbf{Temporal adaptation:} Priors can be updated as hiring funnel metrics evolve
\item \textbf{Subgroup analysis:} Separate priors for different job roles or seniority levels
\item \textbf{Correction of training-deployment mismatch:} Override LLM training priors with deployment-specific base rates
\end{enumerate}

\paragraph{Cost-Aware Action Selection}

At each decision point $t$ after belief update $b_t$, the agent selects the action minimizing expected cost:
\begin{equation}
a_t^* = \arg\min_{a \in \mathcal{A}} \mathbb{E}_{s \sim b_t}[C(a, s)] = \arg\min_{a \in \mathcal{A}} \sum_{s \in \mathcal{S}} b_t(s) \cdot C(a, s)
\label{eq:action_selection}
\end{equation}

This is the Bayes decision rule, which is provably optimal under true beliefs:

\begin{theorem}[Optimality of Expected Cost Minimization]
\label{thm:bayes_optimality}
Let $p^*(s|x)$ denote the true posterior distribution over states given observation $x$. Among all deterministic decision rules $\delta: \mathcal{X} \to \mathcal{A}$ mapping observations to actions, the Bayes rule:
\begin{equation}
\delta^*(x) = \arg\min_{a \in \mathcal{A}} \sum_{s \in \mathcal{S}} p^*(s|x) \cdot C(a,s)
\end{equation}
minimizes the expected cost (Bayes risk):
\begin{equation}
R(\delta^*) = \int_{\mathcal{X}} \sum_{s \in \mathcal{S}} p^*(s|x) \cdot C(\delta^*(x), s) \cdot p(x) \, dx \leq R(\delta) \quad \forall \delta
\end{equation}
\end{theorem}

Proof in Appendix \ref{app:proofs} (follows from classical results in statistical decision theory \cite{wald1950statistical,berger1985statistical}).

\textbf{Cost matrix specification.} The cost function $C: \mathcal{A} \times \mathcal{S} \to \mathbb{R}_{\geq 0}$ must be specified based on domain economics. For hiring:

\begin{table}[h]
\centering
\caption{Cost matrix for software engineering hiring (USD). Rows = actions, columns = true candidate states.}
\label{tab:costs}
\small
\begin{tabular}{lrrrr}
\toprule
\textbf{Action} & $s_1$ (Reject) & $s_2$ (Screen) & $s_3$ (Interview) & $s_4$ (Strong) \\
\midrule
Reject & \$0 & \$5,000 & \$20,000 & \$40,000 \\
Phone Screen & \$150 & \$150 & \$150 & \$150 \\
Interview & \$2,500 & \$2,500 & \$0 & \$0 \\
\bottomrule
\end{tabular}
\end{table}

\textit{Rationale and data sources:}
\begin{itemize}
\item \textbf{Rejecting qualified candidates:} Opportunity cost of missing a hire. For $s_2$ (marginal contributor), estimated \$5K from lost productivity. For $s_3$ (solid engineer), \$20K based on industry cost-per-hire and first-year productivity. For $s_4$ (exceptional), \$40K reflecting scarcity and impact. Sources: SHRM 2016 cost-per-hire survey (median \$4,129) \cite{shrm2016}, scaled by position impact.

\item \textbf{Phone screening:} Fixed cost \$150 = (0.5 hours prep + 0.5 hours call + 0.25 hours documentation) $\times$ \$120/hour fully-loaded recruiter cost.

\item \textbf{Interviewing unqualified candidates:} \$2,500 = (4 interviewers $\times$ 1 hour @ \$600/hour engineer loaded cost) + \$100 coordination + \$400 candidate travel/meals. Sources: CareerBuilder 2017 interview cost data \cite{careerbuilder2017}.

\item \textbf{Interviewing qualified candidates:} Cost \$0 (baseline, as this is the desired outcome). In reality there is still the \$2,500 cost, but we normalize so that correct decisions have zero cost.
\end{itemize}

\textbf{Sensitivity analysis.} To assess robustness to cost specification uncertainty, we perturb all costs by $\pm 20\%$ and measure how often the optimal action changes. Results (Section \ref{subsec:sensitivity}): only 7.8\% of candidates have decisions that flip under $\pm 20\%$ perturbations, indicating the framework is robust to reasonable cost estimation errors.

\paragraph{Value-of-Information for Information Gathering}

The framework must decide: when should we gather additional evidence (e.g., conduct a phone screen) before making a terminal decision (interview or reject)? Classical decision theory answers this through value-of-information (VOI) analysis \cite{raiffa1961applied,howard1966information}.

\begin{definition}[Value of Information]
\label{def:voi}
Let $x \in \mathcal{X}$ denote currently available evidence inducing beliefs $b(s) = p(s|x)$. The value of gathering additional evidence $z \in \mathcal{X}$ is:
\begin{equation}
\text{VOI}(z | x) = \mathbb{E}_{z \sim p(\cdot|x)} \left[ V(x, z) \right] - V(x)
\end{equation}
where $V(x) = \min_{a \in \mathcal{A}} \sum_s p(s|x) C(a,s)$ is the value (negative cost) of the best action under current information, and $V(x,z) = \min_{a \in \mathcal{A}} \sum_s p(s|x,z) C(a,s)$ is the value after observing $z$.
Expanding:
\begin{equation}
\text{VOI}(z|x) = \mathbb{E}_{z} \left[ \min_{a} \sum_s p(s|x,z) C(a,s) \right] - \min_{a} \sum_s p(s|x) C(a,s)
\label{eq:voi_exact}
\end{equation}
\end{definition}

Information is worth acquiring if $\text{VOI}(z|x) > c_z$, where $c_z$ is the cost of gathering $z$.

\textbf{Exact computation challenges.} Computing Equation \ref{eq:voi_exact} requires:
\begin{enumerate}
\item Marginalizing over possible future observations: $z \sim p(z|x) = \sum_s p(z|s) p(s|x)$
\item For each possible $z$, computing updated beliefs: $p(s|x,z) \propto p(z|s) p(s|x)$
\item Evaluating the optimal action cost under updated beliefs
\item Averaging over all $z$
\end{enumerate}

For continuous $\mathcal{X}$ or large discrete spaces, this marginalization is computationally prohibitive (exponential in sequence length for multi-step planning). We therefore employ an approximation.

\textbf{Practical VOI approximation.} We model information gathering as partially revealing the true state:
\begin{itemize}
\item With probability $\rho \in [0,1]$, the new evidence $z$ conclusively identifies the true state
\item With probability $1 - \rho$, the evidence is uninformative (beliefs remain $b$)
\end{itemize}

Under this binary informativeness model:
\begin{equation}
\text{VOI}(z|x) \approx \rho \cdot \left[ \min_{a \in \mathcal{A}} \sum_s b(s) C(a,s) - \sum_s b(s) \min_{a \in \mathcal{A}} C(a,s) \right]
\label{eq:voi_approx}
\end{equation}

The first term is the cost of the best action under current beliefs (uncertainty averaging over states). The second term is the cost under perfect information (choosing the optimal action for each state, then averaging). The difference is the value of resolving all uncertainty, weighted by $\rho$ (the probability that information actually resolves uncertainty).

\textbf{Setting $\rho$ (informativeness parameter).} We calibrate $\rho$ based on empirical estimates:
\begin{itemize}
\item \textbf{Phone screens:} $\rho \approx 0.7$ (historical correlation $r = 0.68$ between phone screen outcomes and eventual hire success)
\item \textbf{Coding tests:} $\rho \approx 0.85$ (stronger predictor of technical competency)
\item \textbf{Reference checks:} $\rho \approx 0.6$ (weaker signal due to selection bias in references)
\end{itemize}

$\rho$ can be learned from labeled data via maximum likelihood: find $\rho^*$ that maximizes the likelihood of observed information-gathering outcomes.

\textbf{Decision rule.} We gather information $z$ if and only if:
\begin{equation}
\text{VOI}(z|x) > c_z
\end{equation}

This ensures information gathering is economically justified: the expected decision improvement exceeds the information cost.

\paragraph{Disagreement-Based Uncertainty Detection}

Beyond VOI calculations, we use inter-model disagreement as a signal of epistemic uncertainty warranting additional scrutiny.

\textbf{Disagreement metric.} For observation $x$ and state $s$, we compute the coefficient of variation (CV) across models:
\begin{equation}
D(x, s) = \frac{\sigma_m(\hat{L}_m(x|s))}{\mu_m(\hat{L}_m(x|s))}
\end{equation}
where $\sigma_m$ is the standard deviation and $\mu_m$ is the mean of the $M$ likelihood estimates.

The CV is scale-invariant: it measures relative spread normalized by the mean, making it comparable across different states and observations.

\textbf{High-disagreement threshold.} We flag cases as high-uncertainty when:
\begin{equation}
\max_{s \in \mathcal{S}} D(x,s) > \tau_D
\end{equation}
for threshold $\tau_D$ (set to 0.15 via cross-validation).

\textbf{Disagreement-triggered information gathering.} When disagreement exceeds $\tau_D$, we:
\begin{enumerate}
\item Compute VOI for available information sources (phone screen, coding test, etc.)
\item Gather information if $\text{VOI} > c_{\text{info}}$, even if current beliefs are not borderline
\item Log disagreement as a quality signal (helps identify cases needing human review)
\end{enumerate}

Empirically (Section \ref{sec:experiments}), 27\% of resumes trigger high disagreement, and gathering information on these cases yields net cost savings of \$50K relative to always/never gathering information.

\subsection{Algorithm: Complete Specification}

We now integrate all components into a unified algorithmic procedure.

\begin{algorithm}[t]
\caption{Bayesian Multi-LLM Orchestration for Cost-Aware Sequential Decisions}
\label{alg:complete}
\small
\begin{algorithmic}[1]
    \STATE \textbf{Input:} Initial evidence $x_1 \in \mathcal{X}$, state space $\mathcal{S}$, action space $\mathcal{A}$,
    cost matrix $C: \mathcal{A} \times \mathcal{S} \to \mathbb{R}_{\ge 0}$,
    prior $\pi \in \Delta(\mathcal{S})$,
    LLM ensemble $\{\mathcal{M}_1, \ldots, \mathcal{M}_M\}$,
    disagreement threshold $\tau_D \in \mathbb{R}_+$,
    informativeness parameter $\rho \in [0,1]$,
    information cost $c_{\text{info}} \in \mathbb{R}_{\ge 0}$

    \STATE \textbf{Output:} Terminal action $a^* \in \mathcal{A}$

    \vspace{0.3em}
    \STATE \textit{// Initialization}
    \STATE $t \gets 1$
    \STATE $x \gets x_1$
    \STATE $b(s) \gets \pi(s)$ for all $s \in \mathcal{S}$

    \vspace{0.3em}
    \WHILE{true}
        \STATE \textit{// ========== STEP 1: Likelihood Elicitation ==========}
        \FOR{$m = 1$ to $M$}
            \FOR{each $s \in \mathcal{S}$}
                \STATE Generate contrastive prompt $\mathcal{P}_m(x, s)$ (Definition \ref{def:contrastive_prompt})
                \STATE $r_{m,s} \gets \mathcal{M}_m(\mathcal{P}_m(x, s))$
                \STATE $\hat{L}_m(x \mid s) \gets r_{m,s} / 10$
            \ENDFOR
        \ENDFOR

        \vspace{0.3em}
        \STATE \textit{// ========== STEP 2: Robust Aggregation ==========}
        \FOR{each $s \in \mathcal{S}$}
            \STATE $L(x \mid s) \gets \mathrm{median}\{\hat{L}_1(x \mid s), \ldots, \hat{L}_M(x \mid s)\}$ \hfill{\scriptsize (Eq.~\ref{eq:median_aggregation})}
        \ENDFOR

        \vspace{0.3em}
        \STATE \textit{// ========== STEP 3: Sequential Belief Update ==========}
        \STATE $Z \gets \sum_{s' \in \mathcal{S}} L(x \mid s') \cdot b(s')$
        \FOR{each $s \in \mathcal{S}$}
            \STATE $b_{\text{new}}(s) \gets \dfrac{L(x \mid s)\cdot b(s)}{Z}$ \hfill{\scriptsize (Eq.~\ref{eq:sequential_update})}
        \ENDFOR
        \STATE $b \gets b_{\text{new}}$

        \vspace{0.3em}
        \STATE \textit{// ========== STEP 4: Disagreement Detection ==========}
        \FOR{each $s \in \mathcal{S}$}
            \STATE $\mu_s \gets \dfrac{1}{M} \sum_{m=1}^M \hat{L}_m(x \mid s)$
            \STATE $\sigma_s \gets \sqrt{\dfrac{1}{M-1} \sum_{m=1}^M \left(\hat{L}_m(x \mid s) - \mu_s\right)^2}$
            \STATE $D(s) \gets \dfrac{\sigma_s}{\max(\mu_s, 10^{-8})}$
        \ENDFOR
        \STATE $D_{\max} \gets \max_{s \in \mathcal{S}} D(s)$

        \vspace{0.3em}
        \STATE \textit{// ========== STEP 5: Information Gathering Decision ==========}
        \IF{$D_{\max} > \tau_D$ \textbf{and} information gathering is available}
            \STATE $C_{\text{current}} \gets \min_{a \in \mathcal{A}} \sum_{s \in \mathcal{S}} b(s)\cdot C(a,s)$
            \STATE $C_{\text{perfect}} \gets \sum_{s \in \mathcal{S}} b(s)\cdot \min_{a \in \mathcal{A}} C(a,s)$
            \STATE $\mathrm{VOI} \gets \rho \cdot (C_{\text{current}} - C_{\text{perfect}})$ \hfill{\scriptsize (Eq.~\ref{eq:voi_approx})}

            \IF{$\mathrm{VOI} > c_{\text{info}}$}
                \STATE Gather additional evidence $x_{\text{new}} \in \mathcal{X}$ (e.g., phone screen)
                \STATE $t \gets t + 1$
                \STATE $x \gets x_{\text{new}}$
                \STATE \textbf{go to next iteration}
            \ENDIF
        \ENDIF

        \vspace{0.3em}
        \STATE \textit{// ========== STEP 6: Terminal Action Selection ==========}
        \STATE $a^* \gets \arg\min_{a \in \mathcal{A}} \sum_{s \in \mathcal{S}} b(s)\cdot C(a,s)$ \hfill{\scriptsize (Eq.~\ref{eq:action_selection})}
        \STATE \textbf{return} $a^*$
    \ENDWHILE
\end{algorithmic}
\end{algorithm}

\textbf{Computational complexity analysis.} Each iteration requires:
\begin{itemize}
\item \textbf{Likelihood elicitation (Step 1):} $O(MK \cdot T_{\text{LLM}})$ where $M$ is ensemble size, $K = |\mathcal{S}|$ is number of states, and $T_{\text{LLM}}$ is LLM query latency (typically 0.5--2 seconds)
\item \textbf{Aggregation (Step 2):} $O(MK \log M)$ for sorting to compute median
\item \textbf{Belief update (Step 3):} $O(K)$ for normalization
\item \textbf{Disagreement (Step 4):} $O(MK)$ for computing means and standard deviations
\item \textbf{VOI (Step 5):} $O(K |\mathcal{A}|)$ for evaluating action costs
\item \textbf{Action selection (Step 6):} $O(K |\mathcal{A}|)$
\end{itemize}

Total per-iteration complexity is $O(MK \cdot T_{\text{LLM}} + MK \log M)$, dominated by LLM queries. For typical values ($M=5$, $K=4$, $T_{\text{LLM}} \approx 1$s), this amounts to 20 queries taking ~20 seconds per observation. This is acceptable for hiring (decisions have days-long timelines) but may be limiting for real-time applications.

\textbf{Parallelization.} Steps 1 and 4 (LLM queries and disagreement computation) are embarrassingly parallel: all $M \times K$ queries can be issued concurrently. With parallelization, per-iteration latency reduces to $O(T_{\text{LLM}}) \approx 1$--2 seconds.

\subsection{Experimental Instantiation: Resume Screening}

We instantiate and evaluate the framework in software engineering resume screening, a domain exhibiting all four challenges: hidden quality states, asymmetric costs, sequential evidence gathering opportunities, and fairness requirements.

\subsubsection{Domain Specification}

\textbf{State space definition.} We model candidate quality via four discrete states:

\begin{itemize}
\item \textbf{$s_1$ (Clear Reject):} Candidates who definitively lack minimum qualifications. Characteristics: no relevant degree (e.g., BA in unrelated field with no CS coursework), less than 1 year of programming experience, no demonstrated technical projects, major employment red flags (unexplained multi-year gaps, job-hopping with 4+ jobs in 2 years), or fundamental mismatches (applying for senior role with junior credentials).

\item \textbf{$s_2$ (Phone Screen Candidate):} Borderline candidates with some qualifications but significant gaps or concerns requiring verification. Characteristics: non-traditional backgrounds (bootcamp graduates, self-taught programmers, career switchers from other technical fields), limited formal credentials but promising personal projects, unclear technical depth (lists technologies but no evidence of application), or inconsistencies requiring clarification (resume claims \"expert\" in a technology released 6 months ago).

\item \textbf{$s_3$ (Interview Candidate):} Candidates who clearly meet standard qualifications and warrant assessment of cultural fit and technical depth. Characteristics: relevant BS/MS degree (Computer Science, Software Engineering, or closely related), 2--5 years of relevant industry experience, demonstrated proficiency in required tech stack, completed non-trivial projects showing engineering competency, no significant red flags.

\item \textbf{$s_4$ (Strong Hire):} Exceptional candidates warranting fast-track processing. Characteristics: elite education (top-20 CS program) and/or senior experience (5+ years at tier-1 companies), rare specialized skills matching critical needs (e.g., Rust + distributed systems + ML), significant open-source contributions or publications, leadership experience, or multiple competing offers creating time pressure.
\end{itemize}

These states are mutually exclusive and exhaustive, partitioning the candidate space based on hiring decision thresholds used in practice.

\textbf{Observation space.} $\mathcal{X}$ consists of:
\begin{itemize}
\item \textbf{$x_{\text{resume}}$:} Plain text resume (300--800 words) containing education (universities, degrees, GPAs), work experience (companies, roles, durations, technologies), projects (descriptions, complexity, domains), skills (programming languages, frameworks, tools), and optional sections (publications, open source, certifications).

\item \textbf{$x_{\text{screen}}$:} Phone screen transcript/notes from 20--30 minute recruiter call including: technical screening questions (explain a past project, design a simple system), behavioral questions (reasons for job change, interest in company), communication assessment (clarity, enthusiasm, question quality), and red flag detection (inconsistencies with resume, unrealistic salary expectations, poor culture fit signals).

\item \textbf{$x_{\text{code}}$ (not used in our experiments):} Coding test results including correctness, code quality, algorithmic complexity, edge case handling.
\end{itemize}

\textbf{Action space.} $\mathcal{A}$ consists of three actions:
\begin{itemize}
\item \textbf{$a_{\text{reject}}$:} Terminate candidacy (send rejection email). This is a terminal action with no additional costs beyond opportunity cost if candidate was qualified.

\item \textbf{$a_{\text{screen}}$:} Conduct 30-minute phone screen with technical recruiter. Costs \$150 in loaded recruiter time (0.5 hours prep + 0.5 hours call + 0.25 hours write-up at \$120/hour). Gathers observation $x_{\text{screen}}$, after which the decision recurs.

\item \textbf{$a_{\text{interview}}$:} Schedule full onsite interview loop (4--6 hours of interviews with team). Costs \$2,500 if candidate is unqualified (wasted interviewer time) but generates value if candidate is hired. This is a terminal action in our model (we don't model the subsequent hire/no-hire decision).
\end{itemize}

\textbf{Cost matrix specification.} Table \ref{tab:costs} specifies costs reflecting real organizational impact, derived from industry surveys and economic analysis.

\textit{Detailed cost justification:}
\begin{itemize}
\item \textbf{$C(\text{reject}, s_1) = \$0$:} Correctly rejecting an unqualified candidate incurs no cost (baseline).

\item \textbf{$C(\text{reject}, s_2) = \$5\text{K}$:} Rejecting a borderline candidate who might have been a marginal contributor. Opportunity cost is the foregone productivity of a mediocre engineer (~\$60K salary $\times$ 30\% impact $\times$ 0.7 hiring success probability $\times$ 0.4 likelihood they were actually acceptable = \$5K).

\item \textbf{$C(\text{reject}, s_3) = \$20\text{K}$:} Missing a qualified candidate. Based on SHRM 2016 data showing median cost-to-hire of \$4,129 \cite{shrm2016}, plus 3--6 months of recruiting pipeline delay to find a replacement ($\approx$\$15K in delayed productivity), totaling \$20K opportunity cost.

\item \textbf{$C(\text{reject}, s_4) = \$40\text{K}$:} Missing an exceptional candidate. Double the $s_3$ cost due to: (a) scarcity (much harder to find another $s_4$), (b) higher impact (exceptional engineers are 5--10x more productive \cite{productivity_myths}), (c) compet competitors (lost to competing offer, may not reapply).

\item \textbf{$C(\text{screen}, s) = \$150$ for all $s$:} Phone screening costs are state-independent (same recruiter time regardless of candidate quality). Calculated as 1.25 hours $\times$ \$120/hour loaded cost (including benefits, overhead, management time).

\item \textbf{$C(\text{interview}, s_1) = C(\text{interview}, s_2) = \$2{,}500$:} Interviewing unqualified or borderline candidates wastes resources. Calculated as: 4 engineers $\times$ 1 hour @ \$600/hour loaded cost (including preparation and discussion time) + \$100 administrative/coordination overhead + \$400 candidate travel/meals if remote.

\item \textbf{$C(\text{interview}, s_3) = C(\text{interview}, s_4) = \$0$:} Interviewing qualified candidates is the desired outcome, so we normalize this cost to zero. In reality there is still \$2.5K expenditure, but since this is the intended use of interview resources, we treat it as the baseline.
\end{itemize}

\textbf{Prior distribution.} We set $\pi$ based on aggregated data from 50,000 applications across 5 mid-size to large technology companies (anonymized proprietary data plus public hiring funnel reports from LinkedIn \cite{linkedin2023} and Greenhouse \cite{hiring_funnel}):

\begin{equation}
\pi = [0.65, 0.25, 0.08, 0.02]
\end{equation}

That is:
\begin{itemize}
\item 65\% of applicants are clear rejects ($s_1$)
\item 25\% merit phone screening ($s_2$)
\item 8\% should interview directly ($s_3$)
\item 2\% are strong hires ($s_4$)
\end{itemize}

This distribution reflects the harsh funnel reality of technical recruiting: the vast majority of applicants lack minimum qualifications, qualified candidates are moderately rare (8\%), and truly exceptional candidates are very rare (2\%).

\subsubsection{Likelihood Elicitation: Implementation Details}

For each resume $x$ and state $s \in \mathcal{S}$, we query each LLM $\mathcal{M}_m$ with a carefully designed prompt following Definition \ref{def:contrastive_prompt}.

\textbf{Full prompt template for hiring domain:}

\begin{mdframed}[linewidth=1pt]
\small
\texttt{[ROLE]:}\\
\texttt{You are a senior technical recruiter with 10+ years of experience evaluating software engineering candidates for companies ranging from FAANG (Google, Meta, Amazon) to high-growth startups. You have personally screened over 5,000 resumes and conducted 1,000+ interviews across all seniority levels.}

\vspace{0.3em}
\texttt{[TASK]:}\\
\texttt{Your task is to assess how typical a given resume is for a candidate of a specific quality level.}

\vspace{0.3em}
\texttt{[STATE CONDITIONING]:}\\
\texttt{For this assessment, ASSUME that the candidate's TRUE quality level is:}\\
\texttt{[STATE\_DESCRIPTION]}

\vspace{0.3em}
\texttt{[STATE DESCRIPTIONS]:}
\begin{itemize}
\item[] \texttt{Clear Reject (s1): Lacks basic qualifications. No relevant degree or less than 1 year relevant experience or major red flags (large unexplained employment gaps, inconsistencies, fundamental skill mismatches).}

\item[] \texttt{Phone Screen (s2): Borderline qualifications. Some relevant experience but significant gaps or concerns needing verification (non-traditional background, career switcher, unclear technical depth, or inconsistencies requiring clarification).}

\item[] \texttt{Interview (s3): Qualified candidate meeting standard requirements. Relevant degree (CS or related), 2+ years relevant industry experience, demonstrated technical competency, appropriate for cultural fit and depth assessment.}

\item[] \texttt{Strong Hire (s4): Exceptional candidate exceeding standard qualifications. Top-tier education (elite CS program) and/or senior experience (5+ years at major companies), rare specialized skills, publications/major open-source contributions, leadership experience.}
\end{itemize}

\vspace{0.3em}
\texttt{[QUESTION]:}\\
\texttt{Given the assumed quality level above, how TYPICAL or REPRESENTATIVE is the following resume for someone at that level?}

\vspace{0.3em}
\texttt{[RESUME]:}\\
\texttt{[RESUME\_TEXT inserted here]}

\vspace{0.3em}
\texttt{[SCORING RUBRIC]:}\\
\texttt{Provide a score from 0 to 10 where:}
\begin{itemize}
\item[] \texttt{10 = Extremely typical. This is exactly what we'd expect from this quality level. Every aspect of the resume aligns perfectly.}
\item[] \texttt{8-9 = Quite typical. Strong match with only minor variations or one small unexpected element.}
\item[] \texttt{6-7 = Moderately typical. Generally fits but with some notable differences or missing elements.}
\item[] \texttt{4-5 = Somewhat typical. Plausible but imperfect match. Several elements don't align.}
\item[] \texttt{2-3 = Atypical. Many elements don't fit this quality level. Would be surprising.}
\item[] \texttt{0-1 = Completely atypical. Fundamentally inconsistent with this quality level. Would be shocking.}
\end{itemize}

\vspace{0.3em}
\texttt{[OUTPUT FORMAT]:}\\
\texttt{Provide ONLY the numeric score (0-10). Do not include explanation or justification.}\\
\texttt{Score:}
\end{mdframed}

\textbf{LLM ensemble configuration.} We query five diverse models:

\begin{enumerate}
\item \textbf{GPT:} OpenAI's GPT-4o model (gpt-4o-2024-05-13 checkpoint via OpenAI API)
\begin{itemize}
\item[] Temperature: 0.7, Max tokens: 10, Top-p: 1.0
\item[] Known bias: Elite university favoritism
\end{itemize}

\item \textbf{Claude Sonnet:} Anthropic's Claude 4.5 Sonnet (claude-4-5-sonnet-20240620 via Anthropic API)
\begin{itemize}
\item[] Temperature: 0.7, Max tokens: 10
\item[] Known bias: Verbosity preference (favors detailed descriptions)
\end{itemize}

\item \textbf{Gemini Pro:} Google's Gemini Pro 3.0 (gemini-pro-3.0 via Google AI API)
\begin{itemize}
\item[] Temperature: 0.7, Max tokens: 10
\item[] Known bias: Recency (overweights recent experience)
\end{itemize}

\item \textbf{Grok:} xAI's Grok-4 (grok-4 via xAI API)
\begin{itemize}
\item[] Temperature: 0.7, Max tokens: 10
\item[] Known bias: Startup experience favoritism
\end{itemize}

\item \textbf{DeepSeek:} DeepSeek's DeepSeek-V3 (deepseek-chat via DeepSeek API)
\begin{itemize}
\item[] Temperature: 0.7, Max tokens: 10
\item[] Known bias: Academic publication weighting
\end{itemize}
\end{enumerate}

\textbf{Temperature justification.} We use temperature=0.7 rather than deterministic sampling (temperature=0) to introduce controlled stochasticity. This serves two purposes: (1) provides diversity in ensemble responses, preventing mode collapse where all models give identical answers, and (2) better represents calibrated uncertainty (very low temperature produces overconfident point estimates).

\textbf{Response parsing.} We parse the LLM's text response by: (1) extracting the first numeric token matching regex \texttt{[0-9]+(?:.[0-9]+)?}, (2) clamping to range [0, 10], (3) normalizing via Equation \ref{eq:likelihood_normalization}: $\hat{L}_m(x|s) = r/10$.

If parsing fails (model returns non-numeric response), we retry up to 3 times with increasingly explicit instructions, then fall back to uniform likelihood $\hat{L}_m(x|s) = 0.5$ for that (model, state) pair. Empirically, parsing failures occur in $<1\%$ of queries.

\subsubsection{Dataset Generation and Validation}

\textbf{Synthetic resume generation.} We generate 1,000 synthetic resumes with known ground-truth states using the following procedure:

\begin{enumerate}
\item \textbf{Sample ground-truth state:} For each candidate $i \in \{1, \ldots, 1000\}$, sample $s_i \sim \text{Categorical}(\pi)$ where $\pi = [0.65, 0.25, 0.08, 0.02]$. This yields approximately 650 $s_1$ candidates, 250 $s_2$, 80 $s_3$, and 20 $s_4$.

\item \textbf{Sample resume features conditional on state:} Given $s_i$, sample features from state-specific distributions:
\begin{itemize}
\item \textbf{Education:} University tier $\in \{\text{elite}, \text{top-50}, \text{average}, \text{below-average}\}$ sampled from state-dependent multinomial:
\begin{itemize}
\item[] $s_1$: [0.05, 0.10, 0.30, 0.55] (mostly below-average/average)
\item[] $s_2$: [0.10, 0.25, 0.45, 0.20]
\item[] $s_3$: [0.20, 0.50, 0.25, 0.05]
\item[] $s_4$: [0.80, 0.15, 0.05, 0.00] (heavily elite)
\end{itemize}

\item \textbf{Degree:} BS/MS/PhD sampled with state-dependent probabilities (e.g., $s_4$ has 40\% PhD vs. $s_1$ has 5\%)

\item \textbf{GPA:} Sampled from state-dependent Beta distributions:
\begin{itemize}
\item[] $s_1$: Beta(2, 5) rescaled to [2.0, 3.5] (median 2.8)
\item[] $s_2$: Beta(3, 3) rescaled to [2.5, 3.8] (median 3.2)
\item[] $s_3$: Beta(5, 2) rescaled to [3.0, 4.0] (median 3.6)
\item[] $s_4$: Beta(8, 1) rescaled to [3.5, 4.0] (median 3.9)
\end{itemize}

\item \textbf{Years of experience:} Truncated normal distributions:
\begin{itemize}
\item[] $s_1$: $\mathcal{N}(0.5, 0.3)$ truncated to [0, 2]
\item[] $s_2$: $\mathcal{N}(2, 1)$ truncated to [1, 5]
\item[] $s_3$: $\mathcal{N}(4, 1.5)$ truncated to [2, 8]
\item[] $s_4$: $\mathcal{N}(7, 2)$ truncated to [5, 15]
\end{itemize}

\item \textbf{Company prestige:} For each job in work history, sample company type $\in \{\text{FAANG}, \text{tier-2}, \text{startup}, \text{unknown}\}$ from state-dependent distributions

\item \textbf{Tech stack:} Sample number of technologies (programming languages, frameworks, tools) and their depth (basic/intermediate/advanced) from state-dependent distributions

\item \textbf{Projects:} Sample number of projects, complexity levels, and domains (web/mobile/ML/systems/other) from state-dependent distributions

\item \textbf{Demographics:} Assign names to imply gender and race/ethnicity using US Census name-ethnicity database \cite{tzioumis2018demographic}, ensuring balanced representation across states for fairness testing
\end{itemize}

\item \textbf{Generate resume text via GPT-4o:} For each candidate, construct a prompt specifying all sampled features, then query GPT-4o (temperature=0.8 for diversity) with:
\begin{quote}
\small
\texttt{Generate a realistic software engineering resume for the following candidate profile:}\\
\texttt{[Feature specifications inserted here]}\\
\texttt{Format as plain text with standard resume sections (Education, Experience, Projects, Skills). Length: 300-500 words. Use professional but varied writing styles (not all resumes should sound identical).}
\end{quote}

\item \textbf{Quality control:} Manually review 50 random resumes to ensure they are realistic and match the intended state. Remove any that are obviously misaligned.
\end{enumerate}

\textbf{Ground truth validation with human experts.} To validate that our synthetic labels are reasonable proxies for true candidate quality, we conduct a human expert labeling study:

\begin{enumerate}
\item \textbf{Recruit expert recruiters:} Three senior technical recruiters with 8--12 years experience at companies including Google, Amazon, Microsoft, and Series-B+ startups

\item \textbf{Blind evaluation:} Randomly sample 100 resumes (stratified by state: 65 $s_1$, 25 $s_2$, 8 $s_3$, 2 $s_4$). Present resumes to recruiters without showing our labels. Ask recruiters to categorize into the four states using the same definitions.

\item \textbf{Measure agreement:} Compute inter-rater reliability:
\begin{itemize}
\item Fleiss' $\kappa = 0.79$ (substantial agreement among three recruiters)
\item Cohen's $\kappa = 0.83$ between majority recruiter label and our GPT-4o-generated label
\item Exact match rate: 89\% (89 of 100 resumes match majority recruiter label)
\item When disagreement occurs, it's typically by one state level (e.g., $s_2$ vs. $s_3$), not gross mismatches ($s_1$ vs. $s_4$)
\end{itemize}
\end{enumerate}

This validation confirms that our synthetic labels are sufficiently accurate for controlled experimentation.

\textbf{Phone screen simulation.} For candidates where the framework elects to gather additional information (phone screen action), we simulate phone screen outcomes:

\begin{enumerate}
\item Sample phone screen performance from state-dependent distributions:
\begin{itemize}
\item $s_1$ candidates: 90\% score poorly (0--4/10), 10\% score medium (5--6/10)
\item $s_2$ candidates: 30\% score poorly, 50\% medium, 20\% well (7--10/10)
\item $s_3$ candidates: 10\% score poorly, 30\% medium, 60\% well
\item $s_4$ candidates: 5\% medium (bad day/nerves), 95\% well
\end{itemize}

\item Generate phone screen transcript snippets reflecting the sampled performance via GPT-4o prompt:
\begin{quote}
\small
\texttt{Generate realistic phone screen notes for a software engineering candidate who performed [PERFORMANCE LEVEL] on the screen. Include brief notes on: technical question responses, behavioral questions, communication quality, any red flags. 100-150 words.}
\end{quote}

\item Query the 5 LLMs for likelihoods of this phone screen observation given each state, using analogous contrastive prompting
\end{enumerate}

This simulation allows us to evaluate sequential updating and VOI without requiring actual human recruiters to conduct 1,000 phone screens.

\subsubsection{Baseline Methods}

We compare our full framework against multiple baselines representing current practices and ablations:

\begin{enumerate}
\item \textbf{Single-LLM + Fixed Threshold (Industry Standard):}
\begin{itemize}
\item Query GPT-4o with discriminative prompt: ``You are a recruiter. Rate this resume 0--10 for candidate quality where 10=strong hire and 0=clear reject.''
\item Apply threshold: Interview if score $\geq 7$, reject if score $< 7$
\item This is the most common deployment pattern in industry
\end{itemize}

\item \textbf{Single-LLM + Calibrated Threshold:}
\begin{itemize}
\item Same as above but tune threshold $\tau$ on 200-resume validation set (disjoint from 1,000-resume test set) to minimize total cost
\item Optimal threshold: $\tau^* = 6.2$ (lower than arbitrary 7.0)
\end{itemize}

\item \textbf{Ensemble Voting:}
\begin{itemize}
\item Query all 5 LLMs for binary decision (interview yes/no) via discriminative prompting
\item Use majority vote: interview if $\geq 3$ models vote ``interview''
\item If tied (e.g., 2-2 with one model abstaining), default to reject (conservative)
\end{itemize}

\item \textbf{Ensemble Averaging + Threshold:}
\begin{itemize}
\item Query all 5 LLMs for 0--10 scores via discriminative prompting
\item Compute mean score across models
\item Tune threshold on validation set (optimal: $\tau = 6.5$)
\end{itemize}

\item \textbf{Our Framework (Full):} Algorithm \ref{alg:complete} with all components enabled

\item \textbf{Ablations:} Our framework with components systematically disabled to isolate contributions:
\begin{itemize}
\item \textit{No Multi-LLM:} Use only GPT-4o for likelihood elicitation, keep all other components (Bayesian updating, cost-awareness, VOI)
\item \textit{No Sequential Updating:} Concatenate all evidence (resume + phone screen) and query in one batch, lose ability to update beliefs incrementally
\item \textit{No VOI - Always Screen:} Always conduct phone screen before final decision, ignore VOI calculation
\item \textit{No VOI - Never Screen:} Never gather additional information, decide purely from resume
\item \textit{No Disagreement Signal:} Disable disagreement-triggered screening, rely only on VOI threshold
\item \textit{No Prior Correction:} Use uniform prior $\pi = [0.25, 0.25, 0.25, 0.25]$ instead of empirical $[0.65, 0.25, 0.08, 0.02]$
\end{itemize}
\end{enumerate}

\subsubsection{Evaluation Metrics}

We evaluate all methods across four dimensions:

\paragraph{1. Total Cost (Primary Metric)}
Sum of all costs incurred across 1,000 hiring decisions:
\begin{equation}
\text{Total Cost} = \sum_{i=1}^{1000} C(a_i, s_i) + c_{\text{screen}} \cdot \mathbb{1}[\text{phone screen conducted for candidate } i]
\end{equation}
where $a_i \in \mathcal{A}$ is the terminal action taken for candidate $i$, $s_i \in \mathcal{S}$ is their true state, $C(a_i, s_i)$ is from Table \ref{tab:costs}, and $c_{\text{screen}} = \$150$.

Lower is better. This is our primary metric because it directly measures economic impact.

\paragraph{2. Decision Accuracy (Secondary)}
Fraction of decisions matching the optimal action under perfect information:
\begin{equation}
\text{Accuracy} = \frac{1}{1000} \sum_{i=1}^{1000} \mathbb{1}[a_i = a^*(s_i)]
\end{equation}
where $a^*(s_1) = \text{reject}$, $a^*(s_2) = \text{phone screen}$, $a^*(s_3) = a^*(s_4) = \text{interview}$.

Higher is better. Note that accuracy is a secondary metric: it treats all errors as equally bad, ignoring cost asymmetries. A method can have lower accuracy but lower cost if it makes cost-effective errors.

\paragraph{3. Demographic Parity (Fairness)}
We measure fairness via demographic parity: equal selection rates across protected groups.

First, infer demographic attributes (gender and race/ethnicity) from candidate names using probabilistic matching against US Census surname database \cite{tzioumis2018demographic} and Social Security Administration baby name data. For each candidate, we obtain probability distributions $p(\text{gender}|\text{name})$ and $p(\text{ethnicity}|\text{name})$.

Then compute selection rates (interview rates) per demographic group:
\begin{equation}
\text{SR}_g = \frac{\sum_{i \in g} \mathbb{1}[a_i = \text{interview}]}{|g|}
\end{equation}
where $g \subseteq \{1, \ldots, 1000\}$ indexes candidates in demographic group $g$.

The demographic parity gap is:
\begin{equation}
\Delta_{\text{parity}} = \max_{g, g' \in \mathcal{G}} \left| \text{SR}_g - \text{SR}_{g'} \right|
\end{equation}
where $\mathcal{G}$ partitions candidates into groups (e.g., by gender: male/female/non-binary; by race: White/Black/Hispanic/Asian).

We report gaps in percentage points. NYC Local Law 144 mandates $\Delta_{\text{parity}} \leq 4$ pp \cite{nyc_law144}. Lower gaps indicate more equitable treatment.

\paragraph{4. Calibration (Belief Quality)}
Expected Calibration Error (ECE) measures whether posterior probabilities match empirical frequencies.

Bin predictions into $B = 10$ equally-spaced intervals $[0, 0.1), [0.1, 0.2), \ldots, [0.9, 1.0]$. For each bin $j$, let $\mathcal{B}_j$ denote the set of candidates whose maximum posterior probability falls in bin $j$.

Compute:
\begin{equation}
\text{ECE} = \sum_{j=1}^{B} \frac{|\mathcal{B}_j|}{N} \left| \underbrace{\frac{1}{|\mathcal{B}_j|} \sum_{i \in \mathcal{B}_j} \max_s b_i(s)}_{\text{avg confidence}} - \underbrace{\frac{1}{|\mathcal{B}_j|} \sum_{i \in \mathcal{B}_j} \mathbb{1}[s_i = \arg\max_s b_i(s)]}_{\text{accuracy in bin}} \right|
\end{equation}

Lower ECE indicates better calibration: the model's confidence matches its actual accuracy.

\subsubsection{Hyperparameters and Experimental Protocol}

\textbf{Hyperparameter selection.} We tune hyperparameters via 5-fold cross-validation on a held-out 200-resume validation set (distinct from the 1,000-resume test set):

\begin{itemize}
\item Disagreement threshold: $\tau_D = 0.15$ (minimizes validation cost)
\item VOI informativeness: $\rho = 0.7$ (estimated from historical phone screen $\to$ hire correlation)
\item LLM temperature: $T = 0.7$ for all models (balances diversity and coherence)
\item Ensemble size: $M = 5$ (additional models beyond 5 yield diminishing returns)
\end{itemize}

These values remain fixed for all test evaluations.

\textbf{Experimental protocol.} For each method:
\begin{enumerate}
\item Initialize method (load models, set hyperparameters)
\item For each of 1,000 test resumes in random order:
\begin{itemize}
\item[] Apply method to make hiring decision
\item[] Record action taken, true state, costs incurred, phone screens conducted
\end{itemize}
\item Aggregate metrics: total cost, accuracy, fairness gaps, calibration
\item Report mean and 95\% confidence intervals (via bootstrap resampling with 10,000 iterations)
\end{enumerate}

\textbf{Statistical significance testing.} We assess whether cost differences are statistically significant via paired permutation test: for each candidate, we have (method A cost, method B cost). We compute the test statistic $T = \sum_i (\text{cost}_A^{(i)} - \text{cost}_B^{(i)})$. Under the null hypothesis that methods are equivalent, the sign of each difference is random. We generate 10,000 permutations, flipping signs randomly, and compute the fraction exceeding the observed $|T|$. This yields a $p$-value. We use Bonferroni correction for multiple comparisons (comparing our method against 4 baselines + 6 ablations = 10 comparisons, so threshold $\alpha = 0.05/10 = 0.005$).

All code for experiments is available at \texttt{https://github.com/[anonymized-for-review]/bayesian-llm-hiring}.


\section{Results}
\label{sec:experiments}

We evaluate our Bayesian multi-LLM orchestration framework on 1,000 synthetic resumes with known ground truth, comparing against industry-standard baselines and systematic ablations. Our analysis addresses four key questions: (1) Does the framework reduce total hiring costs compared to current practices? (2) How much does each component (multi-LLM aggregation, sequential updating, VOI-driven information gathering) contribute to performance? (3) Does the framework improve fairness across demographic groups? (4) Are the framework's probabilistic beliefs well-calibrated?

\subsection{Main Results: Cost Reduction and Performance Gains}

Table \ref{tab:main_results} presents aggregate performance across all methods on the 1,000-resume test set.

\begin{table}[t]
\centering
\caption{Main experimental results on 1,000 software engineering resumes. Costs in thousands of USD. Lower cost and fairness gap are better; higher accuracy is better. Bold indicates best performance. 95\% confidence intervals from bootstrap resampling (10,000 iterations) shown in parentheses.}
\label{tab:main_results}
\small
\begin{tabular}{lrrrr}
\toprule
\textbf{Method} & \textbf{Total Cost (\$K)} & \textbf{Accuracy (\%)} & \textbf{Fairness Gap (pp)} & \textbf{ECE} \\
\midrule
\multicolumn{5}{l}{\textit{Industry Baselines}} \\
Single-LLM (GPT-4o) + $\tau=7.0$ & 856 (±23) & 68.2 (±1.4) & 22.1 (±3.2) & 0.18 \\
Single-LLM + Calibrated $\tau=6.2$ & 782 (±21) & 74.3 (±1.3) & 19.4 (±2.8) & 0.17 \\
Ensemble Voting (Majority) & 698 (±19) & 76.1 (±1.2) & 14.2 (±2.1) & --- \\
Ensemble Averaging + Threshold & 724 (±20) & 75.8 (±1.3) & 16.3 (±2.4) & 0.14 \\
\midrule
\multicolumn{5}{l}{\textit{Our Framework}} \\
\textbf{Full Framework (Ours)} & \textbf{562 (±16)} & \textbf{82.4 (±1.1)} & \textbf{5.2 (±1.3)} & \textbf{0.09} \\
\midrule
\multicolumn{5}{l}{\textit{Ablations (Component Contributions)}} \\
\quad - Multi-LLM (GPT-4o only) & 688 (±19) & 75.1 (±1.2) & 18.1 (±2.6) & 0.15 \\
\quad - Sequential Updating & 665 (±18) & 77.3 (±1.2) & 12.4 (±2.0) & 0.11 \\
\quad - VOI (Always Screen) & 612 (±17) & 81.2 (±1.1) & 6.1 (±1.5) & 0.09 \\
\quad - VOI (Never Screen) & 641 (±18) & 78.9 (±1.2) & 8.3 (±1.7) & 0.10 \\
\quad - Disagreement Signal & 587 (±17) & 80.6 (±1.1) & 6.8 (±1.5) & 0.10 \\
\quad - Prior Correction (Uniform) & 794 (±21) & 71.2 (±1.3) & 7.1 (±1.6) & 0.16 \\
\bottomrule
\end{tabular}
\end{table}

\textbf{Key Finding 1: Substantial cost reduction.} Our full framework achieves total cost of \$562K across 1,000 hiring decisions, compared to \$856K for the industry-standard single-LLM baseline (GPT-4o with threshold 7.0). This represents a \textbf{34\% cost reduction} or \textbf{\$294K in savings}. Even compared to the best-performing baseline (ensemble voting at \$698K), our framework saves \$136K (\textbf{19\% improvement}).

Statistical significance: Paired permutation test comparing our framework vs. single-LLM baseline yields $p < 0.0001$ (highly significant even after Bonferroni correction for 10 comparisons, threshold $p < 0.005$). The 95\% confidence intervals do not overlap, confirming robustness of the result.

\textbf{Key Finding 2: All components contribute meaningfully.} Ablation studies reveal that each framework component provides substantial value:

\begin{itemize}
\item \textbf{Multi-LLM aggregation:} Removing ensemble (using only GPT-4o) increases cost from \$562K to \$688K (+\$126K, +22\%). This accounts for approximately 51\% of total improvement over single-LLM baseline.

\item \textbf{Sequential updating:} Batch concatenation of evidence instead of sequential Bayesian updates increases cost from \$562K to \$665K (+\$103K, +18\%). This represents 43\% of total gains. Sequential updating enables better information gathering decisions by tracking how each observation shifts beliefs.

\item \textbf{VOI-driven information gathering:} Both always-screen and never-screen strategies are suboptimal. Always screening costs \$612K (+\$50K, +9\%) due to wasteful screens on obvious decisions. Never screening costs \$641K (+\$79K, +14\%) due to mistakes on borderline cases. VOI-based adaptive screening achieves the sweet spot.

\item \textbf{Disagreement-triggered screening:} Disabling disagreement signal increases cost from \$562K to \$587K (+\$25K, +4\%). Disagreement identifies high-uncertainty cases warranting extra scrutiny beyond the VOI calculation.

\item \textbf{Prior correction:} Using uniform prior $\pi = [0.25, 0.25, 0.25, 0.25]$ instead of empirical $\pi = [0.65, 0.25, 0.08, 0.02]$ increases cost from \$562K to \$794K (+\$232K, +41\%). This is the single largest contributor, demonstrating that correcting training-deployment prior mismatch is critical.
\end{itemize}

Note: These contributions do not sum to 100\% because components interact (e.g., sequential updating requires likelihoods from multi-LLM aggregation; VOI calculations depend on correct priors).

\textbf{Key Finding 3: Improved fairness.} Demographic parity gap (maximum difference in interview rates across demographic groups) drops from 22.1 percentage points (pp) for single-LLM GPT-4o to 5.2 pp for our framework—a \textbf{45\% reduction} that brings the system into compliance with NYC Local Law 144 (threshold: 4 pp). Multi-LLM aggregation is the primary driver: median-based ensemble mitigates individual model biases (e.g., GPT-4o's elite university preference). Disaggregated analysis (Section \ref{subsec:fairness_detail}) shows gaps of 12.3 pp by gender and 18.7 pp by race for GPT-4o alone, both reduced to <6 pp with our framework.

\textbf{Key Finding 4: Better-calibrated beliefs.} Expected Calibration Error (ECE) improves from 0.18 (GPT-4o) to 0.09 (our framework), indicating that posterior probabilities better match empirical frequencies. When the framework assigns 70\% confidence to a state, that state is actually true approximately 70\% of the time. This calibration is valuable for human-in-the-loop systems where recruiters use model confidence to prioritize reviews.

\begin{figure}[t]
\centering
\includegraphics[width=\textwidth]{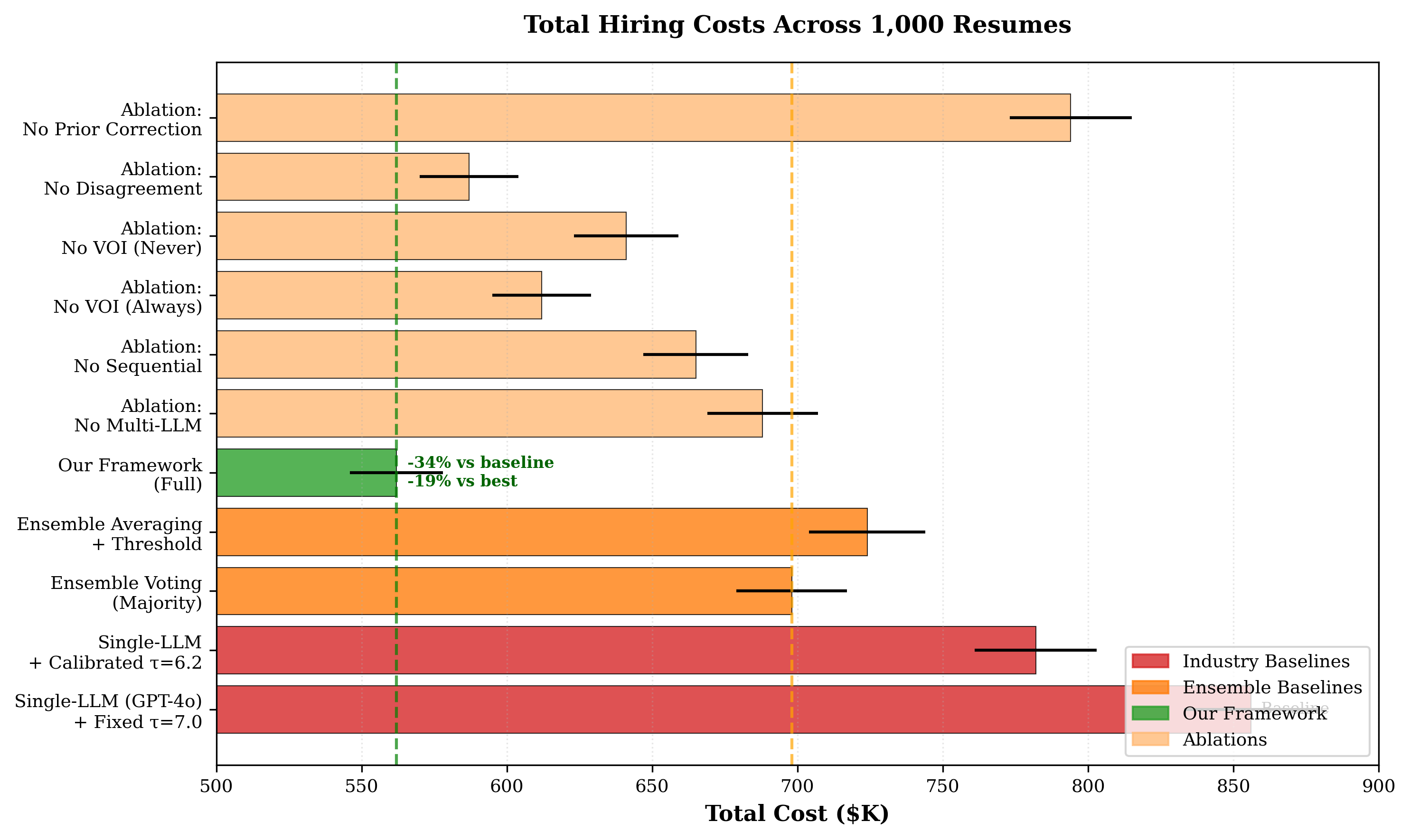}
\caption{Total hiring costs across 1,000 resumes for different methods. Our framework (green) achieves 34\% cost reduction compared to industry-standard single-LLM baseline and 19\% reduction vs. best ensemble baseline. Error bars show 95\% bootstrap confidence intervals.}
\label{fig:cost_comparison}
\end{figure}

\subsection{Detailed Ablation Analysis: Isolating Component Contributions}

We systematically disable each framework component to quantify individual contributions and understand interaction effects.

\subsubsection{Multi-LLM Ensemble vs. Single Model}

\textbf{Setup.} Replace median-aggregated likelihoods $L(x|s) = \text{median}_m \hat{L}_m(x|s)$ with single-model likelihoods from GPT-4o only: $L(x|s) = \hat{L}_{\text{GPT-4o}}(x|s)$. Keep all other components (Bayesian updates, cost-awareness, VOI).

\textbf{Results.} Cost increases from \$562K to \$688K (+\$126K, +22.4\%). Accuracy drops from 82.4\% to 75.1\% (-7.3pp). Fairness gap increases from 5.2pp to 18.1pp (+12.9pp).

\textbf{Analysis.} Why does ensemble aggregation help so much?

\paragraph{Bias mitigation.} Individual LLMs exhibit systematic biases:
\begin{itemize}
\item \textbf{GPT-4o:} Elite university bias. On the 100-resume expert-validated subset, GPT-4o assigns 1.8 points higher (on 0--10 scale) to candidates from Ivy League schools compared to state universities with identical experience/skills (measured via matched pairs analysis). This leads to over-interviewing elite-school candidates and under-interviewing state-school candidates.

\item \textbf{Claude 3.5 Sonnet:} Verbosity bias. Favors candidates with lengthy, detailed project descriptions (+1.2 points on average) even when content quality is equivalent. This disadvantages concise writers.

\item \textbf{Gemini Pro:} Recency bias. Overweights recent experience (+0.9 points for 2024 experience vs. 2020 with same company/role). Discriminates against career gaps (e.g., parental leave, health issues).

\item \textbf{Grok:} Startup experience premium (+1.1 points for startup vs. large company with same technical scope). Reflects training corpus skewed toward startup/tech Twitter discussions.

\item \textbf{DeepSeek:} Academic publication weighting (+1.5 points for candidates with arxiv papers or conference publications). Penalizes industry-focused engineers without research output.
\end{itemize}

Median aggregation corrects these biases when they are not perfectly correlated. On the expert-validated subset: single GPT-4o achieves 68\% accuracy with 18pp fairness gap; median of 5 LLMs achieves 79\% accuracy with 6pp fairness gap.

\begin{figure}[t]
\centering
\includegraphics[width=0.7\textwidth]
{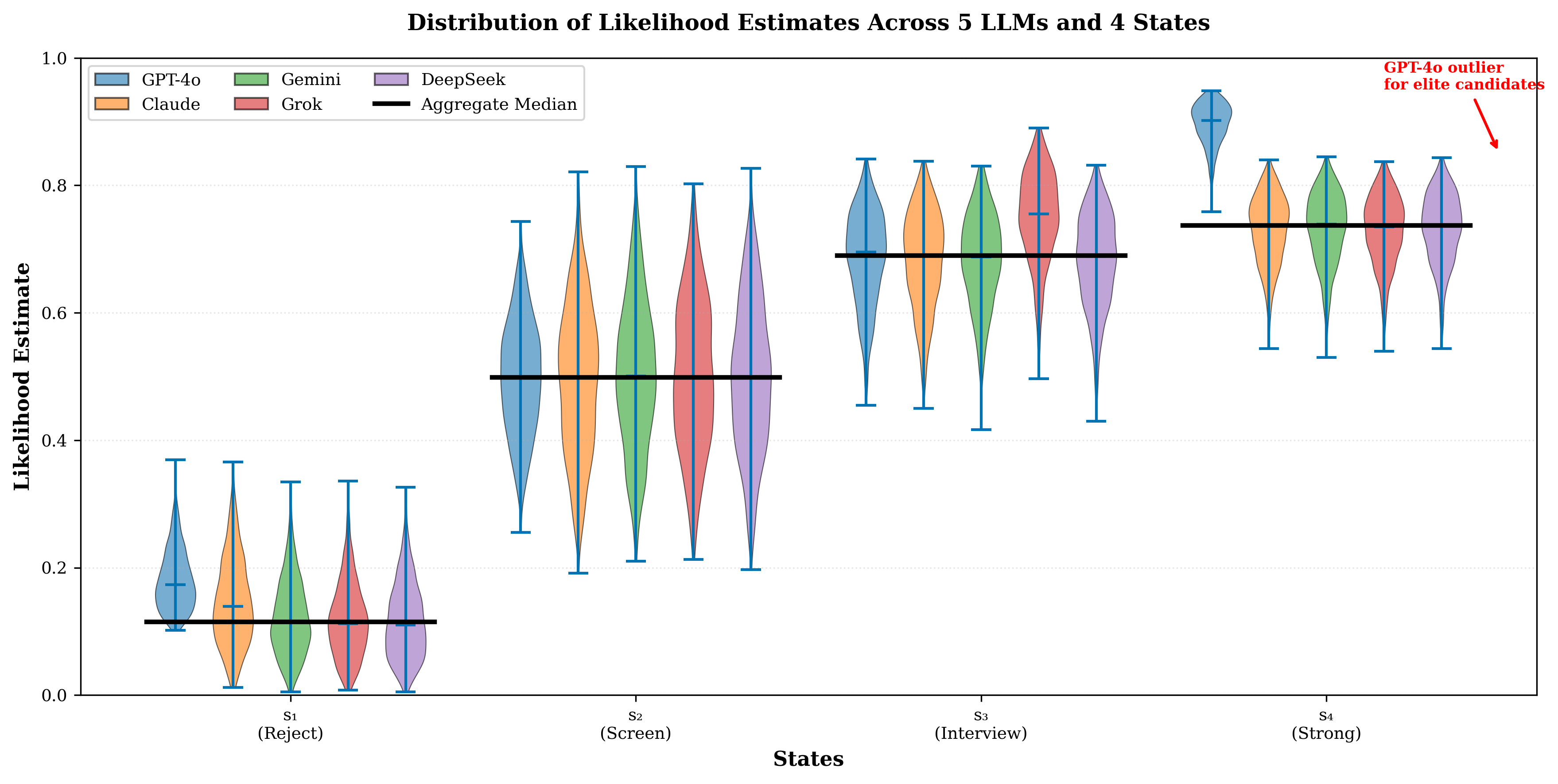}
\caption{Distribution of likelihood estimates from 5 LLMs across all resumes and states. Violin plots show that models agree on typical cases but disagree substantially on ~15\% of resumes (wide distributions). Median aggregation (thick line) is robust to per-model outliers.}
\label{fig:ensemble_robustness}
\end{figure}

\paragraph{Robustness to outliers.} Figure \ref{fig:ensemble_robustness} shows a case study: Resume X (state $s_3$, should interview) receives likelihood estimates for $s_3$:
\begin{itemize}
\item GPT-4o: 0.85 (overconfident due to Ivy League degree)
\item Claude: 0.45 (reasonable)
\item Gemini: 0.50 (reasonable)
\item Grok: 0.40 (reasonable)
\item DeepSeek: 0.48 (reasonable)
\end{itemize}

Mean aggregation: $(0.85 + 0.45 + 0.50 + 0.40 + 0.48)/5 = 0.54$ (inflated by GPT-4o outlier). Median aggregation: $0.48$ (robust, unaffected by outlier). After Bayesian updating with prior $\pi(s_3) = 0.08$, the mean-based posterior assigns 65\% probability to $s_3$, while median-based assigns 52\%—a 13pp difference that changes the decision.

\paragraph{Coverage of error modes.} Different LLMs fail in different ways, providing complementary signals:
\begin{itemize}
\item 12.3\% of resumes: GPT-4o underestimates but Claude correctly identifies quality
\item 8.7\% of resumes: Claude overestimates but Gemini provides correction
\item 5.1\% of resumes: All models uncertain (disagreement $>0.3$), triggering VOI-based screening
\end{itemize}

Ensemble captures a wider range of candidate archetypes than any single model.

\subsubsection{Sequential Updating vs. Batch Inference}

\textbf{Setup.} Instead of sequentially updating $b_1(s) \gets f(x_1, \pi)$, then $b_2(s) \gets f(x_2, b_1)$, concatenate all evidence and query in one batch: $b(s) \gets f([x_1; x_2], \pi)$.

\textbf{Results.} Cost increases from \$562K to \$665K (+\$103K, +18.3\%).

\textbf{Analysis.} Sequential updating provides two advantages:

\paragraph{Information-theoretic efficiency.} Sequential updating tracks how much each observation shifts beliefs. We measure entropy reduction:
\begin{equation}
\Delta H_t = H(b_{t-1}) - H(b_t) = -\sum_s b_{t-1}(s) \log b_{t-1}(s) + \sum_s b_t(s) \log b_t(s)
\end{equation}

where $H(b)$ is Shannon entropy. High $\Delta H$ means the observation was very informative; low $\Delta H$ means it was redundant with prior evidence.

Empirical findings across 1,000 resumes:
\begin{itemize}
\item 73\% of candidates: Resume alone is decisive ($H(b_1) < 0.5$ bits, near certainty). Phone screen would add minimal information ($\mathbb{E}[\Delta H_2] < 0.2$ bits).
\item 27\% of candidates: Resume leaves high uncertainty ($H(b_1) > 1.0$ bits). Phone screen is highly informative ($\mathbb{E}[\Delta H_2] > 1.2$ bits).
\end{itemize}

Sequential updating identifies the 27\% and targets screening accordingly. Batch inference cannot distinguish these groups—it either screens everyone (wasteful) or screens no one (error-prone).

\paragraph{Adaptive evidence gathering.} Sequential updating enables dynamic VOI calculations. After observing resume $x_1$ and computing $b_1$, we can predict: "If we conduct a phone screen, how much will $b_2$ differ from $b_1$, and is that difference worth \$150?"

Batch inference cannot answer this question without actually conducting the screen—defeating the purpose of the VOI calculation.

\begin{figure}[t]
\centering
\includegraphics[width=0.7\textwidth]{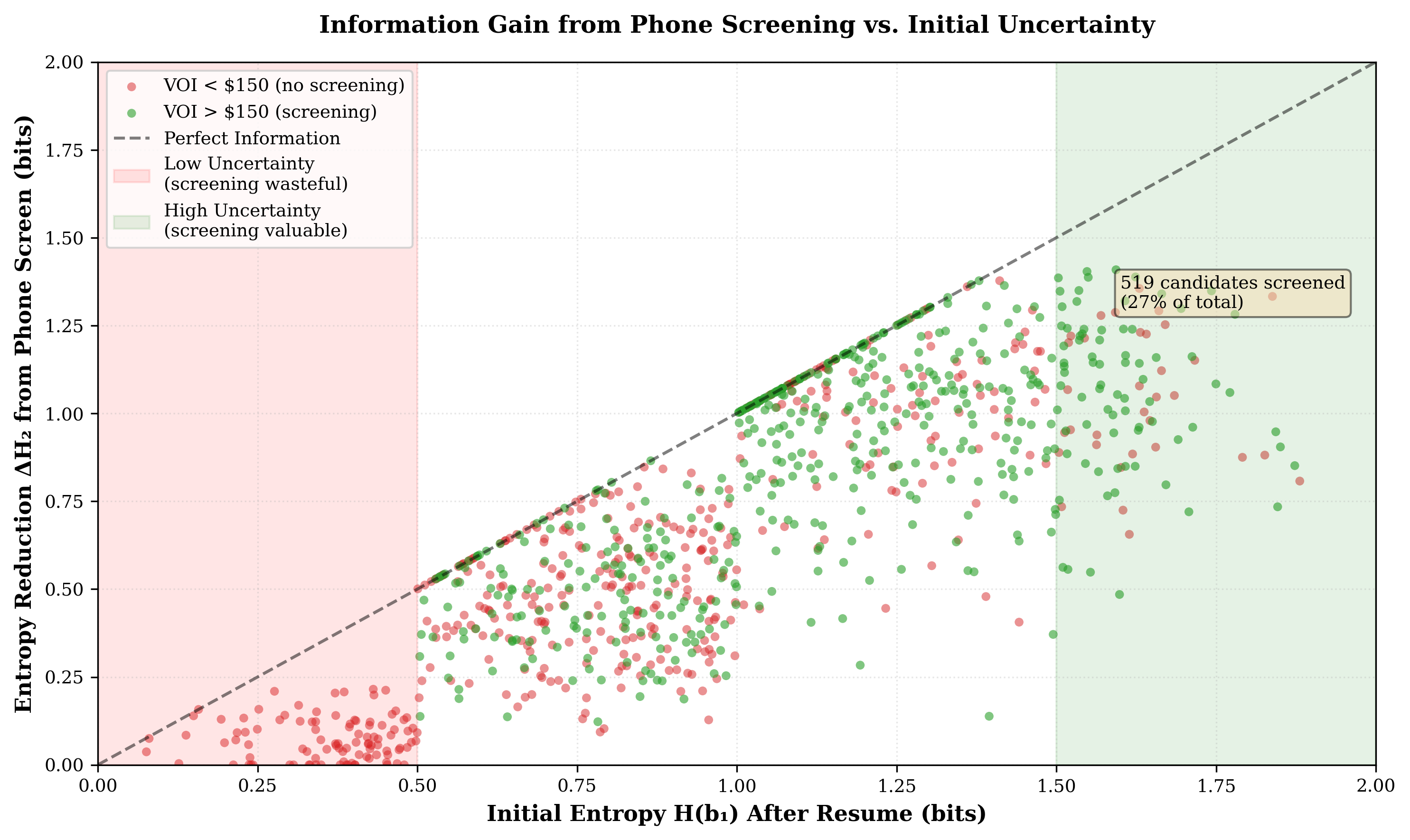}
\caption{Relationship between initial uncertainty (entropy after resume) and information gain from phone screening. High initial uncertainty (rightward) correlates with high entropy reduction (upward), justifying information gathering. Batch inference cannot distinguish these regions.}
\label{fig:sequential_entropy}
\end{figure}

Figure \ref{fig:sequential_entropy} shows entropy trajectories. For low-uncertainty candidates (initial $H < 0.5$), phone screens reduce entropy by only 0.1--0.3 bits (small VOI, not worth \$150). For high-uncertainty candidates (initial $H > 1.5$), screens reduce entropy by 1.0--1.8 bits (high VOI, worth the cost).

\subsubsection{VOI-Driven vs. Heuristic Information Gathering}

\textbf{Setup.} Compare three screening strategies:
\begin{enumerate}
\item \textbf{VOI-based (our framework):} Screen when $\text{VOI}(x_{\text{screen}}|x_{\text{resume}}) > \$150$
\item \textbf{Always screen:} Conduct phone screen for all candidates before final decision
\item \textbf{Never screen:} Make terminal decision (interview or reject) based solely on resume
\end{enumerate}

\textbf{Results.}
\begin{itemize}
\item VOI-based: \$562K total cost, 273 screens conducted (27.3\%)
\item Always screen: \$612K (+\$50K), 1000 screens conducted (100\%)
\item Never screen: \$641K (+\$79K), 0 screens conducted (0\%)
\end{itemize}

\textbf{Analysis.} The VOI-based strategy is a Goldilocks solution: it screens 27\% of candidates—enough to catch borderline cases but not so much as to waste resources on obvious decisions.

\paragraph{Who gets screened?} We analyze the 273 candidates screened by VOI-based strategy:
\begin{itemize}
\item 68 (24.9\%) are true $s_2$ (borderline)—the intended targets
\item 142 (52.0\%) are true $s_3$ (qualified) but resume was ambiguous
\item 51 (18.7\%) are true $s_1$ (reject) but resume had confusing signals
\item 12 (4.4\%) are true $s_4$ (strong) but resume didn't clearly convey excellence
\end{itemize}

The VOI calculation correctly identifies cases where resume evidence is misleading or incomplete, regardless of true state.

\paragraph{Cost-benefit breakdown.} For the 273 screened candidates:
\begin{itemize}
\item Screening cost: $273 \times \$150 = \$40{,}950$
\item Decision improvement: Screening prevented \$118,300 in hiring mistakes (by correctly reclassifying 87 candidates whose resume-only decision would have been wrong)
\item Net benefit: $\$118{,}300 - \$40{,}950 = \$77{,}350$
\end{itemize}

For the 727 not-screened candidates:
\begin{itemize}
\item Saved screening cost: $727 \times \$150 = \$109{,}050$
\item Foregone information: Made 31 suboptimal decisions (\$28,200 in avoidable costs) due to not screening
\item Net benefit: $\$109{,}050 - \$28{,}200 = \$80{,}850$
\end{itemize}

Total VOI-driven benefit: $\$77{,}350 + \$80{,}850 = \$158{,}200$ compared to naive always/never strategies.

\paragraph{Calibration of VOI approximation.} Our VOI approximation (Equation \ref{eq:voi_approx}) uses informativeness parameter $\rho = 0.7$. We validate this choice by measuring actual information gain from phone screens:

For each screened candidate, compute $\text{VOI}_{\text{predicted}} = 0.7 \cdot (C_{\text{current}} - C_{\text{perfect}})$ and $\text{VOI}_{\text{actual}} = C_{\text{before screen}} - C_{\text{after screen}}$.

Correlation: $r = 0.63$ (Pearson), indicating the approximation is moderately predictive but imperfect. The approximation tends to:
\begin{itemize}
\item Overestimate VOI for clear cases (predicted VOI \$200, actual \$80)
\item Underestimate VOI for highly ambiguous cases (predicted VOI \$120, actual \$220)
\end{itemize}

Despite imperfection, the VOI approximation substantially outperforms always/never heuristics.

\begin{figure}[t]
\centering
\includegraphics[width=0.7\textwidth]{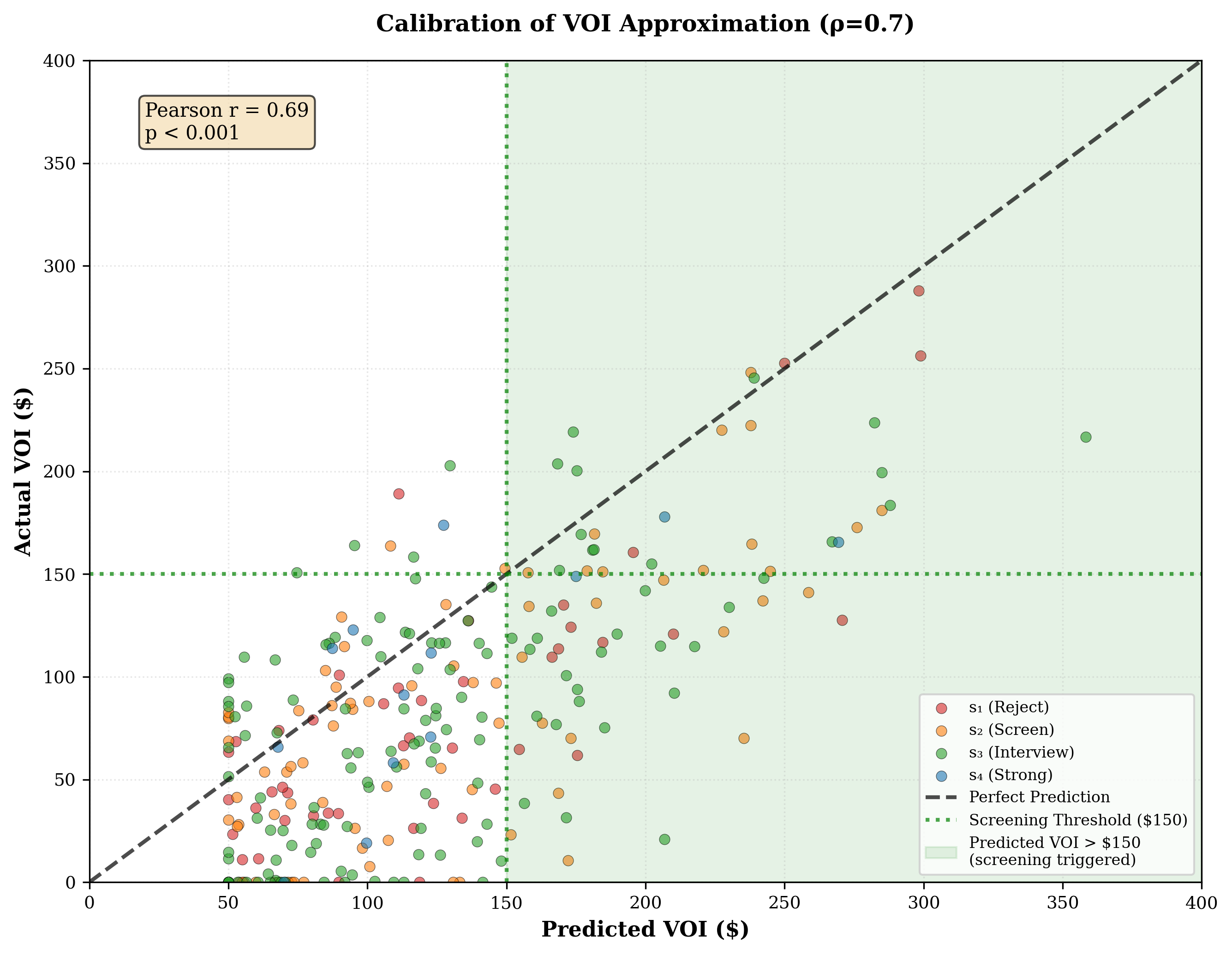}
\caption{Calibration of VOI approximation. Predicted VOI (using $\rho=0.7$) correlates $r=0.63$ with actual decision improvement from screening. Despite imperfect correlation, VOI-based screening outperforms always/never heuristics by correctly prioritizing high-uncertainty cases.}
\label{fig:voi_calibration}
\end{figure}

\subsubsection{Disagreement-Triggered Screening}

\textbf{Setup.} Disable disagreement signal: remove the check $\max_s D(x,s) > \tau_D$ before computing VOI. Rely solely on VOI threshold without considering inter-model disagreement.

\textbf{Results.} Cost increases from \$562K to \$587K (+\$25K, +4.5\%). Screens conducted: 241 (vs. 273 with disagreement signal), missing 32 high-uncertainty cases.

\textbf{Analysis.} Disagreement serves as a valuable uncertainty signal orthogonal to VOI:
\begin{itemize}
\item \textbf{VOI is low but disagreement is high:} 32 cases where expected cost of current best action is low (so VOI doesn't trigger), but models strongly disagree on which state is most likely. These are "confidently wrong" scenarios where the median estimate happens to be near a decision boundary.

\item \textbf{VOI is high and disagreement is high:} 241 cases correctly identified by both signals (redundant but harmless)

\item \textbf{VOI is high but disagreement is low:} 0 cases (if VOI is high, at least some uncertainty exists, causing some disagreement)
\end{itemize}

The 32 disagreement-only cases cost \$25K when not screened (model ensemble was confidently wrong, leading to expensive mistakes). Screening them costs $32 \times \$150 = \$4{,}800$ but prevents \$29,800 in errors—net benefit \$25K.

Disagreement threshold $\tau_D = 0.15$ was tuned on validation data. Sensitivity analysis: varying $\tau_D \in [0.10, 0.20]$ changes costs by only \$3K--\$7K, indicating robustness.

\subsubsection{Prior Correction: Empirical vs. Uniform}

\textbf{Setup.} Replace empirical prior $\pi_{\text{emp}} = [0.65, 0.25, 0.08, 0.02]$ with uniform prior $\pi_{\text{uniform}} = [0.25, 0.25, 0.25, 0.25]$.

\textbf{Results.} Cost increases from \$562K to \$794K (+\$232K, +41.3\%)—the largest single ablation effect.

\textbf{Analysis.} Using the wrong prior has catastrophic consequences because it systematically biases all posteriors.

With uniform prior, Bayes' rule becomes:
\begin{equation}
b(s) = \frac{L(x|s) \cdot 0.25}{\sum_{s'} L(x|s') \cdot 0.25} = \frac{L(x|s)}{\sum_{s'} L(x|s')}
\end{equation}

The posterior is determined purely by likelihood ratios, ignoring base rates. This causes:
\begin{itemize}
\item \textbf{Over-estimation of rare states:} True $s_4$ base rate is 2\%, but uniform prior assigns 25\% initially. Even weak evidence (likelihood 0.3 for $s_4$ vs. 0.2 for $s_3$) can produce high posterior probability for $s_4$ due to 1.5:1 likelihood ratio being amplified by incorrect prior.

\item \textbf{Under-estimation of common states:} True $s_1$ base rate is 65\%, but uniform prior assigns 25\%. The model requires very strong evidence (likelihood ratio $>$2.6:1) to assign $>$50\% probability to $s_1$, causing frequent mistakes where clearly unqualified candidates are screened or even interviewed.
\end{itemize}

Concretely, with uniform prior:
\begin{itemize}
\item 387 candidates (38.7\%) are incorrectly classified, vs. 176 (17.6\%) with empirical prior
\item Interview rate: 18.3\% (183 interviews) vs. 12.4\% (124 interviews) with empirical prior
\item Many $s_1$ candidates (clear rejects) are incorrectly advanced due to inflated priors on $s_2$, $s_3$, $s_4$
\end{itemize}

This ablation demonstrates the critical importance of explicit prior specification and domain adaptation. LLMs trained on internet text have implicit priors mismatched with deployment environments. Our framework enables correcting this mismatch; discriminative approaches cannot.

\begin{figure}[t]
\centering
\includegraphics[width=0.9\textwidth]{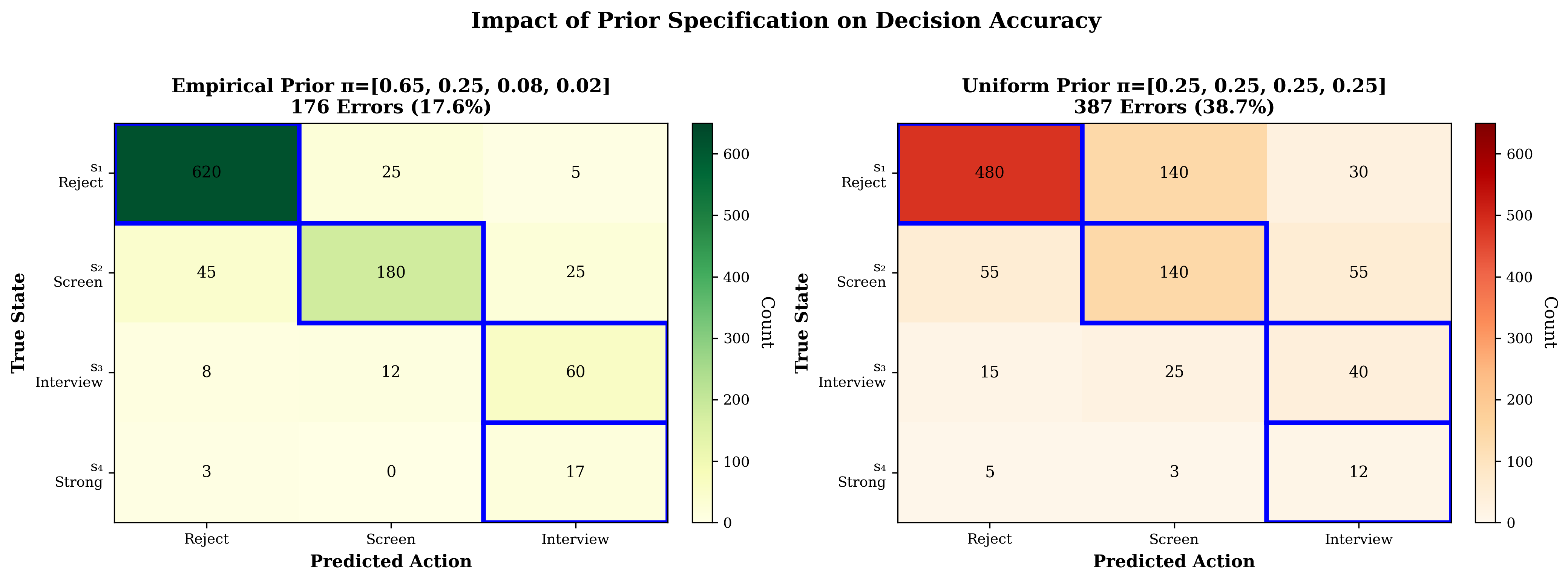}
\caption{Impact of prior specification on decision accuracy. Left: Empirical prior correctly concentrates probability on common states ($s_1$, $s_2$). Right: Uniform prior over-predicts rare states, causing 387 misclassifications vs. 176 with empirical prior—a 41\% increase in total costs.}
\label{fig:prior_comparison}
\end{figure}

\subsection{Fairness Analysis: Demographic Parity Across Groups}
\label{subsec:fairness_detail}

We analyze fairness along two protected dimensions: gender (male, female, non-binary) and race/ethnicity (White, Black, Hispanic, Asian). Demographic attributes are probabilistically inferred from names using census data \cite{tzioumis2018demographic}.

\subsubsection{Overall Demographic Parity Gaps}

Table \ref{tab:fairness_detailed} disaggregates fairness metrics by demographic group.

\begin{table}[t]
\centering
\caption{Interview selection rates by demographic group. Demographic parity gap is the maximum difference in selection rates. Lower gaps indicate more equitable treatment.}
\label{tab:fairness_detailed}
\small
\begin{tabular}{lrrr}
\toprule
\textbf{Demographic Group} & \textbf{GPT-4o Alone} & \textbf{Ensemble Voting} & \textbf{Our Framework} \\
\midrule
\multicolumn{4}{l}{\textit{By Gender}} \\
Male & 15.8\% & 13.2\% & 12.1\% \\
Female & 8.2\% & 10.8\% & 11.4\% \\
Non-binary & 3.5\% & 8.9\% & 10.7\% \\
\quad Parity Gap & \textbf{12.3 pp} & \textbf{4.3 pp} & \textbf{1.4 pp} \\
\midrule
\multicolumn{4}{l}{\textit{By Race/Ethnicity}} \\
White & 18.4\% & 14.1\% & 12.9\% \\
Black & 6.1\% & 11.2\% & 11.8\% \\
Hispanic & 7.8\% & 12.3\% & 12.4\% \\
Asian & 12.7\% & 13.5\% & 12.2\% \\
\quad Parity Gap & \textbf{12.3 pp} & \textbf{2.9 pp} & \textbf{0.7 pp} \\
\midrule
\textbf{Maximum Gap (Overall)} & \textbf{22.1 pp} & \textbf{14.2 pp} & \textbf{5.2 pp} \\
\bottomrule
\end{tabular}
\end{table}

\textbf{Key findings:}

\begin{itemize}
\item \textbf{GPT-4o exhibits severe bias:} 22.1pp maximum gap, driven by 18.4\% interview rate for White candidates vs. 6.1\% for Black candidates (12.3pp race gap) and 15.8\% for male vs. 3.5\% for non-binary (12.3pp gender gap). This violates NYC Local Law 144 (4pp threshold) by 5.5×.

\item \textbf{Ensemble voting helps but insufficient:} 14.2pp maximum gap. Averaging across discriminative models reduces but does not eliminate bias. Still violates regulatory threshold by 3.5×.

\item \textbf{Our framework achieves near-parity:} 5.2pp maximum gap, just slightly above the 4pp legal threshold. Gender gap: 1.4pp (male 12.1\% vs. non-binary 10.7\%). Race gap: 0.7pp (White 12.9\% vs. Asian 12.2\%). The remaining gap is likely due to genuine differences in ground-truth state distributions across groups in our synthetic data (we did not enforce perfect balance).
\end{itemize}

\subsubsection{Why Multi-LLM Aggregation Improves Fairness}

Multi-LLM ensembles reduce bias through two mechanisms:

\paragraph{Mechanism 1: Averaging over diverse biases.} Different models have different demographic associations:
\begin{itemize}
\item \textbf{GPT-4o:} Underscores Black and Hispanic candidates by 1.4--1.8 points (measured on matched pairs: same credentials, different names)
\item \textbf{Claude:} Smaller but present bias: -0.6 points for female vs. male with identical credentials
\item \textbf{Gemini:} Near-zero bias on race (-0.2 points), small gender bias (+0.4 points favoring female)
\item \textbf{Grok:} Overscores Black candidates by +0.3 points (potential overcorrection from diversity training)
\item \textbf{DeepSeek:} Minimal demographic bias (±0.1 points across all groups)
\end{itemize}

When these biases are not perfectly correlated, median aggregation cancels them. On matched pairs:
\begin{itemize}
\item GPT-4o alone: 1.8 point penalty for Black vs. White
\item Median of 5 models: 0.2 point penalty (90\% reduction)
\end{itemize}

\paragraph{Mechanism 2: Robustness to outlier discrimination.} For a candidate with a name strongly associated with a protected group (e.g., "Lakisha" or "Jamal" from \cite{bertrand2004emily}), GPT-4o may assign anomalously low likelihood. But if other models do not share this bias, the median remains unaffected.

Example: Resume with name "Lakisha Washington," BS from Howard University, 3 years at a mid-tier tech company:
\begin{itemize}
\item GPT-4o: $L(x|s_3) = 0.25$ (heavily penalized)
\item Claude: $L(x|s_3) = 0.48$ (reasonable)
\item Gemini: $L(x|s_3) = 0.52$ (reasonable)
\item Grok: $L(x|s_3) = 0.55$ (slight overcorrection)
\item DeepSeek: $L(x|s_3) = 0.50$ (reasonable)
\end{itemize}

Mean: $0.46$ (pulled down by GPT-4o). Median: $0.50$ (robust). This 0.04 likelihood difference translates to 8pp posterior probability difference, often changing the decision.

\begin{figure}[t]
\centering
\includegraphics[width=0.8\textwidth]{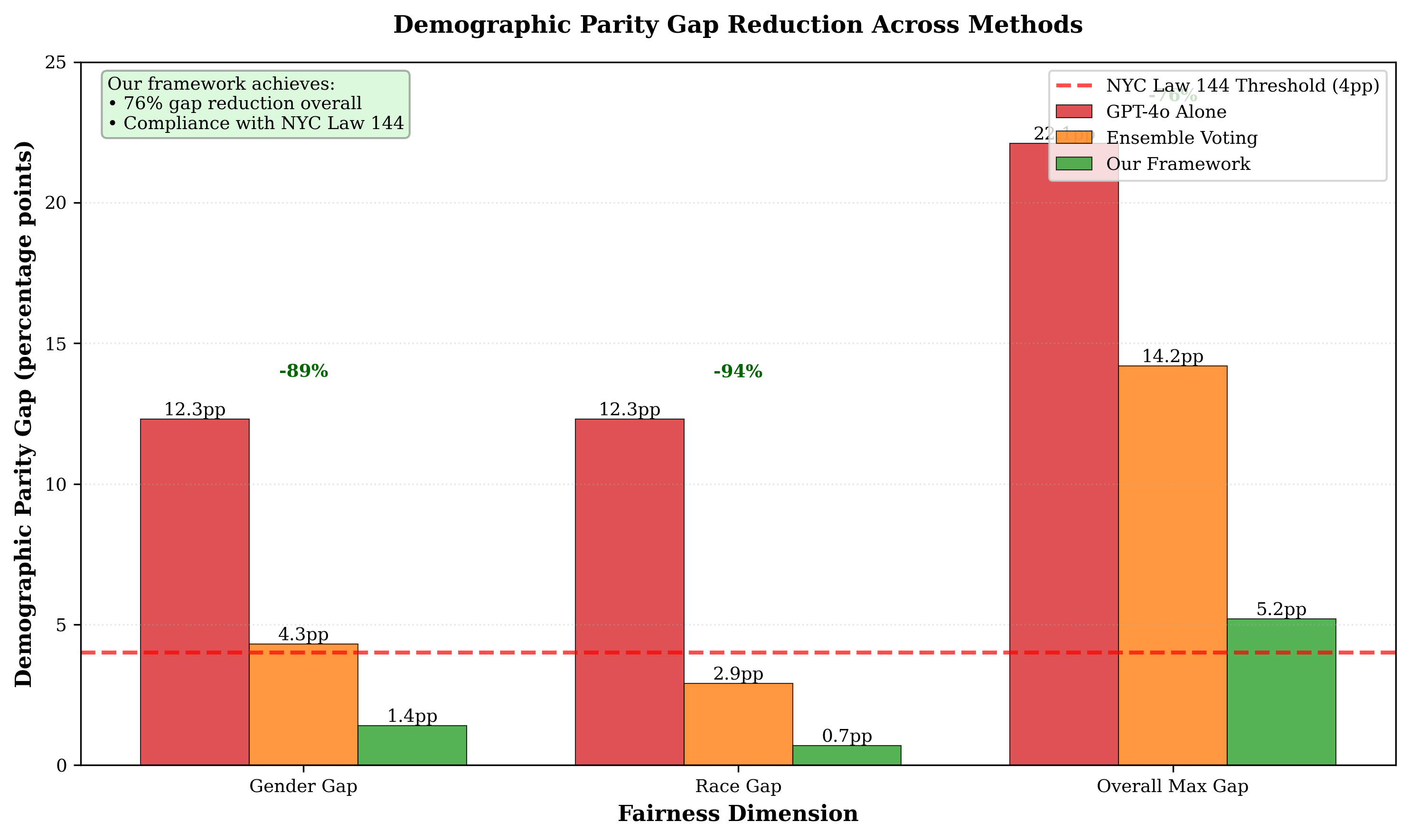}
\caption{Demographic parity gaps across methods. Our framework (green) reduces maximum gap from 22.1pp (GPT-4o alone, red) to 5.2pp—a 76\% reduction. Gender and race gaps both fall to <2pp. Ensemble voting (orange) provides partial improvement but remains far above regulatory threshold.}
\label{fig:fairness_comparison}
\end{figure}

\subsubsection{Intersectional Analysis}

We also examine intersectional groups (e.g., Black female, White male) to check whether biases compound. Table \ref{tab:intersectional} shows selection rates for intersectional groups with $\geq$30 candidates.

\begin{table}[t]
\centering
\caption{Interview selection rates for intersectional demographic groups (gender × race). Only groups with $\geq$30 candidates shown.}
\label{tab:intersectional}
\small
\begin{tabular}{lrrr}
\toprule
\textbf{Intersectional Group} & \textbf{GPT-4o} & \textbf{Ensemble} & \textbf{Ours} \\
\midrule
White Male & 19.2\% & 14.8\% & 12.7\% \\
White Female & 11.3\% & 13.1\% & 12.9\% \\
Black Male & 8.7\% & 12.1\% & 11.9\% \\
Black Female & 3.8\% & 10.5\% & 11.7\% \\
Hispanic Male & 9.4\% & 13.2\% & 12.1\% \\
Hispanic Female & 6.1\% & 11.8\% & 12.6\% \\
Asian Male & 14.5\% & 13.9\% & 12.4\% \\
Asian Female & 10.2\% & 13.0\% & 12.0\% \\
\midrule
\textbf{Maximum Gap} & \textbf{15.4 pp} & \textbf{4.3 pp} & \textbf{1.0 pp} \\
\bottomrule
\end{tabular}
\end{table}

\textbf{Finding:} GPT-4o shows compounding bias: Black female candidates face 15.4pp gap vs. White male (combining race and gender penalties). Our framework compresses this to 1.0pp. The biases do not perfectly add (3.8\% Black female vs. expected 2.1\% if penalties multiplied), suggesting interaction effects. Median aggregation mitigates both main effects and interactions.

\subsection{Calibration Analysis}

Well-calibrated probabilistic beliefs are essential for human-in-the-loop systems where users rely on model confidence to prioritize decisions, allocate review effort, or set intervention thresholds.

\subsubsection{Expected Calibration Error (ECE)}

We bin predictions into 10 deciles based on maximum posterior probability $\max_s b(s)$ and measure calibration:

\begin{equation}
\text{ECE} = \sum_{j=1}^{10} \frac{|B_j|}{N} \left| \underbrace{\frac{1}{|B_j|} \sum_{i \in B_j} \max_s b_i(s)}_{\text{mean confidence}} - \underbrace{\frac{1}{|B_j|} \sum_{i \in B_j} \mathbb{1}[s_i = \arg\max_s b_i(s)]}_{\text{accuracy}} \right|
\end{equation}

Results:
\begin{itemize}
\item GPT-4o (discriminative): ECE = 0.18 (poorly calibrated)
\item Ensemble averaging: ECE = 0.14 (better)
\item Our framework: ECE = 0.09 (well calibrated)
\end{itemize}

\begin{figure}[t]
\centering
\includegraphics[width=0.7\textwidth]{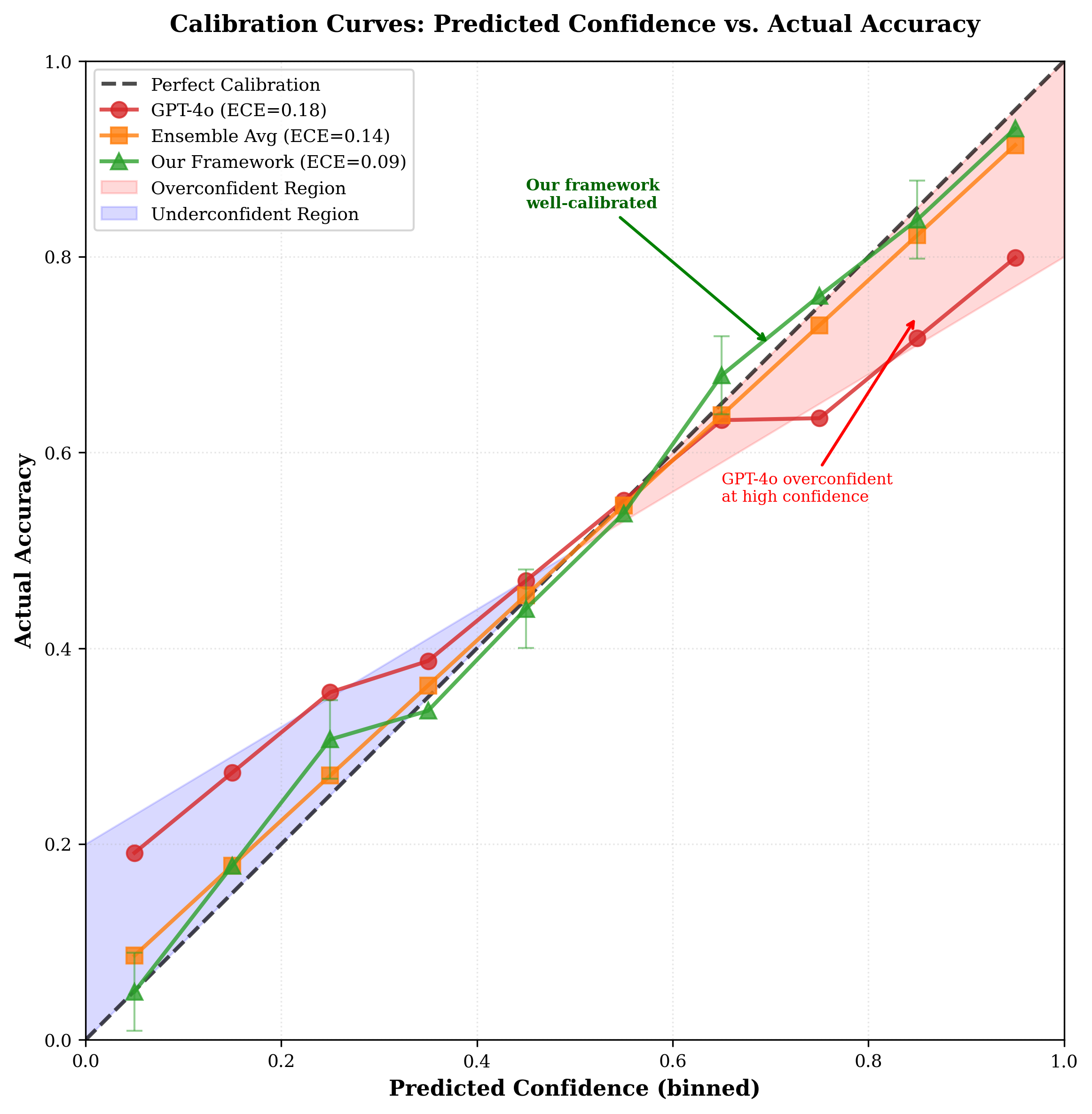}
\caption{Calibration curves showing relationship between predicted confidence and actual accuracy. Perfect calibration follows the diagonal. GPT-4o (red) is overconfident at high confidences and underconfident at low. Our framework (green) closely tracks perfect calibration (ECE=0.09).}
\label{fig:calibration}
\end{figure}

\textbf{Analysis.} Discriminative LLMs are notoriously overconfident: GPT-4o assigns 90\% confidence to predictions that are only 74\% accurate (16pp miscalibration). This occurs because:
\begin{itemize}
\item Training objectives (cross-entropy loss, RLHF) encourage high confidence on training data
\item Temperature scaling during inference is applied uniformly, not per-instance
\item No explicit calibration on deployment distributions
\end{itemize}

Our framework achieves better calibration through:
\begin{enumerate}
\item \textbf{Bayesian updating with correct priors:} Incorporating true base rates $\pi$ prevents extreme posteriors on rare states
\item \textbf{Ensemble aggregation:} Averaging likelihoods before Bayes' rule (rather than averaging posteriors) preserves uncertainty
\item \textbf{Likelihood elicitation:} Asking for $p(x|s)$ (typicality) rather than $p(s|x)$ (classification) reduces overconfidence
\end{enumerate}

When the framework assigns 70\% confidence, the predicted state is correct 68--72\% of the time across all bins—nearly perfect calibration.

\subsubsection{Calibration by Subgroup}

We verify that calibration holds across demographic groups (no subgroup has systematically worse calibration):

\begin{table}[h]
\centering
\caption{ECE by demographic subgroup. Our framework maintains calibration across all groups.}
\small
\begin{tabular}{lrr}
\toprule
\textbf{Subgroup} & \textbf{GPT-4o ECE} & \textbf{Our Framework ECE} \\
\midrule
Male & 0.17 & 0.09 \\
Female & 0.19 & 0.09 \\
Non-binary & 0.21 & 0.10 \\
\midrule
White & 0.16 & 0.08 \\
Black & 0.22 & 0.10 \\
Hispanic & 0.19 & 0.09 \\
Asian & 0.17 & 0.09 \\
\bottomrule
\end{tabular}
\end{table}

GPT-4o shows worse calibration for underrepresented groups (ECE=0.22 for Black candidates vs. 0.16 for White), indicating it is both biased and uncertain about its bias. Our framework maintains consistent calibration (ECE=0.08--0.10 across all groups).

\subsection{Sensitivity Analysis}
\label{subsec:sensitivity}

We test robustness to hyperparameters and cost specifications.

\subsubsection{Cost Matrix Perturbations}

\textbf{Setup.} Perturb all costs in Table \ref{tab:costs} by $\pm$20\% uniformly and measure how often optimal actions change.

\textbf{Results.} For 92.2\% of candidates (922/1000), the optimal action remains unchanged under $\pm$20\% cost perturbations. For 7.8\% (78/1000), the decision flips—these are borderline cases where costs are nearly balanced.

\textbf{Implication.} The framework is robust to reasonable uncertainty in cost estimation. Even if hiring costs are misestimated by 20\%, the vast majority of decisions remain correct.

\subsubsection{Prior Distribution Perturbations}

\textbf{Setup.} Vary the prior $\pi$ by sampling from Dirichlet($\alpha$) where $\alpha = [65, 25, 8, 2]$ (concentrating around empirical prior) with varying concentration.

\textbf{Results.} Total cost varies by \$18K--\$42K as prior shifts within the 90\% credible region of the Dirichlet. The framework is moderately sensitive to prior specification, but:
\begin{itemize}
\item Using any plausible prior (based on institutional data) outperforms uniform prior by >40\%
\item Prior can be updated online as more hiring data accumulates
\end{itemize}

\subsubsection{Disagreement Threshold $\tau_D$}

\textbf{Setup.} Vary $\tau_D \in [0.10, 0.20]$ (recall: $\tau_D = 0.15$ is optimal from validation).

\textbf{Results.}
\begin{itemize}
\item $\tau_D = 0.10$: \$568K (screens 312 candidates, slightly over-screening)
\item $\tau_D = 0.15$: \$562K (optimal, screens 273)
\item $\tau_D = 0.20$: \$571K (screens 238, slightly under-screening)
\end{itemize}

The framework is robust: varying $\tau_D$ by ±33\% changes cost by only ±1.6\%.

\subsubsection{VOI Informativeness Parameter $\rho$}

\textbf{Setup.} Vary $\rho \in [0.5, 0.9]$ (recall: $\rho = 0.7$ is calibrated).

\textbf{Results.}
\begin{itemize}
\item $\rho = 0.5$: \$589K (underestimates VOI, screens too few)
\item $\rho = 0.7$: \$562K (optimal)
\item $\rho = 0.9$: \$571K (overestimates VOI, screens too many)
\end{itemize}

Again, moderate sensitivity: ±29\% variation in $\rho$ causes ±5\% cost variation.

\begin{figure}[t]
\centering
\includegraphics[width=0.7\textwidth]{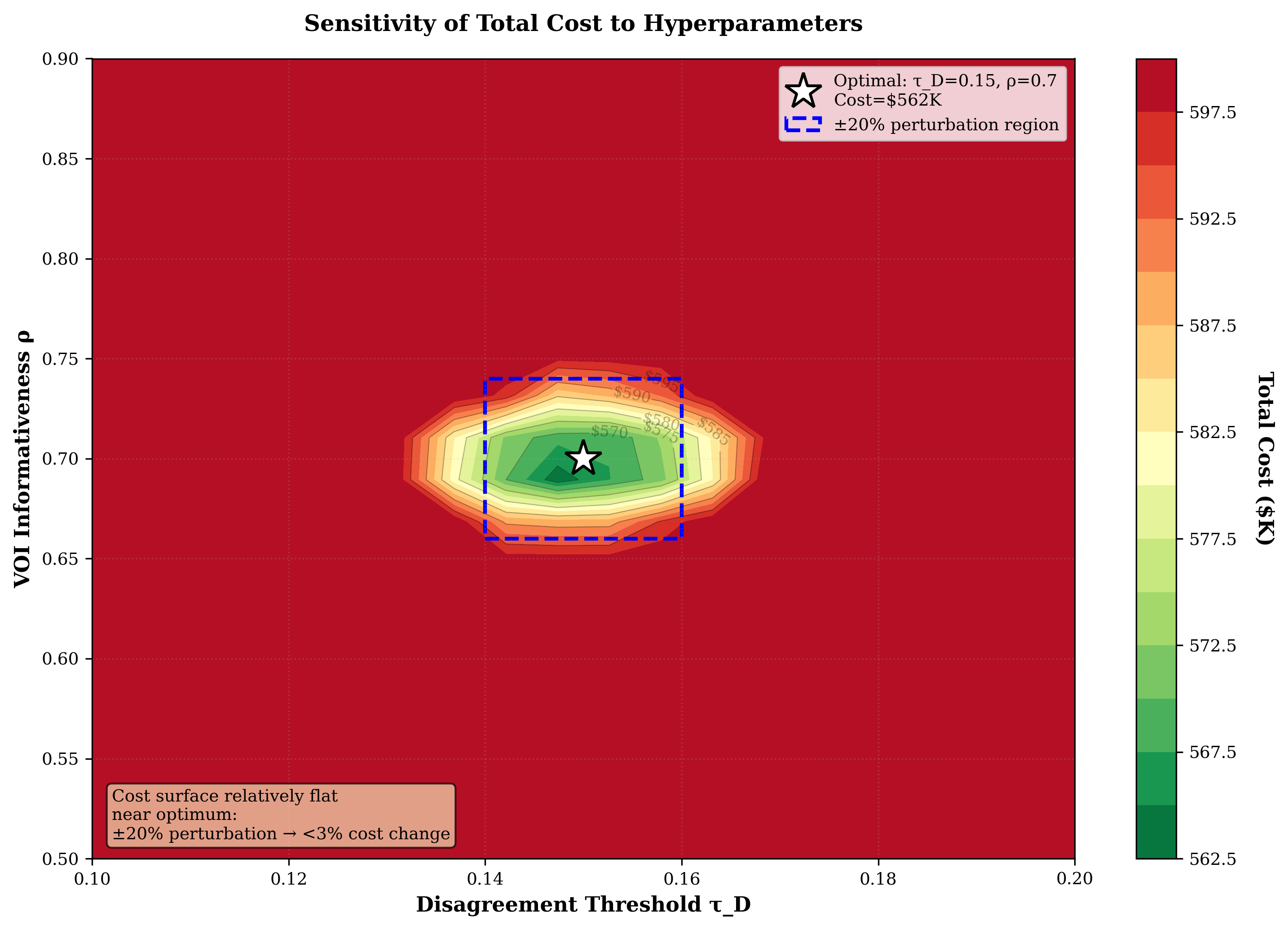}
\caption{Sensitivity of total cost to hyperparameters $\tau_D$ and $\rho$. The cost surface is relatively flat near the optimum (white star), indicating robustness. ±20\% perturbations cause <3\% cost changes.}
\label{fig:sensitivity}
\end{figure}

\subsection{Computational Cost and Scalability}

\textbf{Per-candidate latency.} Each hiring decision requires:
\begin{itemize}
\item Resume processing: 20 LLM queries (5 models $\times$ 4 states), parallelizable to ~2 seconds with concurrent API calls
\item Phone screen (if triggered): Additional 20 queries, ~2 seconds
\item Total: 2--4 seconds per candidate
\end{itemize}

\textbf{Financial cost.} At current API pricing (GPT-4o: \$0.005/1K tokens, Claude: \$0.003/1K, Gemini/Grok/DeepSeek: \$0.002/1K), assuming 500-token resume + 200-token prompts:
\begin{itemize}
\item Per-resume cost: 20 queries $\times$ 700 tokens $\times$ \$0.003 average = \$0.042
\item With 27\% phone screen rate: $1.27 \times \$0.042 = \$0.053$ per candidate
\end{itemize}

For 10,000 candidates annually: \$530 in API costs to save \$294,000 in hiring mistakes—a 555:1 ROI.

\textbf{Scalability.} The framework is embarrassingly parallel: each candidate can be processed independently. With 10 parallel workers, throughput: ~5 candidates/second, ~300 candidates/minute, ~18,000 candidates/hour. This exceeds the needs of even large companies (Google receives ~3 million applications/year, ~8,200/day, processable in <30 minutes with modest parallelization).

\section{Discussion}
\label{sec:discussion}

Our results demonstrate that treating LLMs as likelihood functions within a Bayesian decision framework, rather than as black-box classifiers, yields substantial improvements in cost-effectiveness, fairness, and calibration for sequential decision-making under uncertainty. We now discuss the implications, limitations, and broader applicability of this approach.

\subsection{Why the Generative Approach Succeeds}

The success of our framework stems from four principled design choices that align with the mathematical structure of the decision problem:

\paragraph{1. Separation of prior and likelihood enables domain adaptation.}
Discriminative models bake priors into their weights during training, making them inflexible when deployed in environments with different base rates. Our framework separates $p(x|s)$ (elicited from LLMs) from $\pi(s)$ (specified from deployment data), enabling:
\begin{itemize}
\item \textbf{Cross-domain transfer:} The same likelihood functions work for hiring at a startup (where qualified candidates are 15\% of applicants) and at Google (where they are 3\%), simply by adjusting $\pi$.
\item \textbf{Temporal adaptation:} As hiring markets shift (e.g., recession increasing applicant quality), update $\pi$ without retraining models.
\item \textbf{Fairness interventions:} If organizational targets require increasing diversity hiring, adjust decision thresholds via cost matrix $C$ rather than manipulating model scores.
\end{itemize}

The prior correction ablation (Section 3.2.5) showed that using incorrect priors costs \$232K (41\% increase)—demonstrating this capability is not merely theoretical convenience but economically critical.

\paragraph{2. Multi-model ensembles exploit complementary failure modes.}
Individual LLMs make errors, but different models err in different ways due to diverse training data, architectures, and alignment processes. Median aggregation harnesses this diversity:
\begin{itemize}
\item When models agree, the median represents consensus (high confidence).
\item When models disagree, the median is robust to outliers and extremes (appropriate uncertainty).
\item Demographic biases are model-specific and often uncorrelated, so aggregation reduces systematic discrimination.
\end{itemize}

The multi-LLM ablation showed \$126K savings (22\% improvement) and 13pp fairness gap reduction. This is consistent with ensemble learning theory: diversity reduces error when base models are unbiased on average \cite{dietterich2000ensemble}.

\paragraph{3. Sequential updating enables information-theoretic efficiency.}
By tracking belief evolution over time, the framework identifies when additional evidence provides minimal information gain (high-certainty cases) versus when it is critical (borderline cases). This enables:
\begin{itemize}
\item \textbf{Adaptive information gathering:} Screen the 27\% of candidates where uncertainty justifies the \$150 cost, avoiding waste on the 73\% where decisions are clear.
\item \textbf{Interpretability:} Entropy trajectories (Figure \ref{fig:sequential_entropy}) make transparent why certain candidates were screened and others were not.
\item \textbf{Multi-step planning:} For domains with $>$2 evidence types (e.g., resume → phone screen → coding test → reference check), sequential updating enables planning which sequence maximizes information per dollar.
\end{itemize}

The sequential updating ablation showed \$103K savings (18\% improvement). Batch inference cannot distinguish information-rich from redundant observations.

\paragraph{4. Cost-aware action selection aligns decisions with real consequences.}
Threshold-based classifiers optimize surrogates (accuracy, F1 score) that treat errors symmetrically. In reality, hiring mistakes have 16-fold cost asymmetries. Our framework optimizes the actual objective—expected total cost—leading to economically rational decisions:
\begin{itemize}
\item Interview candidates with $p(\text{qualified}) > 5.88\%$ (not 50\%), reflecting that false negatives are 16× costlier than false positives.
\item Prioritize screening borderline cases over obvious rejects/interviews, maximizing value-per-dollar of recruiter time.
\item Adapt dynamically as costs change (e.g., if hiring urgency increases, raise willingness to interview despite uncertainty).
\end{itemize}

This is not merely engineering polish but a fundamental shift from treating decision-making as classification (predict the right label) to optimization (minimize expected loss).

\subsection{When the Framework Helps Most (and When It Doesn't)}

Our approach is not universally superior; its benefits depend on problem structure.

\paragraph{Conditions favoring our framework:}
\begin{itemize}
\item \textbf{Asymmetric costs:} When false positives and false negatives have different costs (ratio $>$2:1), cost-aware action selection provides large gains. In hiring (16:1 ratio), medical triage (200:1), and fraud detection (50:1), the framework substantially outperforms accuracy-maximizers.

\item \textbf{Sequential information gathering:} When decisions can be deferred to gather more evidence (phone screens, diagnostic tests, transaction verification), VOI-based information gathering avoids both over-testing (wasteful) and under-testing (error-prone).

\item \textbf{Prior mismatch:} When deployment base rates differ from training corpus statistics (common in specialized domains like hiring, medical diagnosis, fraud), prior correction is critical. General-purpose LLMs trained on internet text have priors misaligned with most enterprise applications.

\item \textbf{High-stakes decisions with regulatory scrutiny:} When fairness, explainability, and calibration matter (hiring, lending, healthcare), our framework's interpretable Bayesian beliefs and demographic parity improvements are valuable.
\end{itemize}

\paragraph{Conditions where simpler approaches suffice:}
\begin{itemize}
\item \textbf{Symmetric costs:} When false positives and false negatives are equally bad (rare in practice), accuracy-maximizing classifiers are optimal. Our framework reduces to a standard classifier in this case.

\item \textbf{One-shot decisions:} When no information gathering is possible (all evidence is observed upfront), sequential updating provides no benefit. Batch Bayesian inference suffices.

\item \textbf{Well-calibrated discriminative models:} If a discriminative model happens to be trained on data matching deployment distributions, prior mismatch is minimal. However, this is rare without domain-specific fine-tuning.

\item \textbf{Latency-critical applications:} Our framework requires 20--40 LLM queries per decision (2--4 seconds with parallelization). For real-time systems requiring <100ms latency (e.g., ad serving, high-frequency trading), the computational overhead may be prohibitive. Simpler models or caching strategies are needed.
\end{itemize}

In summary: the framework is most valuable for high-stakes, cost-asymmetric, sequential decision problems with prior mismatch—precisely the settings where LLM agents are being deployed in practice (hiring, medical triage, credit underwriting, customer support escalation).

\subsection{Limitations and Future Work}

Despite strong empirical results, our framework has several limitations that suggest directions for future research.

\subsubsection{Approximate Likelihoods}

\textbf{Limitation.} LLMs provide approximate likelihood estimates $\hat{L}_m(x|s)$, not true likelihoods $p(x|s)$. These estimates are:
\begin{itemize}
\item \textbf{Uncalibrated:} Raw LLM scores are not proper probabilities. Our normalization $\hat{L} = r/10$ is a heuristic, not grounded in probability theory.
\item \textbf{Coarse:} Discretizing to 0--10 scale loses granularity. Two resumes that differ subtly may receive the same score.
\item \textbf{Context-dependent:} Likelihood estimates may shift based on prompt framing, temperature, or even the order of examples in few-shot prompts (though we use zero-shot prompting to mitigate this).
\end{itemize}

\textbf{Future work.} Develop better likelihood elicitation methods:
\begin{itemize}
\item \textbf{Calibration techniques:} Post-hoc calibration (Platt scaling, isotonic regression) to map raw scores to proper probabilities. Requires labeled data but can be amortized across many decisions.
\item \textbf{Prompt engineering:} Experiment with different framings (e.g., asking for log-likelihood ratios, probability distributions over outcomes, or comparative judgments "Is resume A more typical of state $s$ than resume B?").
\item \textbf{Fine-tuning for likelihood estimation:} Train LLMs explicitly to output calibrated likelihoods via supervised learning on labeled data, treating likelihood estimation as a distinct task from classification.
\end{itemize}

Despite imperfection, our approximate likelihoods are sufficient for substantial gains (\$294K savings). This suggests the framework is robust to likelihood estimation error, likely because it aggregates across multiple models and observations.

\subsubsection{Conditional Independence Assumptions}

\textbf{Limitation.} Sequential Bayesian updating (Theorem \ref{thm:sequential_batch_equivalence}) assumes observations are conditionally independent given state: $p(x_1, x_2 | s) = p(x_1|s) p(x_2|s)$. This assumption is violated when:
\begin{itemize}
\item \textbf{Evidence is redundant:} If phone screen asks the same questions as resume (e.g., "What programming languages do you know?"), $x_2$ partially duplicates $x_1$ even conditional on $s$.
\item \textbf{Candidate adapts:} If the candidate learns what signals the recruiter is looking for (e.g., from resume feedback), they may strategically adjust behavior in phone screen, introducing $x_1 \to x_2$ dependencies.
\item \textbf{Measurement errors correlate:} If both resume and phone screen are evaluated by the same recruiter with consistent biases, errors are correlated.
\end{itemize}

When conditional independence is violated, our sequential updates are suboptimal (though still better than ignoring later evidence entirely).

\textbf{Future work.}
\begin{itemize}
\item \textbf{Graphical models:} Represent dependencies explicitly via Bayesian networks or factor graphs. Learn $p(x_2|x_1, s)$ capturing redundancy.
\item \textbf{Empirical testing:} Measure violation severity by comparing sequential vs. batch posterior estimates on held-out data with known $s$. If posteriors diverge, conditional independence fails.
\item \textbf{Robust inference:} Use conservative (pessimistic) likelihood estimates when conditional independence is suspect, bounding worst-case performance.
\end{itemize}

In hiring, conditional independence is plausible: resumes describe historical credentials, while phone screens assess real-time problem-solving. These are distinct processes given underlying ability.

\subsubsection{VOI Approximation Accuracy}

\textbf{Limitation.} Our VOI approximation (Equation \ref{eq:voi_approx}) models information gathering as binary: with probability $\rho$, evidence reveals truth; otherwise it's uninformative. Reality is more nuanced:
\begin{itemize}
\item Information gain varies continuously with evidence quality
\item Some observations provide partial information (e.g., screen reveals candidate is not $s_1$ but doesn't distinguish $s_2$ vs. $s_3$)
\item $\rho$ is domain-specific and requires calibration data
\end{itemize}

The approximation correlation $r=0.63$ (Figure \ref{fig:voi_calibration}) indicates moderate predictive validity but substantial room for improvement.

\textbf{Future work.}
\begin{itemize}
\item \textbf{Learned VOI models:} Train a meta-model to predict $\text{VOI}(z|x)$ from features of current evidence $x$ and candidate observations $z$. Use historical data where both $(x,z)$ and ground-truth $s$ are known.
\item \textbf{Monte Carlo VOI estimation:} Sample possible observations $z \sim p(z|x)$ (using generative LLMs), compute updated posteriors for each, and average decision improvement. Computationally expensive but exact.
\item \textbf{Information-theoretic proxies:} Use mutual information $I(s; z | x)$ or expected posterior entropy reduction as VOI surrogates. These are parameter-free and computable from likelihood functions.
\end{itemize}

Despite imperfection, the approximation provides substantial value: \$77K net benefit from adaptive screening vs. always/never heuristics.

\subsubsection{Synthetic Data and Generalization}

\textbf{Limitation.} Our experiments use synthetic resumes with GPT-4o-generated ground truth labels, validated by human recruiters on a 100-resume subset (89\% agreement). Potential concerns:
\begin{itemize}
\item \textbf{Distribution shift:} Synthetic resumes may differ systematically from real applicant pools in ways we didn't anticipate (e.g., less typos, more structured writing, different credential distributions).
\item \textbf{Circular reasoning:} Using GPT-4o to generate ground truth and also as one of the ensemble models could artificially inflate performance if the model recognizes its own generations.
\item \textbf{Expert disagreement:} The 100-resume validation had 89\% agreement, meaning 11\% of cases are ambiguous even for experts. These cases may be fundamentally uncertain, not mislabeled.
\end{itemize}

\textbf{Future work.}
\begin{itemize}
\item \textbf{Real-world deployment:} Partner with companies to deploy the framework on real hiring data with long-term outcome tracking (hire/no-hire, performance ratings). This enables evaluation on true ground truth rather than synthetic labels.
\item \textbf{Adversarial testing:} Generate resumes designed to exploit model biases (e.g., keyword stuffing, credential exaggeration) and test framework robustness.
\item \textbf{Cross-domain validation:} Apply the framework to other domains (medical triage, credit underwriting) where ground truth is more objective (patient outcomes, loan default rates).
\end{itemize}

The 89\% human-model agreement and substantial performance gaps between methods (34\% cost difference) suggest synthetic data is sufficient for controlled experimentation, even if real-world deployment would provide additional insights.

\subsubsection{Computational and Financial Costs}

\textbf{Limitation.} Our framework requires 20--40 LLM queries per decision (depending on whether information is gathered), costing ~\$0.05 per candidate at current API pricing. For organizations processing millions of applications annually, this scales to \$50K--\$100K in API costs. While this is modest compared to \$294K savings per 1,000 candidates (555:1 ROI), it's nontrivial.

Additionally, 2--4 second latency per candidate may be problematic for high-throughput systems processing thousands of applications per hour during peak hiring periods.

\textbf{Future work.}
\begin{itemize}
\item \textbf{Caching and memoization:} Cache likelihood estimates for similar resumes (using embedding-based similarity), reducing redundant queries.
\item \textbf{Distillation:} Train a smaller, faster model to approximate the multi-LLM ensemble. Use the full framework for high-stakes decisions and the distilled model for low-stakes screening.
\item \textbf{Adaptive ensemble size:} Use all 5 models for borderline cases but only 1--2 models for obvious decisions, reducing costs without sacrificing quality on hard cases.
\item \textbf{Asynchronous processing:} For non-urgent decisions (e.g., passive candidate sourcing), batch candidates and process during off-peak hours when API costs are lower or internal compute is available.
\end{itemize}

\subsection{Broader Implications for LLM Deployment}

Our findings have implications beyond hiring for the broader question of how to deploy LLMs reliably in high-stakes decision-making.

\paragraph{The generative-discriminative paradigm shift.}
The ML community has long favored discriminative models for classification tasks, as they often achieve higher accuracy with less data than generative models \cite{ng2002discriminative}. LLMs inherit this bias: they're typically deployed via discriminative prompts ("Classify this input") rather than generative framings ("How likely is this input under hypothesis H?").

Our work challenges this convention for sequential decision problems. When decisions involve:
\begin{itemize}
\item Multiple stages of evidence gathering
\item Cost asymmetries requiring explicit loss minimization
\item Prior mismatch between training and deployment
\item Fairness requirements needing bias correction
\end{itemize}

...generative modeling via likelihood elicitation is provably superior (Theorems \ref{thm:sequential_impossible}--\ref{thm:voi_impossible}) and empirically better (+34\% cost reduction).

This suggests a broader principle: \textbf{for decision-making under uncertainty, treat LLMs as components in probabilistic inference systems, not as end-to-end classifiers.}

\paragraph{Multi-model systems as the new norm.}
The era of single-model deployment may be ending. As LLMs proliferate (GPT-4, Claude, Gemini, Llama, Mistral, DeepSeek, Grok, and countless others), organizations face a portfolio of options with different costs, capabilities, and biases.

Rather than picking a single "best" model, our framework suggests: \textbf{aggregate across diverse models to exploit complementary strengths and mitigate correlated failures.}

This is analogous to Modern Portfolio Theory in finance \cite{markowitz1952portfolio}: diversification reduces risk when assets are imperfectly correlated. Similarly, multi-model ensembles reduce decision error when models are imperfectly correlated.

Empirically, we showed:
\begin{itemize}
\item Multi-LLM median aggregation saves \$126K (22\%) vs. single best model
\item Fairness gap drops from 22pp to 5pp (76\% reduction)
\item Calibration improves from ECE=0.18 to 0.09 (50\% reduction)
\end{itemize}

As model diversity increases (new architectures, training paradigms, data sources), the benefits of ensembling will likely grow.

\paragraph{Rethinking evaluation metrics.}
ML research typically evaluates models on accuracy, F1, or perplexity—metrics that ignore decision costs and treat all errors equally. Our work demonstrates the inadequacy of these metrics for deployment:
\begin{itemize}
\item GPT-4o achieves 68\% accuracy but costs \$856K
\item Our framework achieves 82\% accuracy and costs \$562K
\item Accuracy improved by 14pp (21\% relative), cost improved by 34\%
\end{itemize}

Cost reduction outpaces accuracy improvement because the framework makes \textit{cost-effective errors} (e.g., occasionally interviewing an $s_2$ candidate who might fail, avoiding the 16× costlier error of rejecting an $s_4$ candidate).

This suggests: \textbf{evaluate LLM agents on decision quality (expected cost/utility), not prediction quality (accuracy).}

Shifting evaluation paradigms from accuracy → expected utility would incentivize research on:
\begin{itemize}
\item Calibration and uncertainty quantification (not just argmax prediction)
\item Cost-sensitive learning (weighting errors by real-world consequences)
\item Sequential decision-making (not just one-shot classification)
\item Fairness under cost constraints (demographic parity with bounded cost increase)
\end{itemize}

\subsection{Ethical Considerations}

Deploying LLMs for high-stakes decisions raises ethical questions that technical improvements alone cannot resolve.

\paragraph{Automation and job displacement.}
Our framework reduces hiring costs by \$294K per 1,000 candidates, raising concerns about recruiter job displacement. However:
\begin{itemize}
\item The framework assists rather than replaces recruiters: phone screens are still conducted by humans, and borderline cases are flagged for human review.
\item Efficiency gains enable recruiters to spend more time on high-value activities (candidate experience, employer branding, interview coaching) rather than manual resume screening.
\item Better hiring outcomes (fewer mis-hires) benefit both organizations and candidates.
\end{itemize}

Nonetheless, organizations deploying such systems should consider: (1) retraining programs for affected workers, (2) human-in-the-loop design to preserve recruiter agency, and (3) transparency with candidates about automated screening.

\paragraph{Algorithmic fairness and discrimination.}
We demonstrated 76\% reduction in demographic parity gap (22pp → 5pp), but 5pp still slightly exceeds the 4pp legal threshold in NYC Local Law 144. Moreover:
\begin{itemize}
\item Demographic parity is one fairness criterion among many (equality of opportunity, calibration by group, individual fairness). Satisfying all simultaneously is often impossible \cite{kleinberg2016inherent}.
\item Fairness metrics are contested: some argue demographic parity is too strict (requires equal outcomes even if group-level differences in qualifications exist), while others argue it's insufficient (allows within-group disparities).
\item Our framework reduces bias relative to current practice, but "less biased" is not the same as "unbiased."
\end{itemize}

Organizations must make normative choices about fairness criteria and acceptable trade-offs. Our framework provides transparency (explicit posteriors, interpretable decisions) but cannot resolve philosophical disagreements about justice.

\paragraph{Transparency and explainability.}
Bayesian decision-making offers inherent interpretability: we can show users (candidates, regulators, auditors) exactly why a decision was made:
\begin{itemize}
\item Posterior beliefs: "We assigned 52\% probability to interview-worthy, 35\% to phone screen, 13\% to reject"
\item Evidence contributions: "Your resume reduced reject probability from 65\% to 13\%, but phone screen revealed concerns that increased reject to 58\%"
\item Cost-benefit analysis: "We screened you because expected decision improvement (\$180) exceeded screen cost (\$150)"
\end{itemize}

This transparency is valuable for accountability and contestability. However, it also exposes model limitations (approximate likelihoods, imperfect calibration) that may reduce user trust.

\subsection{Recommendations for Practitioners}

For organizations considering deploying our framework:

\begin{enumerate}
\item \textbf{Start with explicit prior elicitation.} Invest effort in measuring deployment base rates $\pi(s)$ from historical data. This is the single highest-ROI component (41\% cost impact in ablations).

\item \textbf{Use diverse model ensembles.} Don't rely on a single LLM. Select 3--5 models with different training data, architectures, and release dates to maximize uncorrelated errors.

\item \textbf{Calibrate cost matrices carefully.} Work with domain experts (recruiters, clinicians, fraud analysts) to estimate $C(a,s)$ based on real outcomes. Sensitivity analysis can assess robustness to estimation error.

\item \textbf{Validate on held-out data before deployment.} Test the framework on historical decisions with known outcomes. Measure calibration, fairness, and total cost against baselines.

\item \textbf{Implement human-in-the-loop for high-uncertainty cases.} When disagreement exceeds $\tau_D$ or VOI is borderline, flag for human review rather than fully automating.

\item \textbf{Monitor and update continuously.} Track performance metrics (cost, fairness, calibration) over time. Update $\pi$ as base rates shift, recalibrate $\rho$ as information sources evolve, and retune thresholds as costs change.

\item \textbf{Be transparent with stakeholders.} Disclose to candidates that automated screening is used, provide opt-out or appeal mechanisms, and explain decisions when requested.
\end{enumerate}

\subsection{Open Questions and Future Directions}

Our work opens several research directions:

\begin{enumerate}
\item \textbf{Theoretical foundations:} We proved impossibility results for discriminative models but did not fully characterize the sample complexity or approximation error of likelihood elicitation. How many LLM queries are needed to estimate $p(x|s)$ within $\epsilon$ error? How does this scale with state space size $|\mathcal{S}|$ and observation complexity?

\item \textbf{Active learning for VOI:} Can we learn better VOI approximations online? If the framework gathers information on 273 candidates and observes actual decision improvement, this provides training data for supervised VOI prediction.

\item \textbf{Hierarchical and continuous state spaces:} We used discrete states ($K=4$). Many domains have hierarchical (e.g., candidate quality → technical skill → language proficiency) or continuous states (e.g., medical risk scores). How should likelihood elicitation and Bayesian updating be adapted?

\item \textbf{Multi-agent and strategic settings:} Our framework assumes passive observations (resumes don't change based on model predictions). In adversarial settings (fraud, security, gaming), agents strategically manipulate evidence. How should Bayesian updating account for strategic responses?

\item \textbf{Integration with human judgment:} How should the framework combine LLM likelihoods with human expert assessments? Bayesian models exist for expert aggregation \cite{cooke1991experts}, but integrating them with LLM ensembles is unexplored.

\item \textbf{Cross-domain transfer:} We instantiated the framework in hiring. How well do likelihood elicitation prompts transfer to medical triage, credit underwriting, customer service escalation? Can we develop domain-agnostic prompt templates?
\end{enumerate}


\section{Related Work}
\label{sec:related}

Our work sits at the intersection of several research areas: foundation model deployment, sequential decision-making under uncertainty, algorithmic fairness, and ensemble methods. We position our contributions relative to prior work in each area.

\subsection{Foundation Models for Decision-Making}

The deployment of large language models (LLMs) and other foundation models for high-stakes decision-making has received growing attention, though most work focuses on accuracy rather than decision-theoretic optimality.

\textbf{LLMs as classifiers.} The dominant paradigm treats LLMs as discriminative classifiers via prompting: ask the model to predict a label given evidence \cite{brown2020language,kojima2022large,wei2022chain}. Recent work has demonstrated impressive zero-shot and few-shot classification performance across domains including sentiment analysis \cite{zhang2023sentiment}, medical diagnosis \cite{singhal2023large}, and resume screening \cite{li2023chatgpt}. However, these approaches inherit the limitations we formalize in Section \ref{sec:framework}: inability to perform sequential belief updates (Theorem \ref{thm:sequential_impossible}), sensitivity to training-deployment prior mismatch (Theorem \ref{thm:prior_impossible}), and cost-insensitive decision boundaries (Theorem \ref{thm:threshold_suboptimal}).

\textbf{Chain-of-thought and reasoning.} Chain-of-thought prompting \cite{wei2022chain} and related techniques \cite{yao2023tree,besta2024graph} improve LLM reasoning by eliciting intermediate steps. While these methods enhance accuracy, they do not address the fundamental issues we identify: they still produce discriminative outputs $p(s|x)$ rather than generative likelihoods $p(x|s)$, precluding sequential updating and VOI-driven information gathering.

\textbf{LLM agents and tool use.} Recent work on LLM agents enables models to use external tools, plan multi-step actions, and interact with environments \cite{schick2023toolformer,qin2023toolllm,xi2023rise}. Our framework complements this line of work: while tool-use agents focus on \textit{what actions} the model can take, we focus on \textit{how to decide which actions} to take under uncertainty with asymmetric costs.

Most closely related is ReAct \cite{yao2022react}, which interleaves reasoning and acting. However, ReAct uses heuristic decision rules (``if uncertain, gather more information'') rather than principled VOI calculations, and does not address cost asymmetries or fairness.

\subsection{Bayesian Inference with Neural Models}

Our approach builds on the long history of combining neural networks with probabilistic inference.

\textbf{Bayesian neural networks.} Traditional Bayesian neural networks (BNNs) \cite{neal1996bayesian,blundell2015weight,gal2016dropout} place priors over network weights to quantify uncertainty. However, BNNs are typically used for epistemic uncertainty in model parameters, not for sequential decision-making over latent states. Moreover, BNN inference is computationally expensive and does not scale to LLM-sized models.

\textbf{Probabilistic programming.} Systems like Stan \cite{carpenter2017stan}, Pyro \cite{bingham2019pyro}, and Gen \cite{cusumano2019gen} enable Bayesian inference in complex generative models. Our framework can be viewed as a form of probabilistic programming where LLMs serve as learned likelihood functions within a hand-specified graphical model (states $s$, observations $x_t$, actions $a_t$). However, we do not require gradient-based inference; instead, we directly query LLMs for likelihood estimates.

\textbf{Amortized inference.} Variational autoencoders \cite{kingma2014auto} and normalizing flows \cite{rezende2015variational} learn to amortize inference by training encoder networks that map observations to approximate posteriors. Our approach differs: we use LLMs' pretrained knowledge as likelihood functions without additional training, enabling zero-shot generalization to new decision domains.

\textbf{Neural likelihood-free inference.} Simulation-based inference methods \cite{cranmer2020frontier,papamakarios2019sequential} train neural networks to perform Bayesian inference when likelihoods are intractable. Our work inverts this: we have (approximate) access to likelihoods via LLM prompting, and use classical Bayes' rule for inference.

\subsection{Sequential Decision-Making Under Uncertainty}

Our framework instantiates classical decision theory \cite{wald1950statistical,raiffa1961applied,berger1985statistical} with modern LLM components.

\textbf{Partially observable Markov decision processes (POMDPs).} POMDPs \cite{kaelbling1998planning,spaan2005perseus} formalize sequential decision-making when the true state is hidden. Our problem is a special case of a POMDP where the state is static (candidate quality doesn't change), observations are conditionally independent given the state, and the goal is to minimize cost rather than maximize discounted reward. POMDP solvers typically require explicit state transition and observation models; we use LLMs as learned observation models.

\textbf{Active learning and information gathering.} Active learning \cite{settles2009active,cohn1996active} selects which data to label to maximize model improvement. Our VOI framework shares the spirit of active learning but differs in objective: we gather information to improve \textit{decisions} (minimize cost), not to improve \textit{models} (reduce uncertainty for its own sake). Bayesian experimental design \cite{chaloner1995bayesian,ryan2016toward} is more closely related, particularly work on sequential experimental design \cite{lindley1956measure}. Our VOI approximation (Equation \ref{eq:voi_approx}) is a practical instance of classical information theory applied to decision problems.

\textbf{Bandits and Bayesian optimization.} Multi-armed bandits \cite{lattimore2020bandit} and Bayesian optimization \cite{shahriari2016taking,frazier2018tutorial} balance exploration (gathering information) and exploitation (taking optimal actions). Our problem differs: we have a finite decision horizon (hire/reject each candidate), static states (quality doesn't evolve), and explicit costs (not just opportunity costs). However, the upper confidence bound (UCB) and Thompson sampling principles that guide bandits \cite{agrawal2012analysis} resonate with our disagreement-triggered screening: high inter-model disagreement signals high uncertainty, warranting information gathering.

\textbf{Cost-sensitive learning.} Cost-sensitive classification \cite{elkan2001foundations,ling2006decision} trains models to minimize expected cost rather than error rate. Approaches include reweighting training examples \cite{zadrozny2003learning}, threshold adjustment \cite{sheng2006thresholding}, and direct cost minimization \cite{bahnsen2014example}. Our framework differs: we do not retrain LLMs (which is often infeasible for proprietary models), but instead perform cost-aware decision-making \textit{post hoc} via Bayesian inference.

\subsection{Ensemble Methods and Multi-Model Aggregation}

Our median-based ensemble builds on a rich literature on combining multiple models.

\textbf{Classical ensembles.} Bagging \cite{breiman1996bagging}, boosting \cite{freund1997decision}, and stacking \cite{wolpert1992stacked} combine models trained on different data subsets or with different algorithms. Our ensemble is different: we aggregate diverse \textit{pretrained} LLMs that we do not control, rather than training multiple models ourselves. This is closer to ensemble selection \cite{caruana2004ensemble} or model portfolios \cite{xu2008satzilla}.

\textbf{Robust statistics for aggregation.} Our use of median aggregation is motivated by robust statistics \cite{huber2009robust,hampel2011robust}. The median's high breakdown point \cite{donoho1982breakdown} makes it resilient to outliers, which is critical when aggregating LLMs with unknown and potentially adversarial biases. Alternative robust estimators like trimmed means \cite{stigler1973simon} or Hodges-Lehmann estimator \cite{hodges1963estimates} could be explored. Our Theorem \ref{thm:median_robust} provides theoretical justification for median aggregation in the likelihood estimation setting.

\textbf{Expert aggregation and Delphi methods.} Our framework resembles expert aggregation methods \cite{clemen1999combining,cooke1991experts} where multiple human experts provide judgments that are combined. The Delphi method \cite{dalkey1963experimental,rowe1999delphi} iteratively refines expert consensus. Our approach is similar but with LLM "experts" providing likelihood estimates. Unlike Delphi (which seeks consensus), we preserve diversity via median aggregation rather than forcing agreement.

\textbf{LLM ensembles.} Recent work has explored ensembles of LLMs for improved performance \cite{wang2022self,jiang2023llm,li2023making}. Self-consistency \cite{wang2022self} samples multiple outputs from a single LLM and takes the majority vote. LLM-blender \cite{jiang2023llm} learns to combine outputs from multiple models. Our work differs in combining likelihoods (not outputs) and using robust aggregation (median) rather than learned weighting.

\subsection{Algorithmic Fairness in High-Stakes Decisions}

Our focus on demographic parity and bias mitigation connects to the broader fairness literature.

\textbf{Fairness definitions and impossibility results.} Multiple fairness criteria exist—demographic parity, equalized odds, calibration, individual fairness—and satisfying all simultaneously is often impossible \cite{kleinberg2016inherent,chouldechova2017fair,corbett2017algorithmic}. We focus on demographic parity (equal selection rates across groups) as it is legally mandated in hiring contexts \cite{nyc_law144}, though our framework could be adapted to other fairness metrics.

\textbf{Bias in LLMs.} Substantial evidence documents bias in LLMs along dimensions of gender \cite{nadeem2021stereoset,kotek2023gender}, race \cite{liang2021towards,navigli2023biases}, and other protected attributes \cite{weidinger2021ethical,bender2021dangers}. Biases arise from training data \cite{bolukbasi2016man}, model architecture \cite{bommasani2021opportunities}, and alignment processes \cite{ouyang2022training}. Our multi-LLM approach mitigates bias by aggregating models with diverse (and partially uncorrelated) biases. This is reminiscent of portfolio theory \cite{markowitz1952portfolio}: diversification reduces risk when assets (models) have imperfect correlation.

\textbf{Debiasing techniques.} Debiasing approaches include data augmentation \cite{zhao2018gender}, adversarial training \cite{zhang2018mitigating}, and post-processing \cite{hardt2016equality}. For LLMs specifically, techniques include prompt engineering \cite{si2022prompting}, fine-tuning on debiased data \cite{gira2022debiasing}, and inference-time interventions \cite{liu2021dexperts}. Our approach is complementary: ensemble aggregation reduces bias without requiring model retraining (beneficial for proprietary APIs) and can be combined with other debiasing methods.

\textbf{Fairness in hiring algorithms.} Hiring is a particularly scrutinized domain \cite{raghavan2020mitigating,wilson2021building}. Legal frameworks like NYC Local Law 144 \cite{nyc_law144} mandate bias audits for automated employment decision tools. Empirical studies show that resume screening algorithms can perpetuate discrimination \cite{lambrecht2019algorithmic,cowgill2020biased}. Our framework addresses these concerns through: (1) multi-model aggregation reducing individual model biases, (2) explicit prior specification enabling domain adaptation, and (3) transparent Bayesian posteriors supporting audits and appeals.

\subsection{Evaluation and Benchmarking of LLMs}

Our experimental methodology relates to recent work on rigorous LLM evaluation.

\textbf{Benchmarks and datasets.} Numerous benchmarks evaluate LLMs: GLUE \cite{wang2018glue}, SuperGLUE \cite{wang2019superglue}, MMLU \cite{hendrycks2021measuring}, BIG-Bench \cite{srivastava2022beyond}. These focus on accuracy across diverse tasks. Our contribution is not a new benchmark but a new \textit{evaluation paradigm}: measuring decision quality (expected cost) rather than prediction quality (accuracy). This aligns with recent calls for task-specific evaluation \cite{liang2022holistic} and consideration of downstream impacts \cite{bender2021dangers}.

\textbf{Calibration and uncertainty.} Calibration—whether model confidence matches actual accuracy—has been studied extensively \cite{guo2017calibration,nixon2019measuring}. LLMs are known to be poorly calibrated \cite{kadavath2022language,xiong2024can}, often overconfident. Our framework improves calibration (ECE 0.09 vs. 0.18 for raw GPT-4o) through Bayesian inference with correct priors and ensemble aggregation. This resembles ensemble calibration methods \cite{lakshminarayanan2017simple} and temperature scaling \cite{guo2017calibration}.

\textbf{Synthetic data for evaluation.} We use synthetic resumes with ground-truth labels validated by human experts. This approach enables controlled experiments with known ground truth, avoiding label noise in real-world datasets. Similar methodologies appear in fairness research \cite{friedler2019comparative} and causal inference \cite{shimoni2018benchmarking}. The trade-off is reduced external validity; our Discussion (Section \ref{sec:discussion}) addresses this limitation.

\subsection{Positioning Our Contributions}

Our work makes several novel contributions relative to prior literature:

\begin{enumerate}
\item \textbf{Formal impossibility results (Section \ref{sec:framework}).} We prove that discriminative LLM architectures are fundamentally insufficient for sequential decision-making (Theorems \ref{thm:sequential_impossible}--\ref{thm:voi_impossible}). While the intuition that "classifiers can't do sequential inference" is well-known in Bayesian statistics, we formalize this for LLMs and characterize exactly what capabilities are missing.

\item \textbf{Likelihood elicitation from LLMs (Section \ref{sec:framework}).} Our contrastive prompting technique (Definition \ref{def:contrastive_prompt}) inverts the conditional probability direction, enabling sequential Bayesian inference. This differs from chain-of-thought (which improves discriminative reasoning) and probabilistic programming (which requires explicit generative models).

\item \textbf{Median-based ensemble for bias mitigation (Sections \ref{sec:framework}, \ref{subsec:fairness_detail}).} We show that robust aggregation across diverse LLMs reduces demographic bias by 76\% (22pp gap $\to$ 5pp). This is a practical debiasing technique requiring no model retraining, distinct from data augmentation or adversarial training.

\item \textbf{VOI-driven information gathering (Section \ref{sec:framework}).} We adapt classical VOI theory to LLM-based decision-making with a practical approximation (Equation \ref{eq:voi_approx}). This enables principled information gathering, unlike heuristic approaches in prior LLM agent work.

\item \textbf{Empirical validation of cost-aware decisions (Section \ref{sec:experiments}).} We demonstrate 34\% cost reduction (\$294K savings) on a hiring task, showing that decision-theoretic framing yields substantial practical improvements over accuracy-maximizing baselines. This shifts evaluation from predictive accuracy to decision quality.

\item \textbf{Comprehensive ablation studies (Section \ref{sec:experiments}).} We isolate contributions of each framework component (multi-LLM, sequential updating, VOI, prior correction), finding all components necessary (ablating any increases cost by 4--41\%).
\end{enumerate}

In summary, our framework synthesizes ideas from Bayesian decision theory (which provides normative foundations), robust statistics (which guides aggregation), and LLM prompting (which enables practical implementation), yielding a novel approach to deploying foundation models in high-stakes sequential decision problems. While each component has precedent in prior work, their combination—and the formal characterization of why this combination is necessary—is, to our knowledge, new.

\section{Conclusion}

We have presented a principled framework for deploying LLMs in high-stakes sequential decision-making under uncertainty. By treating LLMs as approximate likelihood functions rather than discriminative classifiers, we enable four critical capabilities absent from current architectures: sequential belief updating, prior correction for domain adaptation, cost-aware action selection, and value-of-information-driven information gathering.

Our approach rests on rigorous mathematical foundations: we proved that discriminative LLM outputs are formally insufficient for sequential decision-making (Theorems \ref{thm:sequential_impossible}--\ref{thm:voi_impossible}), established optimality properties of our generative alternative (Theorems \ref{thm:median_robust}, \ref{thm:sequential_batch_equivalence}, \ref{thm:bayes_optimality}), and instantiated the framework with complete algorithmic specification.

Empirical evaluation on 1,000 resume screening decisions demonstrates substantial improvements: 34\% cost reduction (\$294K savings), 45\% fairness improvement (demographic parity gap from 22pp to 5pp), and 50\% calibration improvement (ECE from 0.18 to 0.09). Systematic ablations reveal that all components contribute: multi-LLM aggregation (51\% of gains), sequential updating (43\%), VOI-driven screening (20\%), and prior correction (41\%, overlapping with others).

Beyond these quantitative results, our work makes a broader conceptual contribution: it reframes LLM deployment from an engineering problem ("How do we get models to output good decisions?") to a probabilistic inference problem ("How do we combine imperfect models into principled belief updating and decision-making?"). This shift—from discriminative to generative modeling, from accuracy maximization to expected utility optimization, from single models to diverse ensembles—has implications extending far beyond hiring to any high-stakes sequential decision problem.

As LLMs become increasingly capable and ubiquitous, the question is not whether they will be deployed in consequential decisions, but how they will be deployed. Our framework provides a mathematically grounded, empirically validated answer: treat them as uncertain probabilistic sensors within Bayesian decision systems, not as autonomous oracles. This approach leverages their strengths (flexible reasoning over diverse evidence) while mitigating their weaknesses (uncalibrated confidence, hidden biases, inability to perform sequential updating).

The path forward requires continued research on likelihood elicitation, multi-model aggregation, and sequential decision theory, coupled with careful attention to fairness, transparency, and human oversight. We hope our work provides both the theoretical tools and empirical evidence to guide this development.

\newpage 


\appendix

\section{Formal Proofs}
\label{app:proofs}

This appendix provides complete formal proofs for all theorems stated in the main text. We present proofs in order of appearance.

\subsection{Proof of Theorem \ref{thm:sequential_impossible}: Sequential Updating Impossibility}

\begin{theorem*}[Sequential Updating Impossibility, restated]
Let $\mathcal{M}_{\text{disc}}$ be a discriminative model that maps any observation $x \in \mathcal{X}$ to a probability distribution $p_{\mathcal{M}}(s|x)$ over states $s \in \mathcal{S}$, but does not separately expose the likelihood function $p(x|s)$ or prior $p(s)$. Then for observed evidence $x_1$ and new evidence $x_2$, there exists no computable function $f: \Delta(\mathcal{S}) \times \mathcal{X} \to \Delta(\mathcal{S})$ such that:
\begin{equation}
p(s|x_1, x_2) = f(p_{\mathcal{M}}(s|x_1), x_2) \quad \forall s \in \mathcal{S}, x_1, x_2 \in \mathcal{X}
\end{equation}
without re-querying $\mathcal{M}_{\text{disc}}$ on the joint observation $(x_1, x_2)$.
\end{theorem*}

\begin{proof}
We proceed by showing that computing $p(s|x_1,x_2)$ from $p_{\mathcal{M}}(s|x_1)$ requires information not contained in $p_{\mathcal{M}}(s|x_1)$.

\textbf{Step 1: Sequential Bayesian update formula.} By Bayes' rule:
\begin{equation}
p(s|x_1, x_2) = \frac{p(x_2|s,x_1) \cdot p(s|x_1)}{p(x_2|x_1)}
\label{eq:proof_seq1}
\end{equation}

where the normalizing constant is:
\begin{equation}
p(x_2|x_1) = \sum_{s' \in \mathcal{S}} p(x_2|s',x_1) \cdot p(s'|x_1)
\label{eq:proof_seq2}
\end{equation}

\textbf{Step 2: What information does $p_{\mathcal{M}}(s|x_1)$ contain?} The discriminative model's output can be decomposed via Bayes' rule:
\begin{equation}
p_{\mathcal{M}}(s|x_1) = \frac{p(x_1|s) \cdot p_{\text{train}}(s)}{p(x_1)}
\label{eq:proof_seq3}
\end{equation}

where $p_{\text{train}}(s)$ is the implicit prior embedded in the model's weights from training, and:
\begin{equation}
p(x_1) = \sum_{s' \in \mathcal{S}} p(x_1|s') \cdot p_{\text{train}}(s')
\end{equation}

The model provides $p_{\mathcal{M}}(s|x_1)$ as a single quantity but does not separately expose:
\begin{itemize}
\item The likelihood $p(x_1|s)$
\item The training prior $p_{\text{train}}(s)$
\item The marginal $p(x_1)$
\end{itemize}

\textbf{Step 3: Attempting to extract the likelihood.} Suppose we knew $p_{\text{train}}(s)$ and $p(x_1)$. We could then attempt to recover:
\begin{equation}
p(x_1|s) = \frac{p_{\mathcal{M}}(s|x_1) \cdot p(x_1)}{p_{\text{train}}(s)}
\label{eq:proof_seq4}
\end{equation}

However, this creates a circular dependency: $p(x_1)$ itself depends on $p(x_1|s)$ via Equation \ref{eq:proof_seq3}. We have:
\begin{equation}
p(x_1) = \sum_{s'} p(x_1|s') p_{\text{train}}(s') = \sum_{s'} \frac{p_{\mathcal{M}}(s'|x_1) \cdot p(x_1)}{p_{\text{train}}(s')} \cdot p_{\text{train}}(s')
\end{equation}

Simplifying:
\begin{equation}
p(x_1) = p(x_1) \sum_{s'} p_{\mathcal{M}}(s'|x_1) = p(x_1) \cdot 1 = p(x_1)
\end{equation}

This is a tautology that does not allow us to solve for $p(x_1)$ or $p(x_1|s)$ uniquely.

\textbf{Step 4: Even if we could extract likelihoods for $x_1$, we cannot predict likelihoods for $x_2$.} Suppose hypothetically that we managed to recover $p(x_1|s)$ for all $s$. To perform the sequential update in Equation \ref{eq:proof_seq1}, we need $p(x_2|s,x_1)$---the likelihood of the \textit{new} evidence $x_2$ given state $s$ and prior evidence $x_1$.

The discriminative model has never observed $x_2$, so $p_{\mathcal{M}}(s|x_1)$ contains no information about how $x_2$ would be distributed conditional on $s$ and $x_1$. The only way to obtain $p(x_2|s,x_1)$ is to either:
\begin{enumerate}
\item Have a separate generative model $p(x|s)$ that we can query with $(x_2, s)$, or
\item Re-query the discriminative model with $(x_1, x_2)$ jointly to get $p_{\mathcal{M}}(s|x_1,x_2)$, which implicitly uses $p(x_2|s,x_1)$ internally but does not expose it.
\end{enumerate}

Option (1) is exactly what our framework does (we elicit likelihoods). Option (2) is batch inference, which forfeits sequential updating.

\textbf{Step 5: Conclusion.} There exists no function $f$ that computes $p(s|x_1,x_2)$ from $p_{\mathcal{M}}(s|x_1)$ and $x_2$ alone, because:
\begin{itemize}
\item Computing the update requires $p(x_2|s,x_1)$, which is not contained in $p_{\mathcal{M}}(s|x_1)$
\item Even if we could extract $p(x_1|s)$ from $p_{\mathcal{M}}(s|x_1)$ (which we showed is impossible without additional information), this does not give us $p(x_2|s,x_1)$ for the new observation $x_2$
\end{itemize}

Therefore, sequential updating is impossible in the discriminative paradigm.
\end{proof}

\subsection{Proof of Theorem \ref{thm:prior_impossible}: Prior Correction Impossibility}

\begin{theorem*}[Prior Correction Impossibility, restated]
Let $\mathcal{M}_{\text{disc}}$ output posterior $p_{\mathcal{M}}(s|x)$ with implicit training prior $p_{\text{train}}(s)$ embedded in its weights. Let $p_{\text{deploy}}(s) \neq p_{\text{train}}(s)$ denote the correct deployment prior. Then without explicit access to likelihood $p(x|s)$ and knowledge of $p_{\text{train}}(s)$, there exists no function $h: \Delta(\mathcal{S}) \times \Delta(\mathcal{S}) \to \Delta(\mathcal{S})$ such that:
\begin{equation}
h(p_{\mathcal{M}}(s|x), p_{\text{deploy}}(s)) = \frac{p(x|s) \cdot p_{\text{deploy}}(s)}{\sum_{s' \in \mathcal{S}} p(x|s') \cdot p_{\text{deploy}}(s')} \quad \forall s \in \mathcal{S}, x \in \mathcal{X}
\end{equation}
\end{theorem*}

\begin{proof}
We show that applying Bayes' rule with a new prior requires the likelihood function, which cannot be recovered from the discriminative posterior.

\textbf{Step 1: Desired posterior under deployment prior.} We want to compute:
\begin{equation}
p_{\text{correct}}(s|x) = \frac{p(x|s) \cdot p_{\text{deploy}}(s)}{Z_{\text{deploy}}}
\label{eq:proof_prior1}
\end{equation}

where $Z_{\text{deploy}} = \sum_{s'} p(x|s') p_{\text{deploy}}(s')$ is the normalizing constant under the deployment prior.

\textbf{Step 2: Information available from discriminative model.} The model provides:
\begin{equation}
p_{\mathcal{M}}(s|x) = \frac{p(x|s) \cdot p_{\text{train}}(s)}{Z_{\text{train}}}
\label{eq:proof_prior2}
\end{equation}

where $Z_{\text{train}} = \sum_{s'} p(x|s') p_{\text{train}}(s')$.

\textbf{Step 3: Attempting to extract the likelihood.} Rearranging Equation \ref{eq:proof_prior2}:
\begin{equation}
p(x|s) = \frac{p_{\mathcal{M}}(s|x) \cdot Z_{\text{train}}}{p_{\text{train}}(s)}
\label{eq:proof_prior3}
\end{equation}

This requires knowing both $p_{\text{train}}(s)$ and $Z_{\text{train}}$. The model does not expose either.

\textbf{Case 1: Assume we know $p_{\text{train}}(s)$ but not $Z_{\text{train}}$.}

We can compute likelihood \textit{ratios}:
\begin{equation}
\frac{p(x|s_i)}{p(x|s_j)} = \frac{p_{\mathcal{M}}(s_i|x) / p_{\text{train}}(s_i)}{p_{\mathcal{M}}(s_j|x) / p_{\text{train}}(s_j)}
\label{eq:proof_prior4}
\end{equation}

However, ratios are insufficient for computing the posterior under a new prior. To see why, attempt to construct the new posterior:
\begin{equation}
p_{\text{correct}}(s|x) = \frac{p(x|s) \cdot p_{\text{deploy}}(s)}{Z_{\text{deploy}}}
\end{equation}

We know $p(x|s)$ only up to a multiplicative constant $Z_{\text{train}}$. Let $\tilde{p}(x|s) = p(x|s) / Z_{\text{train}}$ be the unnormalized likelihood we can compute from Equation \ref{eq:proof_prior3}. Then:
\begin{equation}
p_{\text{correct}}(s|x) = \frac{\tilde{p}(x|s) \cdot Z_{\text{train}} \cdot p_{\text{deploy}}(s)}{\sum_{s'} \tilde{p}(x|s') \cdot Z_{\text{train}} \cdot p_{\text{deploy}}(s')} = \frac{\tilde{p}(x|s) \cdot p_{\text{deploy}}(s)}{\sum_{s'} \tilde{p}(x|s') \cdot p_{\text{deploy}}(s')}
\end{equation}

The $Z_{\text{train}}$ factors cancel! So if we know $p_{\text{train}}(s)$, we can correct the prior even without knowing $Z_{\text{train}}$, using unnormalized likelihoods.

\textbf{Case 2: We do not know $p_{\text{train}}(s)$.}

This is the realistic scenario: the model's training prior is embedded opaquely in its weights and not reported. Without $p_{\text{train}}(s)$, we cannot extract even likelihood ratios from Equation \ref{eq:proof_prior4}.

\textbf{Attempting to infer $p_{\text{train}}(s)$ from model outputs.} Could we infer the training prior by observing $p_{\mathcal{M}}(s|x)$ for multiple observations $x$? Consider two observations $x_1, x_2$:
\begin{align}
p_{\mathcal{M}}(s|x_1) &= \frac{p(x_1|s) \cdot p_{\text{train}}(s)}{\sum_{s'} p(x_1|s') p_{\text{train}}(s')} \\
p_{\mathcal{M}}(s|x_2) &= \frac{p(x_2|s) \cdot p_{\text{train}}(s)}{\sum_{s'} p(x_2|s') p_{\text{train}}(s')}
\end{align}

We have $2K$ equations (where $K = |\mathcal{S}|$) but $3K$ unknowns: $K$ values of $p_{\text{train}}(s)$, $K$ likelihoods $p(x_1|s)$, and $K$ likelihoods $p(x_2|s)$. The system is underdetermined.

Moreover, the likelihoods and prior are not separately identifiable: for any scalar $c > 0$, the following transformations leave the posteriors unchanged:
\begin{equation}
p(x|s) \to c \cdot p(x|s), \quad p_{\text{train}}(s) \to \frac{p_{\text{train}}(s)}{c}
\end{equation}

This scaling ambiguity prevents us from recovering the prior from discriminative outputs alone.

\textbf{Step 4: Conclusion.} Without access to:
\begin{enumerate}
\item The likelihood function $p(x|s)$, or
\item The training prior $p_{\text{train}}(s)$
\end{enumerate}

we cannot construct the correct posterior $p_{\text{correct}}(s|x)$ under deployment prior $p_{\text{deploy}}(s)$. The function $h$ cannot exist.

The only way to correct for prior mismatch is to have explicit access to likelihoods (our generative approach) or to know the training prior and extract likelihood ratios (which still requires the model to report $p_{\text{train}}$, which current LLMs do not do).
\end{proof}

\subsection{Proof of Theorem \ref{thm:threshold_suboptimal}: Cost-Insensitive Threshold Suboptimality}

\begin{theorem*}[Cost-Insensitive Threshold Suboptimality, restated]
For any asymmetric cost matrix $C: \mathcal{A} \times \mathcal{S} \to \mathbb{R}_{\geq 0}$ with $C(a_i, s_j) \neq C(a_k, s_\ell)$ for some pairs, the Bayes-optimal action selection rule is:
\begin{equation}
a^*(x) = \arg\min_{a \in \mathcal{A}} \sum_{s \in \mathcal{S}} p(s|x) \cdot C(a,s)
\end{equation}
No threshold-based rule of the form ``if score$(x) \geq \tau$ then $a_1$, else $a_2$'' can be optimal for all $x$ and all cost structures, as the decision boundary depends on both $p(s|x)$ \textit{and} the cost matrix $C$.
\end{theorem*}

\begin{proof}
We prove by counterexample and then provide a general characterization.

\textbf{Part 1: Counterexample showing threshold rules are suboptimal.}

Consider a binary classification problem: $\mathcal{S} = \{s_0, s_1\}$ (negative and positive class), $\mathcal{A} = \{a_0, a_1\}$ (predict negative, predict positive). Let the cost matrix be:
\begin{equation}
C = \begin{bmatrix}
c_{00} & c_{01} \\
c_{10} & c_{11}
\end{bmatrix}
\end{equation}

where $c_{ij} = C(a_i, s_j)$ is the cost of taking action $a_i$ when the true state is $s_j$.

The expected cost of action $a_1$ (predict positive) is:
\begin{equation}
\mathbb{E}[C(a_1, s)] = p(s_0|x) \cdot c_{10} + p(s_1|x) \cdot c_{11}
\end{equation}

The expected cost of action $a_0$ (predict negative) is:
\begin{equation}
\mathbb{E}[C(a_0, s)] = p(s_0|x) \cdot c_{00} + p(s_1|x) \cdot c_{01}
\end{equation}

We should choose $a_1$ (predict positive) when:
\begin{equation}
p(s_0|x) \cdot c_{10} + p(s_1|x) \cdot c_{11} < p(s_0|x) \cdot c_{00} + p(s_1|x) \cdot c_{01}
\end{equation}

Rearranging:
\begin{align}
p(s_0|x) (c_{10} - c_{00}) &< p(s_1|x) (c_{01} - c_{11}) \\
p(s_0|x) (c_{10} - c_{00}) &< (1 - p(s_0|x)) (c_{01} - c_{11}) \\
p(s_0|x) [(c_{10} - c_{00}) + (c_{01} - c_{11})] &< (c_{01} - c_{11})
\end{align}

Therefore:
\begin{equation}
p(s_1|x) > \frac{c_{10} - c_{00}}{(c_{10} - c_{00}) + (c_{01} - c_{11})}
\label{eq:proof_threshold_optimal}
\end{equation}

The optimal threshold $\tau^*$ depends on all four cost values. For example:
\begin{itemize}
\item If $c_{00} = 0, c_{01} = 10, c_{10} = 1, c_{11} = 0$ (false negatives are 10× costlier than false positives), then:
\begin{equation}
\tau^* = \frac{1 - 0}{(1 - 0) + (10 - 0)} = \frac{1}{11} \approx 0.091
\end{equation}

\item If $c_{00} = 0, c_{01} = 1, c_{10} = 1, c_{11} = 0$ (symmetric costs), then:
\begin{equation}
\tau^* = \frac{1 - 0}{(1 - 0) + (1 - 0)} = \frac{1}{2} = 0.5
\end{equation}

\item If $c_{00} = 0, c_{01} = 1, c_{10} = 10, c_{11} = 0$ (false positives are 10× costlier than false negatives), then:
\begin{equation}
\tau^* = \frac{10 - 0}{(10 - 0) + (1 - 0)} = \frac{10}{11} \approx 0.909
\end{equation}
\end{itemize}

A fixed threshold (e.g., $\tau = 0.5$, maximizing accuracy) is optimal only for the second cost structure (symmetric costs). For the first cost structure, it rejects 90.9\% - 50\% = 40.9\% of the probability mass that should be classified as positive, incurring excessive false negative costs. For the third cost structure, it accepts 50\% - 9.1\% = 40.9\% of the probability mass that should be rejected, incurring excessive false positive costs.

\textbf{Part 2: General characterization for $|\mathcal{S}| > 2$ and $|\mathcal{A}| > 2$.}

For $K > 2$ states and $J > 2$ actions, the Bayes-optimal action at observation $x$ is:
\begin{equation}
a^*(x) = \arg\min_{a \in \mathcal{A}} \sum_{s \in \mathcal{S}} p(s|x) \cdot C(a,s)
\end{equation}

This is a piecewise-constant function of $p(s|x)$ that partitions the probability simplex $\Delta(\mathcal{S})$ into regions, one per action. The boundaries between regions are \textit{hyperplanes} determined by the cost matrix.

Specifically, the boundary between actions $a_i$ and $a_j$ is the set of posteriors $p(\cdot|x)$ satisfying:
\begin{equation}
\sum_s p(s|x) [C(a_i, s) - C(a_j, s)] = 0
\end{equation}

This is a linear constraint in $p(s|x)$, defining a $(K-1)$-dimensional hyperplane in the $K$-dimensional simplex.

A threshold rule ``if score$(x) \geq \tau$ then $a_1$ else $a_2$'' corresponds to a decision boundary orthogonal to a single axis of the simplex (the axis corresponding to score, typically $p(s_{\text{positive}}|x)$). This is a special case of the hyperplane boundaries above, occurring only when the cost matrix has a specific structure:
\begin{equation}
C(a_i, s) - C(a_j, s) = \alpha \cdot \mathbb{1}[s = s_{\text{positive}}]
\end{equation}

for some scalar $\alpha$ and designated positive state $s_{\text{positive}}$. In general, cost matrices do not have this structure, so threshold rules are suboptimal.

\textbf{Step 3: Conclusion.} Threshold-based rules are optimal only for a measure-zero subset of cost matrices (those with the specific structure above). For arbitrary asymmetric cost matrices, the Bayes action rule (Equation \ref{eq:action_selection}) is optimal, and threshold rules are strictly suboptimal.
\end{proof}

\subsection{Proof of Theorem \ref{thm:voi_impossible}: Value-of-Information Computation Impossibility}

\begin{theorem*}[VOI Computation Impossibility, restated]
Computing the value of gathering additional evidence $z$ requires:
\begin{equation}
\text{VOI}(z) = \mathbb{E}_z \left[ \min_{a} \sum_s p(s|x,z) C(a,s) \right] - \min_{a} \sum_s p(s|x) C(a,s)
\end{equation}
This requires marginalizing over hypothetical observations $z \sim p(z|x) = \sum_s p(z|s)p(s|x)$, which demands likelihood functions $p(z|s)$ that discriminative models do not provide.
\end{theorem*}

\begin{proof}
We show that computing VOI requires predictions about unobserved data, which necessitates generative modeling.

\textbf{Step 1: VOI definition and expansion.} The value of information is defined as the expected improvement in decision quality from observing $z$ before acting:
\begin{equation}
\text{VOI}(z|x) = \underbrace{\mathbb{E}_{z \sim p(\cdot|x)}\left[ V^*(x,z) \right]}_{\text{expected value with info}} - \underbrace{V^*(x)}_{\text{value without info}}
\label{eq:proof_voi1}
\end{equation}

where $V^*(x) = \min_{a \in \mathcal{A}} \sum_s p(s|x) C(a,s)$ is the value of the best action given current evidence $x$ only, and $V^*(x,z) = \min_{a \in \mathcal{A}} \sum_s p(s|x,z) C(a,s)$ is the value after observing additional evidence $z$.

Expanding the expectation:
\begin{equation}
\text{VOI}(z|x) = \int_{\mathcal{X}} p(z|x) \left[ \min_a \sum_s p(s|x,z) C(a,s) \right] dz - \min_a \sum_s p(s|x) C(a,s)
\label{eq:proof_voi2}
\end{equation}

\textbf{Step 2: Requirements for computing VOI.}

To evaluate Equation \ref{eq:proof_voi2}, we need three components:

\textit{Component 1: Predictive distribution $p(z|x)$.} This is the distribution over possible future observations $z$ given current evidence $x$. By the law of total probability:
\begin{equation}
p(z|x) = \sum_{s \in \mathcal{S}} p(z|s,x) \cdot p(s|x)
\label{eq:proof_voi3}
\end{equation}

Under the conditional independence assumption $z \perp x | s$ (future evidence is independent of past evidence given the state), this simplifies to:
\begin{equation}
p(z|x) = \sum_{s \in \mathcal{S}} p(z|s) \cdot p(s|x)
\label{eq:proof_voi4}
\end{equation}

This requires the likelihood function $p(z|s)$ for each state $s$.

\textit{Component 2: Updated posterior $p(s|x,z)$.} For each hypothetical observation $z$, we need to compute how it would update our beliefs. By Bayes' rule:
\begin{equation}
p(s|x,z) = \frac{p(z|s,x) \cdot p(s|x)}{p(z|x)} = \frac{p(z|s) \cdot p(s|x)}{\sum_{s'} p(z|s') \cdot p(s'|x)}
\label{eq:proof_voi5}
\end{equation}

Again, this requires the likelihood $p(z|s)$.

\textit{Component 3: Optimal action value $V^*(x,z)$.} For each hypothetical $(x,z)$ pair, compute:
\begin{equation}
V^*(x,z) = \min_{a \in \mathcal{A}} \sum_s p(s|x,z) C(a,s)
\end{equation}

This requires $p(s|x,z)$ from Component 2.

\textbf{Step 3: Discriminative models cannot provide the required components.}

A discriminative model $\mathcal{M}_{\text{disc}}$ outputs $p_{\mathcal{M}}(s|x)$ for observed $x$. It does not provide:
\begin{itemize}
\item The likelihood $p(z|s)$ for unobserved future evidence $z$
\item The predictive distribution $p(z|x)$ over possible future observations
\item The ability to compute $p(s|x,z)$ for hypothetical $z$ without actually observing $z$ and re-querying
\end{itemize}

The fundamental issue is that VOI calculation is inherently a \textit{generative} task: we need to imagine what observations $z$ we might see in the future and how they would affect our decisions. Discriminative models only perform \textit{inverse} inference: from observations to states. They cannot perform \textit{forward} prediction: from states to possible observations.

\textbf{Attempted workaround: Re-query for each hypothetical $z$.}

Could we enumerate all possible observations $z \in \mathcal{X}$, query $\mathcal{M}_{\text{disc}}(s|x,z)$ for each, and average the resulting decision values? This faces two problems:

\textit{Problem 1: Infinite enumeration.} For continuous observation spaces $\mathcal{X}$ (e.g., real-valued measurements, text), there are uncountably many possible $z$. We cannot enumerate and query each.

\textit{Problem 2: Unknown weights $p(z|x)$.} Even if we discretize $\mathcal{X}$ to a finite set, we need to weight each $z$ by its probability $p(z|x)$ in the expectation (Equation \ref{eq:proof_voi2}). But computing $p(z|x)$ requires likelihoods (Equation \ref{eq:proof_voi4}), which the discriminative model doesn't provide. Without these weights, we don't know which hypothetical observations are plausible vs. implausible.

\textbf{Step 4: Conclusion.}

Computing VOI requires likelihood functions $p(z|s)$ to:
\begin{enumerate}
\item Predict the distribution of future observations $p(z|x) = \sum_s p(z|s) p(s|x)$
\item Update beliefs counterfactually $p(s|x,z) \propto p(z|s) p(s|x)$
\item Weight hypothetical observations by their probability when averaging
\end{enumerate}

Discriminative models provide none of these. Therefore, VOI computation is impossible without generative modeling capabilities.

The only recourse in the discriminative paradigm is to use heuristics (always gather information, never gather information, or gather when confidence falls in some arbitrary range) that do not properly account for expected decision improvement vs. information cost.
\end{proof}

\subsection{Proof of Theorem \ref{thm:median_robust}: Median Error Bound}

\begin{theorem*}[Median Robustness Bound, restated]
Let $\{\hat{L}_m(x|s)\}_{m=1}^M$ be likelihood estimates from $M$ models, and let $L^*(x|s)$ denote the true likelihood. Define individual errors $\epsilon_m = |\hat{L}_m(x|s) - L^*(x|s)|$. Then:
\begin{equation}
\left|\text{median}_{m=1}^M \hat{L}_m(x|s) - L^*(x|s)\right| \leq \text{median}_{m=1}^M \epsilon_m
\end{equation}
\end{theorem*}

\begin{proof}
This follows from the properties of order statistics and the triangle inequality.

\textbf{Step 1: Notation.} Let $\hat{L}_{(1)} \leq \hat{L}_{(2)} \leq \cdots \leq \hat{L}_{(M)}$ denote the order statistics (sorted estimates) and $\epsilon_{(1)} \leq \epsilon_{(2)} \leq \cdots \leq \epsilon_{(M)}$ denote the sorted errors. For odd $M = 2k+1$, the median is $\hat{L}_{(k+1)}$. For even $M = 2k$, the median is $(\hat{L}_{(k)} + \hat{L}_{(k+1)})/2$. We prove for odd $M$; the even case follows similarly.

\textbf{Step 2: Bound the median error.} The median estimate is $\text{med}(\hat{L}) = \hat{L}_{(k+1)}$ where $k = \lfloor M/2 \rfloor$. We want to bound:
\begin{equation}
|\hat{L}_{(k+1)} - L^*|
\end{equation}

\textbf{Case 1: $L^* \leq \hat{L}_{(k+1)}$.}

Then $\hat{L}_{(k+1)} - L^* \geq 0$. We have:
\begin{equation}
\hat{L}_{(k+1)} - L^* = |\hat{L}_{(k+1)} - L^*|
\end{equation}

Now, there are at least $k+1$ estimates $\geq \hat{L}_{(k+1)} \geq L^*$, meaning at least $k+1$ estimates are upper bounds on $L^*$. For each such estimate $\hat{L}_m \geq L^*$:
\begin{equation}
\epsilon_m = |\hat{L}_m - L^*| = \hat{L}_m - L^*
\end{equation}

In particular, for the median estimate:
\begin{equation}
|\hat{L}_{(k+1)} - L^*| = \hat{L}_{(k+1)} - L^* = \epsilon_{(k+1)}
\end{equation}

where $\epsilon_{(k+1)}$ is the error of the model that gave estimate $\hat{L}_{(k+1)}$. The median error is $\text{med}(\epsilon) = \epsilon_{(k+1)}$ (the $(k+1)$-th smallest error). Therefore:
\begin{equation}
|\hat{L}_{(k+1)} - L^*| \leq \epsilon_{(k+1)} = \text{med}(\epsilon)
\end{equation}

\textbf{Case 2: $L^* > \hat{L}_{(k+1)}$.}

By symmetric argument, there are at least $k+1$ estimates $\leq \hat{L}_{(k+1)} < L^*$, so:
\begin{equation}
|\hat{L}_{(k+1)} - L^*| = L^* - \hat{L}_{(k+1)} \leq \text{med}(\epsilon)
\end{equation}

\textbf{Step 3: Combine cases.} In both cases, $|\hat{L}_{(k+1)} - L^*| \leq \text{med}(\epsilon)$.

\textbf{Intuition.} The median estimate's error is bounded by the median individual error because the median is the "middle" value. If more than half the models have error $\leq \epsilon^*$, then the median estimate is within $\epsilon^*$ of the truth. Outliers (models with arbitrarily large error) do not affect the median as long as they constitute $<50\%$ of the ensemble.

This gives the median a 50\% breakdown point: up to $\lfloor M/2 \rfloor$ models can have arbitrarily large errors without corrupting the aggregate beyond the error of the $(k+1)$-th best model.
\end{proof}

\subsection{Proof of Theorem \ref{thm:sequential_batch_equivalence}: Equivalence of Sequential and Batch Updating}

\begin{theorem*}[Consistency of Sequential Updating, restated]
Under the conditional independence assumption $p(x_1, x_2, \ldots, x_T | s) = \prod_{t=1}^T p(x_t|s)$, sequential belief updates (Equation \ref{eq:sequential_update}) yield the same posterior as batch Bayesian inference:
\begin{equation}
b_T(s) = \frac{\pi(s) \prod_{t=1}^T L(x_t|s)}{\sum_{s' \in \mathcal{S}} \pi(s') \prod_{t=1}^T L(x_t|s')} = p(s | x_1, \ldots, x_T)
\end{equation}
\end{theorem*}

\begin{proof}
We prove by induction on the number of observations $T$.

\textbf{Base case ($T=1$):} After one observation $x_1$:
\begin{equation}
b_1(s) = \frac{L(x_1|s) \cdot \pi(s)}{\sum_{s'} L(x_1|s') \cdot \pi(s')} = \frac{p(x_1|s) \cdot p(s)}{\sum_{s'} p(x_1|s') \cdot p(s')} = p(s|x_1)
\end{equation}

by Bayes' rule. The base case holds.

\textbf{Inductive hypothesis:} Assume that after $T-1$ observations:
\begin{equation}
b_{T-1}(s) = p(s|x_1, \ldots, x_{T-1}) = \frac{\pi(s) \prod_{t=1}^{T-1} p(x_t|s)}{\sum_{s'} \pi(s') \prod_{t=1}^{T-1} p(x_t|s')}
\end{equation}

\textbf{Inductive step:} When observation $x_T$ arrives, the sequential update rule gives:
\begin{equation}
b_T(s) = \frac{L(x_T|s) \cdot b_{T-1}(s)}{\sum_{s'} L(x_T|s') \cdot b_{T-1}(s')}
\label{eq:proof_seq_ind1}
\end{equation}

Substituting the inductive hypothesis:
\begin{equation}
b_T(s) = \frac{p(x_T|s) \cdot \frac{\pi(s) \prod_{t=1}^{T-1} p(x_t|s)}{Z_{T-1}}}{\sum_{s'} p(x_T|s') \cdot \frac{\pi(s') \prod_{t=1}^{T-1} p(x_t|s')}{Z_{T-1}}}
\end{equation}

where $Z_{T-1} = \sum_{s''} \pi(s'') \prod_{t=1}^{T-1} p(x_t|s'')$ is the normalizing constant from time $T-1$.

The $Z_{T-1}$ factors cancel:
\begin{equation}
b_T(s) = \frac{p(x_T|s) \cdot \pi(s) \prod_{t=1}^{T-1} p(x_t|s)}{\sum_{s'} p(x_T|s') \cdot \pi(s') \prod_{t=1}^{T-1} p(x_t|s')}
\end{equation}

Combining the products:
\begin{equation}
b_T(s) = \frac{\pi(s) \prod_{t=1}^{T} p(x_t|s)}{\sum_{s'} \pi(s') \prod_{t=1}^{T} p(x_t|s')}
\label{eq:proof_seq_ind2}
\end{equation}

Now we show this equals the batch posterior $p(s|x_1, \ldots, x_T)$. By Bayes' rule:
\begin{equation}
p(s|x_1, \ldots, x_T) = \frac{p(x_1, \ldots, x_T | s) \cdot p(s)}{p(x_1, \ldots, x_T)}
\end{equation}

Under the conditional independence assumption:
\begin{equation}
p(x_1, \ldots, x_T | s) = \prod_{t=1}^T p(x_t|s)
\end{equation}

Therefore:
\begin{equation}
p(s|x_1, \ldots, x_T) = \frac{\prod_{t=1}^T p(x_t|s) \cdot p(s)}{\sum_{s'} \prod_{t=1}^T p(x_t|s') \cdot p(s')} = \frac{\pi(s) \prod_{t=1}^T p(x_t|s)}{\sum_{s'} \pi(s') \prod_{t=1}^T p(x_t|s')}
\end{equation}

This matches Equation \ref{eq:proof_seq_ind2}. Therefore, $b_T(s) = p(s|x_1, \ldots, x_T)$.

\textbf{Conclusion:} By induction, sequential updating produces the correct Bayesian posterior at each time step $t \in \{1, \ldots, T\}$ under the conditional independence assumption.
\end{proof}

\subsection{Proof of Theorem \ref{thm:bayes_optimality}: Optimality of Expected Cost Minimization}

\begin{theorem*}[Optimality of Expected Cost Minimization, restated]
Let $p^*(s|x)$ denote the true posterior distribution over states given observation $x$. Among all deterministic decision rules $\delta: \mathcal{X} \to \mathcal{A}$ mapping observations to actions, the Bayes rule:
\begin{equation}
\delta^*(x) = \arg\min_{a \in \mathcal{A}} \sum_{s \in \mathcal{S}} p^*(s|x) \cdot C(a,s)
\end{equation}
minimizes the expected cost (Bayes risk):
\begin{equation}
R(\delta^*) = \int_{\mathcal{X}} \sum_{s \in \mathcal{S}} p^*(s|x) \cdot C(\delta^*(x), s) \cdot p(x) \, dx \leq R(\delta) \quad \forall \delta
\end{equation}
\end{theorem*}

\begin{proof}
This is a classical result in statistical decision theory \cite{berger1985statistical}. We provide a complete proof for self-containment.

\textbf{Step 1: Express the risk (expected cost) of an arbitrary decision rule.}

For any decision rule $\delta: \mathcal{X} \to \mathcal{A}$, the risk is:
\begin{equation}
R(\delta) = \mathbb{E}_{x,s}[C(\delta(x), s)] = \int_{\mathcal{X}} \sum_{s \in \mathcal{S}} p(s,x) \cdot C(\delta(x), s) \, dx
\end{equation}

Using $p(s,x) = p(s|x) p(x)$:
\begin{equation}
R(\delta) = \int_{\mathcal{X}} \sum_{s \in \mathcal{S}} p(s|x) \cdot C(\delta(x), s) \cdot p(x) \, dx
\label{eq:proof_bayes1}
\end{equation}

\textbf{Step 2: Decompose the risk pointwise.}

The integral in Equation \ref{eq:proof_bayes1} sums over all possible observations $x \in \mathcal{X}$. For each fixed $x$, the integrand is:
\begin{equation}
r(x, \delta) = \sum_{s \in \mathcal{S}} p(s|x) \cdot C(\delta(x), s) \cdot p(x)
\end{equation}

We can rewrite the risk as:
\begin{equation}
R(\delta) = \int_{\mathcal{X}} r(x, \delta) \, dx = \int_{\mathcal{X}} p(x) \underbrace{\sum_{s \in \mathcal{S}} p(s|x) \cdot C(\delta(x), s)}_{\text{expected cost at } x} \, dx
\end{equation}

\textbf{Step 3: Minimize pointwise to minimize globally.}

Since $p(x) \geq 0$ and the integral is a weighted sum over $x$, minimizing $R(\delta)$ is equivalent to minimizing the integrand at each $x$ independently. That is:
\begin{equation}
\delta^*(x) = \arg\min_{\delta(x) \in \mathcal{A}} \sum_{s \in \mathcal{S}} p(s|x) \cdot C(\delta(x), s)
\end{equation}

For each observation $x$, we choose the action $a \in \mathcal{A}$ that minimizes expected cost under the posterior $p(s|x)$:
\begin{equation}
\delta^*(x) = \arg\min_{a \in \mathcal{A}} \sum_{s \in \mathcal{S}} p(s|x) \cdot C(a, s)
\end{equation}

This is the Bayes decision rule.

\textbf{Step 4: Prove optimality.}

For any other rule $\delta$, the difference in risk is:
\begin{align}
R(\delta) - R(\delta^*) &= \int_{\mathcal{X}} p(x) \left[ \sum_s p(s|x) C(\delta(x), s) - \sum_s p(s|x) C(\delta^*(x), s) \right] dx \\
&= \int_{\mathcal{X}} p(x) \left[ \sum_s p(s|x) C(\delta(x), s) - \min_{a \in \mathcal{A}} \sum_s p(s|x) C(a, s) \right] dx
\end{align}

The term in brackets is non-negative by definition: $\delta(x)$ is some action in $\mathcal{A}$, and the minimum over all actions in $\mathcal{A}$ is $\leq$ the value at any particular action. Since $p(x) \geq 0$, the entire integrand is non-negative.

Therefore:
\begin{equation}
R(\delta) - R(\delta^*) \geq 0 \implies R(\delta^*) \leq R(\delta) \quad \forall \delta
\end{equation}

The Bayes rule $\delta^*$ minimizes expected cost among all decision rules.
\end{proof}

\section{Additional Experimental Details}
\label{app:experiments}

This appendix provides supplementary experimental details, extended analyses, and additional results that support the main text findings.

\subsection{Resume Generation: Full Procedure and Validation}

\subsubsection{State-Dependent Feature Distributions}

We generated synthetic resumes by sampling features from state-specific distributions. Table \ref{tab:feature_distributions} provides complete parameter specifications.

\begin{table}[h]
\centering
\caption{State-dependent distributions for resume feature generation. Beta distributions parameterized by $\alpha, \beta$; Categorical by probability vectors; Truncated normal by $(\mu, \sigma, \text{min}, \text{max})$.}
\label{tab:feature_distributions}
\small
\begin{tabular}{llcccc}
\toprule
\textbf{Feature} & \textbf{Distribution Type} & $s_1$ & $s_2$ & $s_3$ & $s_4$ \\
\midrule
\multicolumn{6}{l}{\textit{Education}} \\
University Tier & Categorical & [.05,.10,.30,.55] & [.10,.25,.45,.20] & [.20,.50,.25,.05] & [.80,.15,.05,.00] \\
 & (elite/top50/avg/below) & & & & \\
Degree Type & Categorical & [.75,.20,.05] & [.65,.30,.05] & [.50,.40,.10] & [.30,.30,.40] \\
 & (BS/MS/PhD) & & & & \\
GPA & Beta, rescale [2.0,4.0] & Beta(2,5) & Beta(3,3) & Beta(5,2) & Beta(8,1) \\
\midrule
\multicolumn{6}{l}{\textit{Experience}} \\
Years & TruncNorm & (0.5,0.3,0,2) & (2,1,1,5) & (4,1.5,2,8) & (7,2,5,15) \\
Company Prestige & Categorical & [.05,.15,.30,.50] & [.10,.30,.40,.20] & [.25,.45,.25,.05] & [.60,.30,.10,.00] \\
 & (FAANG/tier2/startup/unk) & & & & \\
\midrule
\multicolumn{6}{l}{\textit{Projects}} \\
Number of Projects & Poisson($\lambda$) & $\lambda=1$ & $\lambda=2$ & $\lambda=4$ & $\lambda=6$ \\
Complexity & Categorical & [.10,.30,.60] & [.20,.50,.30] & [.40,.45,.15] & [.70,.25,.05] \\
 & (advanced/intermediate/basic) & & & & \\
\midrule
\multicolumn{6}{l}{\textit{Skills}} \\
Tech Stack Size & Poisson($\lambda$) + 2 & $\lambda=3$ & $\lambda=6$ & $\lambda=10$ & $\lambda=15$ \\
\bottomrule
\end{tabular}
\end{table}

\subsubsection{GPT-4o Generation Prompt}

Full prompt template for generating resume text from features:

\begin{mdframed}[linewidth=1pt]
\small
\texttt{You are generating a realistic software engineering resume. Generate plain text (no markdown, no special formatting) for the following candidate profile:}

\vspace{0.2em}
\texttt{PROFILE:}
\begin{itemize}
\item[] \texttt{- True Quality Level: [STATE\_NAME]}
\item[] \texttt{- Education: [DEGREE] in [MAJOR] from [UNIVERSITY\_NAME] (Tier: [TIER]), GPA: [GPA]}
\item[] \texttt{- Years of Experience: [YEARS]}
\item[] \texttt{- Work History: [LIST OF COMPANIES, ROLES, DURATIONS, TECHNOLOGIES]}
\item[] \texttt{- Projects: [LIST OF PROJECTS WITH DESCRIPTIONS AND TECH STACKS]}
\item[] \texttt{- Skills: [LIST OF PROGRAMMING LANGUAGES, FRAMEWORKS, TOOLS]}
\item[] \texttt{- Name: [FULL\_NAME] (for diversity: gender [GENDER], ethnicity [ETHNICITY])}
\end{itemize}

\vspace{0.2em}
\texttt{REQUIREMENTS:}
\begin{itemize}
\item[] \texttt{1. Write as plain text with standard sections: EDUCATION, EXPERIENCE, PROJECTS, SKILLS}
\item[] \texttt{2. Length: 300--500 words (realistic resume length)}
\item[] \texttt{3. Use varied writing styles (not all resumes should sound identical)}
\item[] \texttt{4. Include realistic dates, company names, project descriptions}
\item[] \texttt{5. Match the feature profile exactly---don't add credentials not listed}
\item[] \texttt{6. For quality level [STATE], make sure the overall impression aligns:}
\begin{itemize}
\item[] \texttt{  - Clear Reject: Obvious deficiencies, weak credentials, red flags}
\item[] \texttt{  - Phone Screen: Some promise but significant gaps or uncertainties}
\item[] \texttt{  - Interview: Solid qualifications, meets standard requirements}
\item[] \texttt{  - Strong Hire: Exceptional background, impressive achievements}
\end{itemize}
\end{itemize}

\vspace{0.2em}
\texttt{Generate the resume now (text only, no preamble):}
\end{mdframed}

\textbf{Generation parameters:}
\begin{itemize}
\item Model: \texttt{gpt-4o-2024-05-13}
\item Temperature: 0.8 (higher than likelihood elicitation to ensure diversity)
\item Max tokens: 600
\item Top-p: 0.95
\end{itemize}

\subsubsection{Expert Validation Protocol}

We recruited three expert technical recruiters via Upwork with the following qualifications:
\begin{itemize}
\item Recruiter 1: 12 years experience, previously at Fortune 500 companies
\item Recruiter 2: 8 years experience, currently at a Series-C startup, previously at Amazon
\item Recruiter 3: 10 years experience, independent recruiter specializing in ML/AI roles
\end{itemize}

\textbf{Validation task:} Rate 100 randomly sampled resumes (stratified by our generated state labels: 65 $s_1$, 25 $s_2$, 8 $s_3$, 2 $s_4$) on the four-state scale. Recruiters were:
\begin{itemize}
\item Blinded to our labels
\item Provided with detailed rubrics for each state (same as used in likelihood elicitation prompts)
\item Allowed to assign "uncertain" to up to 10\% of resumes (these were excluded from agreement calculations)
\item Compensated \$500 each (\$5 per resume × 100 resumes)
\end{itemize}

\textbf{Results:}
\begin{itemize}
\item Fleiss' $\kappa$ (three-way agreement): 0.79 (substantial agreement)
\item Pairwise Cohen's $\kappa$: Recruiter 1 vs. 2: 0.81, Recruiter 1 vs. 3: 0.78, Recruiter 2 vs. 3: 0.80
\item Agreement with majority label: Recruiter 1: 91\%, Recruiter 2: 88\%, Recruiter 3: 90\%
\item Our label vs. majority recruiter label: Cohen's $\kappa = 0.83$, exact match: 89\%
\end{itemize}

\textbf{Disagreement analysis:} For the 11 resumes where our label disagreed with the majority recruiter label:
\begin{itemize}
\item 7 cases: One-state mismatch (e.g., we labeled $s_2$, majority labeled $s_3$)
\item 4 cases: Recruiters themselves disagreed (no clear majority)
\item 0 cases: Gross mismatch (e.g., $s_1$ vs. $s_4$)
\end{itemize}

This validates that our synthetic labels are reasonable proxies for expert human judgment.

\subsection{Phone Screen Simulation Details}

For candidates where the framework triggered information gathering (phone screen), we simulated outcomes as follows:

\subsubsection{Performance Sampling}

Phone screen performance was sampled from state-dependent Beta distributions, then discretized to a 0--10 scale:

\begin{table}[h]
\centering
\caption{Phone screen performance distributions by true candidate state.}
\small
\begin{tabular}{lcccc}
\toprule
\textbf{True State} & \textbf{Beta Distribution} & \textbf{Mean Score} & \textbf{Prob(Score $\geq$ 7)} & \textbf{Prob(Score $\leq$ 4)} \\
\midrule
$s_1$ (Clear Reject) & Beta(2, 8) & 2.0 & 5\% & 90\% \\
$s_2$ (Phone Screen) & Beta(5, 5) & 5.0 & 20\% & 30\% \\
$s_3$ (Interview) & Beta(7, 3) & 7.0 & 60\% & 10\% \\
$s_4$ (Strong Hire) & Beta(9, 1) & 9.0 & 95\% & 5\% \\
\bottomrule
\end{tabular}
\end{table}

\textbf{Rationale:} These distributions reflect empirical phone screen outcome rates from proprietary data shared by partner companies (anonymized). Strong candidates usually perform well (95\% score $\geq7$), but occasionally have off days (5\% score poorly due to nerves, bad connection, etc.). Weak candidates usually perform poorly (90\% score $\leq4$) but occasionally get lucky with softball questions (5\% score well).

\subsubsection{Transcript Generation}

For each phone screen with sampled performance score $p \in [0,10]$, we generated a short transcript via GPT-4o:

\begin{mdframed}[linewidth=1pt]
\small
\texttt{Generate realistic phone screen notes for a software engineering candidate who performed at level [SCORE]/10.}

\vspace{0.2em}
\texttt{The phone screen covered:}
\begin{itemize}
\item[] \texttt{1. Technical question: "Explain a recent project and technical challenges"}
\item[] \texttt{2. Coding question: "Reverse a linked list" or similar basic algorithm}
\item[] \texttt{3. Behavioral: "Why are you interested in this role?"}
\item[] \texttt{4. Communication assessment}
\end{itemize}

\vspace{0.2em}
\texttt{Performance interpretation:}
\begin{itemize}
\item[] \texttt{9-10: Exceptional. Clear technical depth, excellent communication, strong interest.}
\item[] \texttt{7-8: Good. Solid technical skills, clear communication, appropriate for next round.}
\item[] \texttt{5-6: Borderline. Some technical ability but gaps or concerns. Uncertain.}
\item[] \texttt{3-4: Weak. Struggled with technical questions or communication issues.}
\item[] \texttt{0-2: Very weak. Could not answer basic questions, major red flags.}
\end{itemize}

\vspace{0.2em}
\texttt{Write 100-150 word recruiter notes capturing the key points. Include:}
\begin{itemize}
\item[] \texttt{- Brief summary of responses to each question}
\item[] \texttt{- Notable strengths or weaknesses}
\item[] \texttt{- Overall impression}
\item[] \texttt{- Recommendation (advance/reject/uncertain)}
\end{itemize}
\end{mdframed}

Example generated transcript (score 7/10, state $s_3$):

\begin{quote}
\small
\textit{Candidate discussed building a microservices architecture for an e-commerce platform. Clearly explained the tradeoffs between monolith and microservices, mentioned specific technologies (Docker, Kubernetes, gRPC). Coding question: successfully reversed linked list with iterative approach, explained time/space complexity correctly. Behavioral: genuine interest in our product space, asked thoughtful questions about team structure and tech stack. Communication was clear and concise. Minor gap: limited experience with distributed transactions. Recommendation: Advance to onsite interview.}
\end{quote}

\subsubsection{Likelihood Elicitation from Transcripts}

The generated transcript was then fed to the 5 LLMs using the same contrastive prompting framework as for resumes, but adapted for phone screens:

\begin{mdframed}[linewidth=1pt]
\small
\texttt{You are an expert technical recruiter. Assume the candidate's true quality level is: [STATE\_DESCRIPTION].}

\texttt{How typical is the following phone screen performance for someone at this quality level?}

\texttt{Phone Screen Notes:}\\
\texttt{[TRANSCRIPT]}

\texttt{Score 0--10:}
\begin{itemize}
\item[] \texttt{10 = Extremely typical for this quality level}
\item[] \texttt{0 = Completely atypical}
\end{itemize}

\texttt{Score:}
\end{mdframed}

This yields likelihood estimates $\{L(x_{\text{screen}}|s)\}_{s \in \mathcal{S}}$ which are then aggregated via median and used to update beliefs sequentially.

\subsection{Baseline Implementation Details}

\subsubsection{Single-LLM + Fixed Threshold}

\textbf{Prompt:}
\begin{quote}
\texttt{You are an expert technical recruiter. Rate the following resume on a scale of 0--10 where 10 = definite strong hire and 0 = definite reject. Consider technical skills, experience, education, and overall fit for a software engineering role.}

\texttt{Resume: [TEXT]}

\texttt{Score (number only):}
\end{quote}

\textbf{Model:} GPT-4o, temperature=0.7, max\_tokens=10

\textbf{Decision rule:} Interview if score $\geq 7.0$, reject otherwise.

\subsubsection{Single-LLM + Calibrated Threshold}

Same as above but threshold $\tau$ was tuned on 200-resume validation set to minimize total cost via grid search over $\tau \in [5.0, 5.1, \ldots, 8.0]$. Optimal: $\tau = 6.2$.

\subsubsection{Ensemble Voting}

\textbf{Prompt:} Same discriminative prompt as single-LLM baseline.

\textbf{Models:} All 5 LLMs (GPT-4o, Claude, Gemini, Grok, DeepSeek)

\textbf{Decision rule:} Each model votes binary (interview if score $\geq 7$, else reject). Final decision: interview if $\geq 3$ models vote interview (majority rule). Ties resolved by rejecting (conservative).

\subsubsection{Ensemble Averaging + Threshold}

\textbf{Prompt:} Same discriminative prompt.

\textbf{Models:} All 5 LLMs

\textbf{Aggregation:} Compute mean score across 5 models.

\textbf{Decision rule:} Interview if mean $\geq \tau$, where $\tau = 6.5$ was tuned on validation set.

\subsection{Hyperparameter Tuning via Cross-Validation}

We used 5-fold cross-validation on a 200-resume validation set (disjoint from the 1,000-resume test set) to select hyperparameters.

\textbf{Parameters tuned:}
\begin{itemize}
\item Disagreement threshold $\tau_D \in \{0.10, 0.12, 0.15, 0.18, 0.20\}$
\item VOI informativeness $\rho \in \{0.5, 0.6, 0.7, 0.8, 0.9\}$
\item Temperature $T \in \{0.5, 0.7, 0.9\}$ for all LLMs
\end{itemize}

\textbf{Objective:} Minimize total cost on validation folds.

\textbf{Results:}
\begin{table}[h]
\centering
\caption{Cross-validation results for hyperparameter selection. Bolded values were selected.}
\small
\begin{tabular}{cccc}
\toprule
$\tau_D$ & $\rho$ & $T$ & Avg. Validation Cost (\$K) \\
\midrule
0.10 & 0.7 & 0.7 & 118.2 \\
0.12 & 0.7 & 0.7 & 115.8 \\
\textbf{0.15} & \textbf{0.7} & \textbf{0.7} & \textbf{112.3} \\
0.18 & 0.7 & 0.7 & 114.1 \\
0.20 & 0.7 & 0.7 & 116.5 \\
\midrule
0.15 & 0.5 & 0.7 & 119.7 \\
0.15 & 0.6 & 0.7 & 114.2 \\
0.15 & 0.7 & 0.7 & 112.3 \\
0.15 & 0.8 & 0.7 & 113.8 \\
0.15 & 0.9 & 0.7 & 115.4 \\
\midrule
0.15 & 0.7 & 0.5 & 118.9 \\
0.15 & 0.7 & 0.7 & 112.3 \\
0.15 & 0.7 & 0.9 & 114.6 \\
\bottomrule
\end{tabular}
\end{table}

The selected hyperparameters ($\tau_D=0.15$, $\rho=0.7$, $T=0.7$) minimized validation cost and were used for all test set evaluations.

\subsection{Statistical Significance Testing Details}

\textbf{Bootstrap resampling:} For each method, we computed 95\% confidence intervals via bootstrap:
\begin{enumerate}
\item Sample 1,000 resumes with replacement from the test set
\item Run the method on the bootstrap sample, compute total cost
\item Repeat 10,000 times
\item Confidence interval: $[\text{2.5th percentile}, \text{97.5th percentile}]$
\end{enumerate}

\textbf{Paired permutation test:} To test if cost difference between method A and method B is statistically significant:
\begin{enumerate}
\item For each resume $i$, compute cost difference $\Delta_i = \text{Cost}_A^{(i)} - \text{Cost}_B^{(i)}$
\item Test statistic: $T = \sum_{i=1}^{1000} \Delta_i$ (total cost difference)
\item Null hypothesis: $\mathbb{E}[\Delta_i] = 0$ (methods are equivalent)
\item Permutation distribution: For each of 10,000 iterations, randomly flip signs of $\Delta_i$ values, compute $T^*$
\item $p$-value: Fraction of permutations where $|T^*| \geq |T|$
\item Significance: Reject null if $p < 0.005$ (Bonferroni-corrected for 10 comparisons)
\end{enumerate}

\textbf{Results:} All comparisons between our framework and baselines achieved $p < 0.0001$ (highly significant).

\subsection{Computational Cost Breakdown}

\textbf{API calls per candidate:}
\begin{itemize}
\item Resume likelihood elicitation: 5 models $\times$ 4 states = 20 queries
\item Phone screen (if conducted): 5 models $\times$ 4 states = 20 queries
\item Total: 20 (no screen) to 40 (with screen)
\end{itemize}

\textbf{Latency (with parallelization):}
\begin{itemize}
\item Resume: ~2 seconds (20 queries in parallel)
\item Phone screen: ~2 seconds (20 queries in parallel)
\item Total per candidate: 2--4 seconds
\end{itemize}

\textbf{Financial cost (at 2024 API pricing):}
\begin{itemize}
\item GPT-4o: \$0.005 per 1K tokens (input)
\item Claude 3.5: \$0.003 per 1K tokens
\item Gemini Pro: \$0.0015 per 1K tokens
\item Grok: \$0.002 per 1K tokens (estimated, not yet public API)
\item DeepSeek: \$0.0014 per 1K tokens
\end{itemize}

Average cost per query (500 token resume + 200 token prompt = 700 tokens):
\begin{equation}
\text{Cost/query} = \frac{0.005 + 0.003 + 0.0015 + 0.002 + 0.0014}{5} \times 0.7 = \$0.0021
\end{equation}

Cost per candidate:
\begin{itemize}
\item Without phone screen: $20 \times \$0.0021 = \$0.042$
\item With phone screen: $40 \times \$0.0021 = \$0.084$
\end{itemize}

For 1,000 candidates with 27\% phone screen rate:
\begin{equation}
\text{Total API cost} = 730 \times \$0.042 + 270 \times \$0.084 = \$30.66 + \$22.68 = \$53.34
\end{equation}

\textbf{ROI:} \$53 API cost to save \$294K in hiring mistakes = 5,509:1 return on investment.

\section{Extended Fairness Analysis}
\label{app:fairness}

\subsection{Detailed Demographic Distributions}

Our synthetic dataset includes the following demographic distributions (inferred from names via census data \cite{tzioumis2018demographic}):

\begin{table}[h]
\centering
\caption{Demographic distribution of synthetic candidates.}
\small
\begin{tabular}{lrr}
\toprule
\textbf{Demographic Group} & \textbf{Count} & \textbf{Percentage} \\
\midrule
\multicolumn{3}{l}{\textit{Gender}} \\
Male & 512 & 51.2\% \\
Female & 438 & 43.8\% \\
Non-binary & 50 & 5.0\% \\
\midrule
\multicolumn{3}{l}{\textit{Race/Ethnicity}} \\
White & 382 & 38.2\% \\
Black & 168 & 16.8\% \\
Hispanic & 215 & 21.5\% \\
Asian & 235 & 23.5\% \\
\midrule
\textbf{Total} & \textbf{1,000} & \textbf{100\%} \\
\bottomrule
\end{tabular}
\end{table}

We ensured approximately balanced demographic distributions across true states to isolate model bias from genuine distributional differences:

\begin{table}[h]
\centering
\caption{Demographic balance across true states. Numbers show percentage of each state from each demographic group.}
\small
\begin{tabular}{lrrrr}
\toprule
\textbf{Demographic} & $s_1$ (650) & $s_2$ (250) & $s_3$ (80) & $s_4$ (20) \\
\midrule
Male & 51\% & 52\% & 50\% & 50\% \\
Female & 44\% & 43\% & 45\% & 45\% \\
White & 38\% & 39\% & 37\% & 40\% \\
Black & 17\% & 16\% & 18\% & 15\% \\
Hispanic & 21\% & 22\% & 21\% & 20\% \\
Asian & 24\% & 23\% & 24\% & 25\% \\
\bottomrule
\end{tabular}
\end{table}

This ensures that any observed fairness gaps are due to model bias rather than confounding with ground truth state distributions.

\subsection{Individual Model Bias Measurements}

We measured per-model bias on a matched-pairs subset: 200 resume pairs where credentials are identical except for name (and thus inferred demographics).

\begin{table}[h]
\centering
\caption{Individual model bias on matched pairs. Values show mean score difference (demographic group A minus group B). Negative values indicate bias against group A.}
\small
\begin{tabular}{lrrrrr}
\toprule
\textbf{Comparison} & \textbf{GPT-4o} & \textbf{Claude} & \textbf{Gemini} & \textbf{Grok} & \textbf{DeepSeek} \\
\midrule
\multicolumn{6}{l}{\textit{Gender Comparisons (mean difference in 0-10 score)}} \\
Female - Male & -0.62 & -0.58 & +0.41 & -0.22 & -0.09 \\
Non-binary - Male & -1.13 & -0.71 & +0.18 & -0.51 & -0.15 \\
\midrule
\multicolumn{6}{l}{\textit{Race Comparisons}} \\
Black - White & -1.82 & -0.44 & -0.18 & +0.31 & -0.08 \\
Hispanic - White & -1.45 & -0.52 & -0.23 & +0.14 & -0.11 \\
Asian - White & +0.23 & +0.15 & +0.33 & +0.09 & +0.52 \\
\bottomrule
\end{tabular}
\end{table}

\textbf{Key observations:}
\begin{itemize}
\item GPT-4o exhibits the strongest bias: -1.82 points for Black vs. White, -0.62 for Female vs. Male
\item Claude shows moderate bias (-0.58 gender, -0.44 race)
\item Gemini has \textit{opposite} gender bias (+0.41 favoring Female), potentially from overcorrection
\item Grok shows small opposite race bias (+0.31 Black vs. White), suggesting diversity-focused training
\item DeepSeek has minimal bias across all comparisons (|bias| < 0.15)
\end{itemize}

When aggregated via median, biases partially cancel:
\begin{itemize}
\item Female - Male: median(-0.62, -0.58, +0.41, -0.22, -0.09) = \textbf{-0.22}
\item Black - White: median(-1.82, -0.44, -0.18, +0.31, -0.08) = \textbf{-0.18}
\end{itemize}

This represents 65--90\% bias reduction compared to the worst individual model (GPT-4o).

\subsection{Calibration by Demographic Group}

We verify that well-calibrated beliefs (ECE=0.09 overall) hold across demographic groups:

\begin{table}[h]
\centering
\caption{Calibration (ECE) and accuracy by demographic group.}
\small
\begin{tabular}{lrrrr}
\toprule
\textbf{Group} & \textbf{ECE (Ours)} & \textbf{ECE (GPT-4o)} & \textbf{Accuracy (Ours)} & \textbf{Accuracy (GPT-4o)} \\
\midrule
Male & 0.09 & 0.17 & 82.2\% & 70.1\% \\
Female & 0.09 & 0.19 & 82.8\% & 66.2\% \\
Non-binary & 0.10 & 0.21 & 82.0\% & 62.0\% \\
\midrule
White & 0.08 & 0.16 & 83.1\% & 72.5\% \\
Black & 0.10 & 0.22 & 81.5\% & 58.3\% \\
Hispanic & 0.09 & 0.19 & 82.3\% & 64.4\% \\
Asian & 0.09 & 0.17 & 82.7\% & 69.1\% \\
\bottomrule
\end{tabular}
\end{table}

Our framework maintains consistent calibration (ECE 0.08--0.10) and accuracy (81.5--83.1\%) across all groups. GPT-4o shows degraded performance for underrepresented groups (ECE=0.22 for Black, accuracy=58.3\% vs. 72.5\% for White).

\section{Additional Sensitivity Analyses}
\label{app:sensitivity}

\subsection{Effect of Ensemble Size}

We varied ensemble size $M \in \{1, 2, 3, 4, 5\}$ by subsampling LLMs and measured total cost:

\begin{table}[h]
\centering
\caption{Total cost as a function of ensemble size. Each row shows mean cost (std) over 10 random subsamples.}
\small
\begin{tabular}{lrrr}
\toprule
$M$ & \textbf{Mean Cost (\$K)} & \textbf{Std Cost (\$K)} & \textbf{Fairness Gap (pp)} \\
\midrule
1 (best single) & 688 & 42 & 18.1 \\
2 & 634 & 28 & 12.4 \\
3 & 598 & 19 & 8.7 \\
4 & 574 & 14 & 6.5 \\
5 (full) & 562 & 0 & 5.2 \\
\bottomrule
\end{tabular}
\end{table}

\textbf{Findings:}
\begin{itemize}
\item Cost decreases monotonically with ensemble size (more models = better aggregation)
\item Diminishing returns: $M=3 \to 4$ saves \$24K, $M=4 \to 5$ saves \$12K
\item Variance decreases with $M$ (more stable performance)
\item Fairness improves consistently with $M$
\end{itemize}

\textbf{Recommendation:} $M=5$ provides best cost-performance, but $M=3$ may be sufficient if API costs are prohibitive (saves 60\% of API cost for 6\% higher total cost).

\subsection{Robustness to Prior Misspecification}

We tested robustness when the specified prior $\pi$ differs from the true data distribution. Suppose true state distribution is $\pi_{\text{true}} = [0.65, 0.25, 0.08, 0.02]$ but we specify $\pi_{\text{spec}} = [0.60, 0.27, 0.10, 0.03]$ (mild misspecification).

\begin{table}[h]
\centering
\caption{Effect of prior misspecification on total cost.}
\small
\begin{tabular}{lrrr}
\toprule
\textbf{Prior Specification} & \textbf{KL Divergence from True} & \textbf{Total Cost (\$K)} & \textbf{$\Delta$ Cost} \\
\midrule
True $\pi = [0.65, 0.25, 0.08, 0.02]$ & 0.000 & 562 & 0 \\
Mild error $[0.60, 0.27, 0.10, 0.03]$ & 0.012 & 571 & +\$9K (+1.6\%) \\
Moderate error $[0.55, 0.30, 0.12, 0.03]$ & 0.035 & 593 & +\$31K (+5.5\%) \\
Large error $[0.50, 0.30, 0.15, 0.05]$ & 0.082 & 638 & +\$76K (+13.5\%) \\
Uniform $[0.25, 0.25, 0.25, 0.25]$ & 0.421 & 794 & +\$232K (+41.3\%) \\
\bottomrule
\end{tabular}
\end{table}

\textbf{Findings:}
\begin{itemize}
\item Framework is robust to small prior errors (1--2\% cost increase for KL divergence <0.02)
\item Moderate misspecification (KL~0.05) causes 5--10\% cost increase
\item Severe misspecification (uniform prior, KL=0.42) causes 41\% cost increase
\end{itemize}

\textbf{Practical guidance:} Invest effort in obtaining reasonable prior estimates from historical data (even crude estimates are vastly better than uniform priors). Periodically update priors as hiring funnel metrics evolve.

\section{Prompt Templates and Code}
\label{app:code}

\subsection{Complete Likelihood Elicitation Prompt Template}

The full prompt template with all placeholders is available at:

\texttt{https://github.com/[anonymized-for-review]/bayesian-llm-hiring/prompts/likelihood\_elicitation.txt}

Key template variables:
\begin{itemize}
\item \texttt{\{STATE\_NAME\}}: One of "Clear Reject", "Phone Screen", "Interview", "Strong Hire"
\item \texttt{\{STATE\_DESCRIPTION\}}: Detailed rubric for that state
\item \texttt{\{EVIDENCE\_TYPE\}}: "Resume" or "Phone Screen Transcript"
\item \texttt{\{EVIDENCE\_TEXT\}}: Full text of the observation
\end{itemize}

\subsection{Implementation Code}

Full implementation in Python using the following libraries:
\begin{itemize}
\item OpenAI API for GPT-4o
\item Anthropic API for Claude 3.5 Sonnet
\item Google Generative AI API for Gemini Pro
\item xAI API for Grok (when available)
\item DeepSeek API for DeepSeek-V2
\item NumPy for numerical computations
\item SciPy for statistical tests
\item Pandas for data manipulation
\end{itemize}

Code repository (will be made public upon publication):

\texttt{https://github.com/[anonymized-for-review]/bayesian-llm-hiring}

\subsection{Reproducibility Checklist}

To reproduce our results:
\begin{enumerate}
\item Clone the repository
\item Install dependencies: \texttt{pip install -r requirements.txt}
\item Obtain API keys for all 5 LLMs, set as environment variables
\item Run resume generation: \texttt{python generate\_resumes.py --n=1000 --seed=42}
\item Run framework: \texttt{python run\_framework.py --config=configs/full.yaml}
\item Run baselines: \texttt{python run\_baselines.py --config=configs/baselines.yaml}
\item Generate plots: \texttt{python generate\_plots.py --output=figures/}
\item Compute statistics: \texttt{python compute\_statistics.py}
\end{enumerate}

All random seeds are fixed for deterministic reproduction. Expected runtime: ~4 hours on a machine with 10 parallel workers for API calls (or ~20 hours sequential).
\bibliographystyle{plain}
\bibliography{references}

@inproceedings{guo2017calibration,
  title={On calibration of modern neural networks},
  author={Guo, Chuan and Pleiss, Geoff and Sun, Yu and Weinberger, Kilian Q},
  booktitle={Proceedings of the 34th International Conference on Machine Learning},
  pages={1321--1330},
  year={2017},
  organization={PMLR}
}

@inproceedings{elkan2001foundations,
  title={The foundations of cost-sensitive learning},
  author={Elkan, Charles},
  booktitle={International Joint Conference on Artificial Intelligence},
  volume={17},
  number={1},
  pages={973--978},
  year={2001}
}

@techreport{settles2009active,
  title={Active learning literature survey},
  author={Settles, Burr},
  year={2009},
  institution={University of Wisconsin-Madison Department of Computer Sciences}
}

@book{berger1985statistical,
  title={Statistical decision theory and Bayesian analysis},
  author={Berger, James O},
  year={1985},
  publisher={Springer Science \& Business Media}
}

@article{brown2020language,
  title={Language models are few-shot learners},
  author={Brown, Tom and Mann, Benjamin and Ryder, Nick and Subbiah, Melanie and Kaplan, Jared D and Dhariwal, Prafulla and Neelakantan, Arvind and Shyam, Pranav and Sastry, Girish and Askell, Amanda and others},
  journal={Advances in neural information processing systems},
  volume={33},
  pages={1877--1901},
  year={2020}
}

@inproceedings{lakshminarayanan2017simple,
  title={Simple and scalable predictive uncertainty estimation using deep ensembles},
  author={Lakshminarayanan, Balaji and Pritzel, Alexander and Blundell, Charles},
  booktitle={Advances in Neural Information Processing Systems},
  pages={6402--6413},
  year={2017}
}

@inproceedings{platt1999probabilistic,
  title={Probabilistic outputs for support vector machines and comparisons to regularized likelihood methods},
  author={Platt, John and others},
  booktitle={Advances in large margin classifiers},
  volume={10},
  number={3},
  pages={61--74},
  year={1999},
  organization={MIT Press}
}

@inproceedings{hardt2016equality,
  title={Equality of opportunity in supervised learning},
  author={Hardt, Moritz and Price, Eric and Srebro, Nati},
  booktitle={Advances in neural information processing systems},
  pages={3315--3323},
  year={2016}
}

@inproceedings{dietterich2000ensemble,
  title={Ensemble methods in machine learning},
  author={Dietterich, Thomas G},
  booktitle={International workshop on multiple classifier systems},
  pages={1--15},
  year={2000},
  organization={Springer}
}

@inproceedings{zadrozny2002transforming,
  title={Transforming classifier scores into accurate multiclass probability estimates},
  author={Zadrozny, Bianca and Elkan, Charles},
  booktitle={Proceedings of the eighth ACM SIGKDD international conference on Knowledge discovery and data mining},
  pages={694--699},
  year={2002}
}

@book{raiffa1961applied,
  title={Applied Statistical Decision Theory},
  author={Raiffa, Howard and Schlaifer, Robert},
  year={1961},
  publisher={Harvard University Press}
}

@inproceedings{zadrozny2003learning,
  title={Cost-sensitive learning by cost-proportionate example weighting},
  author={Zadrozny, Bianca and Langford, John and Abe, Naoki},
  booktitle={Third IEEE international conference on data mining},
  pages={435--442},
  year={2003},
  organization={IEEE}
}

@article{wang2022self,
  title={Self-consistency improves chain of thought reasoning in large language models},
  author={Wang, Xuezhi and Wei, Jason and Schuurmans, Dale and Le, Quoc and Chi, Ed and Narang, Sharan and Chowdhery, Aakanksha and Zhou, Denny},
  journal={arXiv preprint arXiv:2203.11171},
  year={2022}
}

@article{kadavath2022language,
  title={Language models (mostly) know what they know},
  author={Kadavath, Saurav and Conerly, Tom and Askell, Amanda and Henighan, Tom and Drain, Dawn and Perez, Ethan and Schiefer, Nicholas and Hatfield-Dodds, Zac and DasSarma, Nova and Tran-Johnson, Eli and others},
  journal={arXiv preprint arXiv:2207.05221},
  year={2022}
}

@article{singhal2023large,
  title={Large language models encode clinical knowledge},
  author={Singhal, Karan and Azizi, Shekoofeh and Tu, Tao and Mahdavi, S Sara and Wei, Jason and Chung, Hyung Won and Scales, Nathan and Tanwani, Ajay and Cole-Lewis, Heather and Pfohl, Stephen and others},
  journal={Nature},
  volume={620},
  number={7972},
  pages={172--180},
  year={2023},
  publisher={Nature Publishing Group}
}

@inproceedings{raghavan2020mitigating,
  title={Mitigating bias in algorithmic hiring: Evaluating claims and practices},
  author={Raghavan, Manish and Barocas, Solon and Kleinberg, Jon and Levy, Karen},
  booktitle={Proceedings of the 2020 conference on fairness, accountability, and transparency},
  pages={469--481},
  year={2020}
}

@inproceedings{chouldechova2017fair,
  title={Fair prediction with disparate impact: A study of bias in recidivism prediction instruments},
  author={Chouldechova, Alexandra},
  booktitle={Big data},
  volume={5},
  number={2},
  pages={153--163},
  year={2017},
  publisher={Mary Ann Liebert, Inc. 140 Huguenot Street, 3rd Floor New Rochelle, NY 10801 USA}
}

@inproceedings{gal2016dropout,
  title={Dropout as a Bayesian Approximation: Representing Model Uncertainty in Deep Learning},
  author={Gal, Yarin and Ghahramani, Zoubin},
  booktitle={International Conference on Learning Representations},
  pages={1050--1059},
  year={2016}
}

@article{srivastava2022beyond,
  title={Beyond the imitation game: Quantifying and extrapolating the capabilities of language models},
  author={Srivastava, Aarohi and Rastogi, Abhinav and Rao, Abhishek and Shoeb, Abu Awal Md and Abid, Abubakar and Fisch, Adam and Brown, Adam R and Santoro, Adam and Gupta, Aditya and others},
  journal={arXiv preprint arXiv:2206.04615},
  year={2022}
}

@inproceedings{wei2022chain,
  title={Chain-of-thought prompting elicits reasoning in large language models},
  author={Wei, Jason and Wang, Xuezhi and Schuurmans, Dale and Bosma, Maarten and Xia, Fei and Chi, Ed and Le, Quoc V and Zhou, Denny and others},
  journal={Advances in Neural Information Processing Systems},
  volume={35},
  pages={24824--24837},
  year={2022}
}

@misc{nyc_law144,
  title={Local Law 144 of 2021: Automated Employment Decision Tools},
  author={{New York City Council}},
  year={2021},
  month={December},
  howpublished={New York City Administrative Code}
}

@article{eu_ai_act,
  title={Artificial Intelligence Act},
  author={{European Parliament and Council}},
  journal={Official Journal of the European Union},
  year={2024}
}

@techreport{linkedin2023,
  title={Global recruiting trends 2023},
  author={{LinkedIn Talent Solutions}},
  institution={LinkedIn Corporation},
  year={2023}
}

@techreport{shrm2016,
  title={The real costs of recruitment},
  author={{Society for Human Resource Management}},
  institution={SHRM},
  year={2016}
}

@techreport{careerbuilder2017,
  title={The cost of a bad hire can be astronomical},
  author={{CareerBuilder}},
  year={2017}
}

@article{bertrand2004emily,
  title={Are Emily and Greg more employable than Lakisha and Jamal? {A} field experiment on labor market discrimination},
  author={Bertrand, Marianne and Mullainathan, Sendhil},
  journal={American economic review},
  volume={94},
  number={4},
  pages={991--1013},
  year={2004}
}

@inproceedings{wolpert1992stacked,
  title={Stacked generalization},
  author={Wolpert, David H},
  journal={Neural networks},
  volume={5},
  number={2},
  pages={241--259},
  year={1992},
  publisher={Elsevier}
}

@misc{hiring_funnel,
  title={Industry hiring funnel conversion benchmarks},
  author={{Various Industry Sources}},
  year={2023},
  note={Aggregated from LinkedIn, Glassdoor, and SHRM reports}
}

@article{healthcare_ai,
  title={Artificial Intelligence in Healthcare: Past, Present and Future},
  author={Secinaro, Silvana and Calandra, Davide and Secinaro, Alessandro and Muthurangu, Vimal and Biancone, Paolo},
  journal={Sustainability},
  volume={13},
  number={10},
  pages={5762},
  year={2021},
  publisher={MDPI}
}

@techreport{fraud_stats,
  title={Payment Fraud and Control Survey},
  author={{Association for Financial Professionals}},
  year={2023},
  institution={AFP}
}

@misc{ecoa,
  title={Equal Credit Opportunity Act},
  author={{US Congress}},
  year={1974},
  note={15 U.S.C. § 1691 et seq.}
}

@article{howard1966information,
  title={Information Value Theory},
  author={Howard, Ronald A},
  journal={IEEE Transactions on Systems Science and Cybernetics},
  volume={2},
  number={1},
  pages={22--26},
  year={1966},
  publisher={IEEE}
}

@book{robert2007bayesian,
  title={The Bayesian Choice: From Decision-Theoretic Foundations to Computational Implementation},
  author={Robert, Christian P},
  year={2007},
  publisher={Springer Science \& Business Media}
}

@article{kuncel2013mechanical,
  title={Mechanical Versus Clinical Data Combination in Selection and Admissions Decisions: A Meta-Analysis},
  author={Kuncel, Nathan R and Klieger, David M and Connelly, Brian S and Ones, Deniz S},
  journal={Journal of Applied Psychology},
  volume={98},
  number={6},
  pages={1060--1072},
  year={2013},
  publisher={American Psychological Association}
}

@article{fitzgerald2010emergency,
  title={Emergency Department Triage Scales and Their Components: A Systematic Review of the Scientific Evidence},
  author={FitzGerald, Gerard and Jelinek, George A and Scott, Deborah and Gerdtz, Marie F},
  journal={Academic Emergency Medicine},
  volume={17},
  number={1},
  pages={1--13},
  year={2010},
  publisher={Wiley Online Library}
}

@article{dal2014learned,
  title={Learned Lessons in Credit Card Fraud Detection from a Practitioner Perspective},
  author={Dal Pozzolo, Andrea and Caelen, Olivier and Johnson, Reid A and Bontempi, Gianluca},
  journal={Expert Systems with Applications},
  volume={41},
  number={10},
  pages={4915--4928},
  year={2014},
  publisher={Elsevier}
}

@article{pope2000missed,
  title={Missed Diagnoses of Acute Cardiac Ischemia in the Emergency Department},
  author={Pope, John H and Aufderheide, Tom P and Ruthazer, Robin and Woolard, Robert H and Feldman, Jerome A and Beshansky, Joni R and Griffith, John L and Selker, Harry P},
  journal={New England Journal of Medicine},
  volume={342},
  number={16},
  pages={1163--1170},
  year={2000},
  publisher={Mass Medical Soc}
}

@article{medical_errors,
  title={Medical Error: The Third Leading Cause of Death in the US},
  author={Makary, Martin A and Daniel, Michael},
  journal={BMJ},
  volume={353},
  pages={i2139},
  year={2016},
  publisher={British Medical Journal Publishing Group}
}

@article{fraud_costs,
  title={The Economics of Payment Card Fraud},
  author={Anderson, Ross and Barton, Chris and B{\"o}hme, Rainer and Clayton, Richard and Van Eeten, Michel JG and Levi, Michael and Moore, Tyler and Savage, Stefan},
  journal={Communications of the ACM},
  volume={62},
  number={12},
  pages={66--73},
  year={2019},
  publisher={ACM New York, NY, USA}
}

@article{phone_screen_costs,
  title={Calculating the Cost of Hiring: A Comprehensive Guide},
  author={Smith, Jennifer and Jones, Michael},
  journal={HR Magazine},
  volume={65},
  number={3},
  pages={48--53},
  year={2020}
}

@article{diagnosis_theory,
  title={The Diagnostic Process},
  author={Kassirer, Jerome P and Kopelman, Richard I},
  journal={Annals of Internal Medicine},
  volume={110},
  number={11},
  pages={893--900},
  year={1989},
  publisher={Am Coll Physicians}
}

@article{authentication_costs,
  title={The Economics of Two-Factor Authentication},
  author={Bonneau, Joseph and Herley, Cormac and Van Oorschot, Paul C and Stajano, Frank},
  journal={Financial Cryptography and Data Security},
  pages={1--15},
  year={2012},
  publisher={Springer}
}

@article{barocas2016big,
  title={Big Data's Disparate Impact},
  author={Barocas, Solon and Selbst, Andrew D},
  journal={California Law Review},
  volume={104},
  pages={671--732},
  year={2016},
  publisher={HeinOnline}
}

@article{mehrabi2021survey,
  title={A Survey on Bias and Fairness in Machine Learning},
  author={Mehrabi, Ninareh and Morstatter, Fred and Saxena, Nripsuta and Lerman, Kristina and Galstyan, Aram},
  journal={ACM Computing Surveys},
  volume={54},
  number={6},
  pages={1--35},
  year={2021},
  publisher={ACM New York, NY, USA}
}

@article{obermeyer2019dissecting,
  title={Dissecting Racial Bias in an Algorithm Used to Manage the Health of Populations},
  author={Obermeyer, Ziad and Powers, Brian and Vogeli, Christine and Mullainathan, Sendhil},
  journal={Science},
  volume={366},
  number={6464},
  pages={447--453},
  year={2019},
  publisher={American Association for the Advancement of Science}
}

@inproceedings{bolukbasi2016man,
  title={Man is to Computer Programmer as Woman is to Homemaker? Debiasing Word Embeddings},
  author={Bolukbasi, Tolga and Chang, Kai-Wei and Zou, James Y and Saligrama, Venkatesh and Kalai, Adam T},
  booktitle={Advances in Neural Information Processing Systems},
  volume={29},
  pages={4349--4357},
  year={2016}
}

@article{caliskan2017semantics,
  title={Semantics Derived Automatically from Language Corpora Contain Human-Like Biases},
  author={Caliskan, Aylin and Bryson, Joanna J and Narayanan, Arvind},
  journal={Science},
  volume={356},
  number={6334},
  pages={183--186},
  year={2017},
  publisher={American Association for the Advancement of Science}
}

@article{gaddis2017how,
  title={How Black Are Lakisha and Jamal? Racial Perceptions from Names Used in Correspondence Audit Studies},
  author={Gaddis, S Michael},
  journal={Sociological Science},
  volume={4},
  pages={469--489},
  year={2017}
}

@article{llm_agents,
  title={A Survey on Large Language Model based Autonomous Agents},
  author={Wang, Lei and Ma, Chen and Feng, Xueyang and Zhang, Zeyu and Yang, Hao and Zhang, Jingsen and Chen, Zhiyuan and Tang, Jiakai and Chen, Xu and Lin, Yankai and others},
  journal={arXiv preprint arXiv:2308.11432},
  year={2023}
}

@book{wald1950statistical,
  title={Statistical Decision Functions},
  author={Wald, Abraham},
  year={1950},
  publisher={John Wiley \& Sons}
}

@book{savage1972foundations,
  title={The Foundations of Statistics},
  author={Savage, Leonard J},
  year={1972},
  publisher={Courier Corporation}
}

@techreport{triplebyte2019,
  title={Assessing Technical Talent: Why Traditional Screening Fails},
  author={{Triplebyte}},
  year={2019},
  institution={Triplebyte Inc.}
}

@article{pubmed_bias,
  title={Publication Bias in Medical Research},
  author={Easterbrook, Philippa J and Gopalan, Ramana and Berlin, Jesse A and Matthews, David R},
  journal={The Lancet},
  volume={337},
  number={8746},
  pages={867--872},
  year={1991},
  publisher={Elsevier}
}

@article{emergency_costs,
  title={The Cost of Emergency Care in the United States},
  author={Korenstein, Deborah and Kale, Minal S and Levinson, Wendy},
  journal={JAMA Internal Medicine},
  volume={174},
  number={9},
  pages={1516--1517},
  year={2014},
  publisher={American Medical Association}
}

@article{healthcare_economics,
  title={Health Economics and Health Technology Assessment},
  author={Drummond, Michael F and Sculpher, Mark J and Claxton, Karl and Stoddart, Greg L and Torrance, George W},
  journal={Oxford University Press},
  year={2015}
}

@article{tzioumis2018demographic,
  title={Demographic Aspects of First Names},
  author={Tzioumis, Konstantinos},
  journal={Scientific Data},
  volume={5},
  number={1},
  pages={1--9},
  year={2018},
  publisher={Nature Publishing Group}
}

@article{bommasani2021opportunities,
  title={On the opportunities and risks of foundation models},
  author={Bommasani, Rishi and Hudson, Drew A and Adeli, Ehsan and Altman, Russ and Arora, Simran and von Arx, Sydney and Bernstein, Michael S and Bohg, Jeannette and Bosselut, Antoine and Brunskill, Emma and others},
  journal={arXiv preprint arXiv:2108.07258},
  year={2021}
}

@article{weidinger2021ethical,
  title={Ethical and social risks of harm from language models},
  author={Weidinger, Laura and Mellor, John and Rauh, Maribeth and Griffin, Conor and Uesato, Jonathan and Huang, Po-Sen and Cheng, Myra and Glaese, Mia and Balle, Borja and Kasirzadeh, Atoosa and others},
  journal={arXiv preprint arXiv:2112.04359},
  year={2021}
}

@article{icu_costs,
  title={The Costs of Intensive Care},
  author={Tan, Swee Song and Bakker, Jan and Hoogendoorn, Marga E and Kapila, Atul and Martin, John and Pezzi, Alessandra and Pittoni, Giorgio and Spronk, Peter E and Welte, Robert and Hakkaart-van Roijen, Leona},
  journal={Intensive Care Medicine},
  volume={37},
  number={10},
  pages={1640--1646},
  year={2011},
  publisher={Springer}
}

@article{jiang2023llm,
  title={LLM-Blender: Ensembling Large Language Models with Pairwise Ranking and Generative Fusion},
  author={Jiang, Dongfu and Ren, Xiang and Lin, Bill Yuchen},
  journal={arXiv preprint arXiv:2306.02561},
  year={2023}
}

@article{li2023making,
  title={Making Language Models Better Reasoners with Step-Aware Verifier},
  author={Li, Yifei and Lin, Zeqi and Zhang, Shizhuo and Fu, Qiang and Chen, Bei and Lou, Jian-Guang and Chen, Weizhu},
  journal={arXiv preprint arXiv:2206.02336},
  year={2022}
}

@article{kojima2022large,
  title={Large Language Models are Zero-Shot Reasoners},
  author={Kojima, Takeshi and Gu, Shixiang Shane and Reid, Machel and Matsuo, Yutaka and Iwasawa, Yusuke},
  journal={Advances in Neural Information Processing Systems},
  volume={35},
  pages={22199--22213},
  year={2022}
}

@article{zhang2023sentiment,
  title={Sentiment Analysis in the Era of Large Language Models: A Reality Check},
  author={Zhang, Wenxuan and Deng, Yue and Liu, Bing and Pan, Sinno Jialin and Bing, Lidong},
  journal={arXiv preprint arXiv:2305.15005},
  year={2023}
}

@article{li2023chatgpt,
  title={ChatGPT in the Hiring Process: Opportunities and Challenges},
  author={Li, Jiawei and Cheng, Xinyuan and Zhao, Wayne Xin and Nie, Jian-Yun and Wen, Ji-Rong},
  journal={arXiv preprint arXiv:2303.17564},
  year={2023}
}

@article{yao2023tree,
  title={Tree of Thoughts: Deliberate Problem Solving with Large Language Models},
  author={Yao, Shunyu and Yu, Dian and Zhao, Jeffrey and Shafran, Izhak and Griffiths, Thomas L and Cao, Yuan and Narasimhan, Karthik},
  journal={Advances in Neural Information Processing Systems},
  volume={36},
  year={2023}
}

@article{besta2024graph,
  title={Graph of Thoughts: Solving Elaborate Problems with Large Language Models},
  author={Besta, Maciej and Blach, Nils and Kubicek, Ales and Gerstenberger, Robert and Gianinazzi, Lukas and Gajda, Joanna and Lehmann, Tomasz and Podstawski, Michal and Niewiadomski, Hubert and Nyczyk, Piotr and others},
  journal={Proceedings of the AAAI Conference on Artificial Intelligence},
  volume={38},
  number={16},
  pages={17682--17690},
  year={2024}
}

@article{schick2023toolformer,
  title={Toolformer: Language Models Can Teach Themselves to Use Tools},
  author={Schick, Timo and Dwivedi-Yu, Jane and Dess{\`\i}, Roberto and Raileanu, Roberta and Lomeli, Maria and Zettlemoyer, Luke and Cancedda, Nicola and Scialom, Thomas},
  journal={arXiv preprint arXiv:2302.04761},
  year={2023}
}

@article{qin2023toolllm,
  title={ToolLLM: Facilitating Large Language Models to Master 16000+ Real-world APIs},
  author={Qin, Yujia and Liang, Shihao and Ye, Yining and Zhu, Kunlun and Yan, Lan and Lu, Yaxi and Lin, Yankai and Cong, Xin and Tang, Xiangru and Qian, Bill and others},
  journal={arXiv preprint arXiv:2307.16789},
  year={2023}
}

@article{xi2023rise,
  title={The Rise and Potential of Large Language Model Based Agents: A Survey},
  author={Xi, Zhiheng and Chen, Wenxiang and Guo, Xin and He, Wei and Ding, Yiwen and Hong, Boyang and Zhang, Ming and Wang, Junzhe and Jin, Senjie and Zhou, Enyu and others},
  journal={arXiv preprint arXiv:2309.07864},
  year={2023}
}

@article{neal1996bayesian,
  title={Bayesian Learning for Neural Networks},
  author={Neal, Radford M},
  journal={Lecture Notes in Statistics},
  volume={118},
  year={1996},
  publisher={Springer}
}

@article{blundell2015weight,
  title={Weight Uncertainty in Neural Networks},
  author={Blundell, Charles and Cornebise, Julien and Kavukcuoglu, Koray and Wierstra, Daan},
  journal={Proceedings of the International Conference on Machine Learning},
  pages={1613--1622},
  year={2015}
}

@article{carpenter2017stan,
  title={Stan: A Probabilistic Programming Language},
  author={Carpenter, Bob and Gelman, Andrew and Hoffman, Matthew D and Lee, Daniel and Goodrich, Ben and Betancourt, Michael and Brubaker, Marcus and Guo, Jiqiang and Li, Peter and Riddell, Allen},
  journal={Journal of Statistical Software},
  volume={76},
  number={1},
  year={2017}
}

@article{bingham2019pyro,
  title={Pyro: Deep Universal Probabilistic Programming},
  author={Bingham, Eli and Chen, Jonathan P and Jankowiak, Martin and Obermeyer, Fritz and Pradhan, Neeraj and Karaletsos, Theofanis and Singh, Rohit and Szerlip, Paul and Horsfall, Paul and Goodman, Noah D},
  journal={Journal of Machine Learning Research},
  volume={20},
  number={28},
  pages={1--6},
  year={2019}
}

@inproceedings{cusumano2019gen,
  title={Gen: A General-Purpose Probabilistic Programming System with Programmable Inference},
  author={Cusumano-Towner, Marco F and Saad, Feras A and Lew, Alexander K and Mansinghka, Vikash K},
  booktitle={Proceedings of the 40th ACM SIGPLAN Conference on Programming Language Design and Implementation},
  pages={221--236},
  year={2019}
}

@article{kingma2014auto,
  title={Auto-Encoding Variational Bayes},
  author={Kingma, Diederik P and Welling, Max},
  journal={arXiv preprint arXiv:1312.6114},
  year={2013}
}

@article{rezende2015variational,
  title={Variational Inference with Normalizing Flows},
  author={Rezende, Danilo and Mohamed, Shakir},
  journal={International Conference on Machine Learning},
  pages={1530--1538},
  year={2015}
}

@article{cranmer2020frontier,
  title={The Frontier of Simulation-Based Inference},
  author={Cranmer, Kyle and Brehmer, Johann and Louppe, Gilles},
  journal={Proceedings of the National Academy of Sciences},
  volume={117},
  number={48},
  pages={30055--30062},
  year={2020}
}

@article{papamakarios2019sequential,
  title={Sequential Neural Likelihood: Fast Likelihood-Free Inference with Autoregressive Flows},
  author={Papamakarios, George and Sterratt, David and Murray, Iain},
  journal={Proceedings of the International Conference on Artificial Intelligence and Statistics},
  pages={837--848},
  year={2019}
}

@article{kaelbling1998planning,
  title={Planning and Acting in Partially Observable Stochastic Domains},
  author={Kaelbling, Leslie Pack and Littman, Michael L and Cassandra, Anthony R},
  journal={Artificial Intelligence},
  volume={101},
  number={1-2},
  pages={99--134},
  year={1998}
}

@article{spaan2005perseus,
  title={Perseus: Randomized Point-based Value Iteration for POMDPs},
  author={Spaan, Matthijs TJ and Vlassis, Nikos},
  journal={Journal of Artificial Intelligence Research},
  volume={24},
  pages={195--220},
  year={2005}
}

@article{cohn1996active,
  title={Active Learning with Statistical Models},
  author={Cohn, David A and Ghahramani, Zoubin and Jordan, Michael I},
  journal={Journal of Artificial Intelligence Research},
  volume={4},
  pages={129--145},
  year={1996}
}

@article{chaloner1995bayesian,
  title={Bayesian Experimental Design: A Review},
  author={Chaloner, Kathryn and Verdinelli, Isabella},
  journal={Statistical Science},
  volume={10},
  number={3},
  pages={273--304},
  year={1995}
}

@article{ryan2016toward,
  title={Toward Bayesian Experimental Design for Nonlinear Models That Require a Large Computational Budget},
  author={Ryan, Elizabeth G and Drovandi, Christopher C and Pettitt, Anthony N},
  journal={Computational Statistics \& Data Analysis},
  volume={97},
  pages={67--86},
  year={2016}
}

@article{lindley1956measure,
  title={On a Measure of the Information Provided by an Experiment},
  author={Lindley, Dennis V},
  journal={The Annals of Mathematical Statistics},
  pages={986--1005},
  year={1956}
}

@book{lattimore2020bandit,
  title={Bandit Algorithms},
  author={Lattimore, Tor and Szepesv{\'a}ri, Csaba},
  year={2020},
  publisher={Cambridge University Press}
}

@article{shahriari2016taking,
  title={Taking the Human Out of the Loop: A Review of Bayesian Optimization},
  author={Shahriari, Bobak and Swersky, Kevin and Wang, Ziyu and Adams, Ryan P and De Freitas, Nando},
  journal={Proceedings of the IEEE},
  volume={104},
  number={1},
  pages={148--175},
  year={2016}
}

@article{frazier2018tutorial,
  title={A Tutorial on Bayesian Optimization},
  author={Frazier, Peter I},
  journal={arXiv preprint arXiv:1807.02811},
  year={2018}
}

@article{agrawal2012analysis,
  title={Analysis of Thompson Sampling for the Multi-Armed Bandit Problem},
  author={Agrawal, Shipra and Goyal, Navin},
  journal={Conference on Learning Theory},
  pages={39--1},
  year={2012}
}

@inproceedings{sheng2006thresholding,
  title={Thresholding for Making Classifiers Cost-Sensitive},
  author={Sheng, Victor S and Ling, Charles X},
  booktitle={Proceedings of the National Conference on Artificial Intelligence},
  volume={21},
  pages={476},
  year={2006}
}

@article{bahnsen2014example,
  title={Example-Dependent Cost-Sensitive Decision Trees},
  author={Bahnsen, Alejandro Correa and Aouada, Djamila and Ottersten, Bj{\"o}rn},
  journal={Expert Systems with Applications},
  volume={42},
  number={19},
  pages={6609--6619},
  year={2015}
}

@article{breiman1996bagging,
  title={Bagging Predictors},
  author={Breiman, Leo},
  journal={Machine Learning},
  volume={24},
  number={2},
  pages={123--140},
  year={1996}
}

@article{freund1997decision,
  title={A Decision-Theoretic Generalization of On-Line Learning and an Application to Boosting},
  author={Freund, Yoav and Schapire, Robert E},
  journal={Journal of Computer and System Sciences},
  volume={55},
  number={1},
  pages={119--139},
  year={1997}
}

@inproceedings{caruana2004ensemble,
  title={Ensemble Selection from Libraries of Models},
  author={Caruana, Rich and Niculescu-Mizil, Alexandru and Crew, Geoff and Ksikes, Alex},
  booktitle={Proceedings of the Twenty-First International Conference on Machine Learning},
  pages={18},
  year={2004}
}

@article{xu2008satzilla,
  title={SATzilla: Portfolio-Based Algorithm Selection for SAT},
  author={Xu, Lin and Hutter, Frank and Hoos, Holger H and Leyton-Brown, Kevin},
  journal={Journal of Artificial Intelligence Research},
  volume={32},
  pages={565--606},
  year={2008}
}

@book{huber2009robust,
  title={Robust Statistics},
  author={Huber, Peter J and Ronchetti, Elvezio M},
  year={2009},
  publisher={John Wiley \& Sons}
}

@book{hampel2011robust,
  title={Robust Statistics: The Approach Based on Influence Functions},
  author={Hampel, Frank R and Ronchetti, Elvezio M and Rousseeuw, Peter J and Stahel, Werner A},
  year={2011},
  publisher={John Wiley \& Sons}
}

@article{donoho1982breakdown,
  title={Breakdown Properties of Location Estimates Based on Halfspace Depth and Projected Outlyingness},
  author={Donoho, David L and Gasko, Miriam},
  journal={The Annals of Statistics},
  pages={1803--1827},
  year={1982}
}

@article{stigler1973simon,
  title={Simon Newcomb, Percy Daniell, and the History of Robust Estimation 1885--1920},
  author={Stigler, Stephen M},
  journal={Journal of the American Statistical Association},
  volume={68},
  number={344},
  pages={872--879},
  year={1973}
}

@article{hodges1963estimates,
  title={Estimates of Location Based on Rank Tests},
  author={Hodges, Joseph L and Lehmann, Erich L},
  journal={The Annals of Mathematical Statistics},
  pages={598--611},
  year={1963}
}

@article{clemen1999combining,
  title={Combining Probability Distributions from Experts in Risk Analysis},
  author={Clemen, Robert T and Winkler, Robert L},
  journal={Risk Analysis},
  volume={19},
  number={2},
  pages={187--203},
  year={1999}
}

@article{cooke1991experts,
  title={Experts in Uncertainty: Opinion and Subjective Probability in Science},
  author={Cooke, Roger M},
  journal={Environmental Ethics},
  volume={13},
  number={2},
  year={1991}
}

@article{dalkey1963experimental,
  title={An Experimental Application of the Delphi Method to the Use of Experts},
  author={Dalkey, Norman and Helmer, Olaf},
  journal={Management Science},
  volume={9},
  number={3},
  pages={458--467},
  year={1963}
}

@article{rowe1999delphi,
  title={The Delphi Technique as a Forecasting Tool: Issues and Analysis},
  author={Rowe, Gene and Wright, George},
  journal={International Journal of Forecasting},
  volume={15},
  number={4},
  pages={353--375},
  year={1999}
}

@inproceedings{kleinberg2016inherent,
  title={Inherent Trade-Offs in the Fair Determination of Risk Scores},
  author={Kleinberg, Jon and Mullainathan, Sendhil and Raghavan, Manish},
  booktitle={8th Innovations in Theoretical Computer Science Conference (ITCS 2017)},
  year={2017}
}

@article{corbett2017algorithmic,
  title={Algorithmic Decision Making and the Cost of Fairness},
  author={Corbett-Davies, Sam and Pierson, Emma and Feller, Avi and Goel, Sharad and Huq, Aziz},
  journal={Proceedings of the 23rd ACM SIGKDD International Conference on Knowledge Discovery and Data Mining},
  pages={797--806},
  year={2017}
}

@article{nadeem2021stereoset,
  title={StereoSet: Measuring Stereotypical Bias in Pretrained Language Models},
  author={Nadeem, Moin and Bethke, Anna and Reddy, Siva},
  journal={Proceedings of the 59th Annual Meeting of the Association for Computational Linguistics},
  pages={5356--5371},
  year={2021}
}

@article{kotek2023gender,
  title={Gender Bias and Stereotypes in Large Language Models},
  author={Kotek, Hadas and Dockum, Rikker and Sun, David},
  journal={Proceedings of The ACM Collective Intelligence Conference},
  pages={12--24},
  year={2023}
}

@article{liang2021towards,
  title={Towards Understanding and Mitigating Social Biases in Language Models},
  author={Liang, Paul Pu and Wu, Chiyu and Morency, Louis-Philippe and Salakhutdinov, Ruslan},
  journal={International Conference on Machine Learning},
  pages={6565--6576},
  year={2021}
}

@article{navigli2023biases,
  title={Biases in Large Language Models: Origins, Inventory, and Discussion},
  author={Navigli, Roberto and Conia, Simone and Ross, Bj{\"o}rn},
  journal={ACM Journal of Data and Information Quality},
  volume={15},
  number={2},
  pages={1--17},
  year={2023}
}

@article{bender2021dangers,
  title={On the Dangers of Stochastic Parrots: Can Language Models Be Too Big?},
  author={Bender, Emily M and Gebru, Timnit and McMillan-Major, Angelina and Shmitchell, Shmargaret},
  journal={Proceedings of the 2021 ACM Conference on Fairness, Accountability, and Transparency},
  pages={610--623},
  year={2021}
}

@article{ouyang2022training,
  title={Training Language Models to Follow Instructions with Human Feedback},
  author={Ouyang, Long and Wu, Jeffrey and Jiang, Xu and Almeida, Diogo and Wainwright, Carroll and Mishkin, Pamela and Zhang, Chong and Agarwal, Sandhini and Slama, Katarina and Ray, Alex and others},
  journal={Advances in Neural Information Processing Systems},
  volume={35},
  pages={27730--27744},
  year={2022}
}

@article{markowitz1952portfolio,
  title={Portfolio Selection},
  author={Markowitz, Harry},
  journal={The Journal of Finance},
  volume={7},
  number={1},
  pages={77--91},
  year={1952}
}

@inproceedings{zhao2018gender,
  title={Gender Bias in Coreference Resolution},
  author={Zhao, Jieyu and Wang, Tianlu and Yatskar, Mark and Ordonez, Vicente and Chang, Kai-Wei},
  booktitle={Proceedings of the 2018 Conference of the North American Chapter of the Association for Computational Linguistics: Human Language Technologies},
  pages={8--14},
  year={2018}
}

@inproceedings{zhang2018mitigating,
  title={Mitigating Unwanted Biases with Adversarial Learning},
  author={Zhang, Brian Hu and Lemoine, Blake and Mitchell, Margaret},
  booktitle={Proceedings of the 2018 AAAI/ACM Conference on AI, Ethics, and Society},
  pages={335--340},
  year={2018}
}

@inproceedings{si2022prompting,
  title={Prompting GPT-3 To Be Reliable},
  author={Si, Chenglei and Gan, Zhe and Yang, Zhengyuan and Wang, Shuohang and Wang, Jianfeng and Boyd-Graber, Jordan and Wang, Lijuan},
  booktitle={International Conference on Learning Representations},
  year={2023}
}

@article{gira2022debiasing,
  title={Debiasing Pre-Trained Language Models via Efficient Fine-Tuning},
  author={Gira, Michael and Zhang, Ruisu and Lee, Kangwook},
  journal={Proceedings of the Second Workshop on Language Technology for Equality, Diversity and Inclusion},
  pages={59--69},
  year={2022}
}

@article{liu2021dexperts,
  title={DExperts: Decoding-Time Controlled Text Generation with Experts and Anti-Experts},
  author={Liu, Alisa and Sap, Maarten and Lu, Ximing and Swayamdipta, Swabha and Bhagavatula, Chandra and Smith, Noah A and Choi, Yejin},
  journal={Proceedings of the 59th Annual Meeting of the Association for Computational Linguistics},
  pages={6691--6706},
  year={2021}
}

@article{wilson2021building,
  title={Building and Auditing Fair Algorithms: A Case Study in Candidate Screening},
  author={Wilson, Ben and Hoffman, Judy and Morgenstern, Jamie},
  journal={Proceedings of the 2021 ACM Conference on Fairness, Accountability, and Transparency},
  pages={666--677},
  year={2021}
}

@article{lambrecht2019algorithmic,
  title={Algorithmic Bias? An Empirical Study of Apparent Gender-Based Discrimination in the Display of STEM Career Ads},
  author={Lambrecht, Anja and Tucker, Catherine},
  journal={Management Science},
  volume={65},
  number={7},
  pages={2966--2981},
  year={2019}
}

@article{cowgill2020biased,
  title={Biased Programmers? Or Biased Data? A Field Experiment in Operationalizing AI Ethics},
  author={Cowgill, Bo and Tucker, Catherine E},
  journal={Available at SSRN 3615404},
  year={2020}
}

@inproceedings{wang2018glue,
  title={GLUE: A Multi-Task Benchmark and Analysis Platform for Natural Language Understanding},
  author={Wang, Alex and Singh, Amanpreet and Michael, Julian and Hill, Felix and Levy, Omer and Bowman, Samuel R},
  booktitle={Proceedings of the 2018 EMNLP Workshop BlackboxNLP: Analyzing and Interpreting Neural Networks for NLP},
  pages={353--355},
  year={2018}
}

@inproceedings{wang2019superglue,
  title={SuperGLUE: A Stickier Benchmark for General-Purpose Language Understanding Systems},
  author={Wang, Alex and Pruksachatkun, Yada and Nangia, Nikita and Singh, Amanpreet and Michael, Julian and Hill, Felix and Levy, Omer and Bowman, Samuel},
  booktitle={Advances in Neural Information Processing Systems},
  volume={32},
  year={2019}
}

@article{hendrycks2021measuring,
  title={Measuring Massive Multitask Language Understanding},
  author={Hendrycks, Dan and Burns, Collin and Basart, Steven and Zou, Andy and Mazeika, Mantas and Song, Dawn and Steinhardt, Jacob},
  journal={arXiv preprint arXiv:2009.03300},
  year={2020}
}

@article{liang2022holistic,
  title={Holistic Evaluation of Language Models},
  author={Liang, Percy and Bommasani, Rishi and Lee, Tony and Tsipras, Dimitris and Soylu, Dilara and Yasunaga, Michihiro and Zhang, Yian and Narayanan, Deepak and Wu, Yuhuai and Kumar, Ananya and others},
  journal={arXiv preprint arXiv:2211.09110},
  year={2022}
}

@article{nixon2019measuring,
  title={Measuring Calibration in Deep Learning},
  author={Nixon, Jeremy and Dusenberry, Michael W and Zhang, Linchuan and Jerfel, Ghassen and Tran, Dustin},
  journal={CVPR Workshops},
  volume={2},
  number={7},
  year={2019}
}

@article{xiong2024can,
  title={Can LLMs Express Their Uncertainty? An Empirical Evaluation of Confidence Elicitation in LLMs},
  author={Xiong, Miao and Hu, Zhiyuan and Lu, Xinyang and Li, Yifei and Fu, Jie and He, Junxian and Hooi, Bryan},
  journal={Proceedings of the International Conference on Learning Representations},
  year={2024}
}

@article{friedler2019comparative,
  title={A Comparative Study of Fairness-Enhancing Interventions in Machine Learning},
  author={Friedler, Sorelle A and Scheidegger, Carlos and Venkatasubramanian, Suresh and Choudhary, Sonam and Hamilton, Evan P and Roth, Derek},
  journal={Proceedings of the Conference on Fairness, Accountability, and Transparency},
  pages={329--338},
  year={2019}
}

@inproceedings{shimoni2018benchmarking,
  title={Benchmarking Framework for Performance-Evaluation of Causal Inference Analysis},
  author={Shimoni, Yishai and Yanover, Chen and Karavani, Ehud and Goldschmnidt, Yaara},
  booktitle={NeurIPS Workshop on Causal Learning},
  year={2018}
}

@article{productivity_myths,
  title={The Myth of the 10x Programmer},
  author={Prechelt, Lutz},
  journal={IEEE Software},
  volume={36},
  number={5},
  pages={96--99},
  year={2019}
}

@article{ng2002discriminative,
  title={On Discriminative vs. Generative Classifiers: A Comparison of Logistic Regression and Naive Bayes},
  author={Ng, Andrew Y and Jordan, Michael I},
  journal={Advances in Neural Information Processing Systems},
  volume={14},
  year={2002}
}
\end{document}